%% file: main.tex
\documentclass[10pt,twocolumn,letterpaper]{article}

\usepackage{iccv}              %

\input{preamble}
\usepackage{tikz}

\usepackage{graphicx}
\setkeys{Gin}{keepaspectratio}
\usepackage{amsmath}
\usepackage{amssymb}
\usepackage{booktabs}
\usepackage{float}
\usepackage{pgfplots}
\usepackage{tikz}
\usetikzlibrary{patterns}
\usepackage[precision=2, unit=mm]{lengthconvert}
\usepackage{diagbox}
\usepackage{enumitem}
\usepackage{tabularx}
\usepackage{multirow}
\usepackage{wrapfig}
\usepackage{cuted}
\usepackage{overpic}
\usepackage[rightcaption]{sidecap}
\usepackage{calc}
\usepackage{bbm}
\usepackage{adjustbox}
\pgfplotsset{compat=1.18} 
\usepackage[accsupp]{axessibility}

\definecolor{mycolor1}{RGB}{90,130,213}%
\definecolor{mycolor3}{RGB}{231,92,46}%
\definecolor{mycolor11}{RGB}{134,168,235}%
\definecolor{mycolor33}{RGB}{231,156,130}%
\definecolor{mycolor2}{RGB}{230,130,46}%
\definecolor{cBLUE}{RGB}{90,130,213}%
\definecolor{cBLUE1}{RGB}{90,183,214}%
\definecolor{cBLUE2}{RGB}{132,217,226}%
\definecolor{cRED}{RGB}{231,92,46}%
\definecolor{cPINK}{RGB}{200,57,170}%
\definecolor{cPINKLIGHT}{RGB}{255,209,245}%
\definecolor{cGREEN}{RGB}{80,150,80}%
\definecolor{cYELLOW}{RGB}{247,179,43}%
\definecolor{cORANGE}{RGB}{242,105,0}%
\definecolor{cGRAY}{RGB}{129,141,146}%

\newcommand{\authorspace}{\hspace{0.7cm}}
\newcommand{\affiliationspace}{\hspace{0.12cm}}

\newcolumntype{?}{!{\vrule width 1pt}}

\newcommand\blfootnote[1]{%
  \begingroup
  \renewcommand\thefootnote{}\footnote{#1}%
  \addtocounter{footnote}{-1}%
  \endgroup
}

\definecolor{iccvblue}{rgb}{0.21,0.49,0.74}
\usepackage[pagebackref,breaklinks,colorlinks,allcolors=iccvblue]{hyperref}

\def\paperID{11223} %
\def\confName{ICCV}
\def\confYear{2025}

\title{
$\chi$: Symmetry Understanding of 3D Shapes via Chirality Disentanglement
}

\author{Weikang Wang$^{*}$
\authorspace
Tobias Weißberg$^{*}$
\authorspace
Nafie El Amrani
\authorspace
Florian Bernard \\
University of Bonn
\affiliationspace \& \affiliationspace
Lamarr Institute
}

\newif\ifhighlight
\highlightfalse %

\newcommand{\changed}[1]{%
  \ifhighlight
    \textcolor{red}{#1}%
  \else
    #1%
  \fi
}

\begin{document}

\twocolumn[{%
\renewcommand\twocolumn[1][]{#1}%
\maketitle
\begin{center}
    \captionsetup{type=figure}
    \input{teaser/teaser}
    \caption{\textbf{Left:} Our chirality features can disentangle the left and right parts of 3D shapes. By leveraging the generalisation abilities of foundation image models, our method remains effective even on partial shapes. \textbf{Right:} Our chirality features resolve left-right ambiguities in 3D shape matching by augmenting state-of-the-art vertex features like Diff3F \cite{dutt2024diffusion} with structural information.}
    \label{fig:teaser}
\end{center}
}]

\blfootnote{
$^*$ Authors contributed equally.}

\input{sec/0_abstract}    
\input{sec/1_introduction}

\input{sec/2_related_works}
\input{sec/3_chirality_optimizations}
\input{sec/4_experiments}

\input{sec/5_ablation}

\input{sec/6_discussion}
\input{sec/7_conclusion}
\input{sec/8_acknowledge}
{
    \small
    \bibliographystyle{ieeenat_fullname}
    \bibliography{main}
}

\input{supplementary}

\end{document}

%% file: teaser/teaser.tex
\newcommand{\imageheightat}{0.12\textheight}
\newcommand{\imageheightbt}{0.12\textheight}
\newcommand{\imagespacingt}{\hspace{0.1cm}}

\begin{tabular}{ccc?cccc} 

\multicolumn{3}{c?}{\textbf{Chirality Features}}
& \multicolumn{4}{c}{\textbf{Shape Matching Results}} \\

& & & Source & Target & Diff3F & Ours \\

\adjustbox{valign=m}{\includegraphics[height=0.12\textheight]{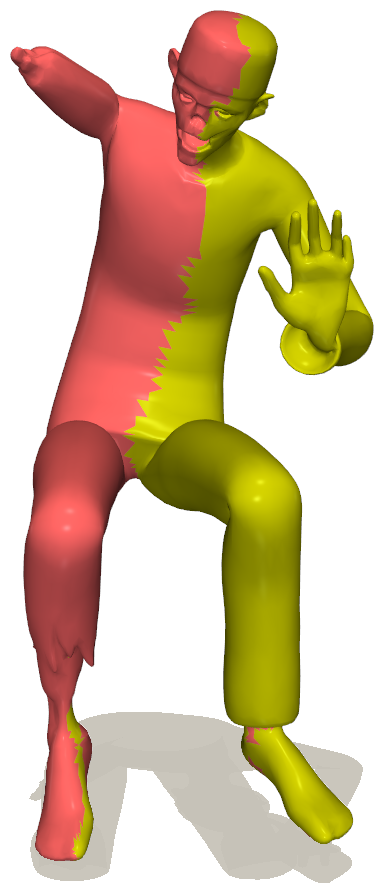}} &
\imagespacingt
\adjustbox{valign=m}{\includegraphics[height=0.12\textheight]{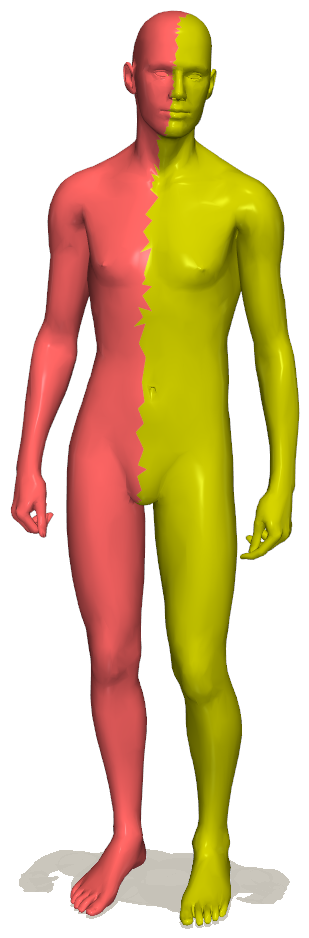}} &
\imagespacingt
\adjustbox{valign=m}{\includegraphics[height=0.1\textheight]{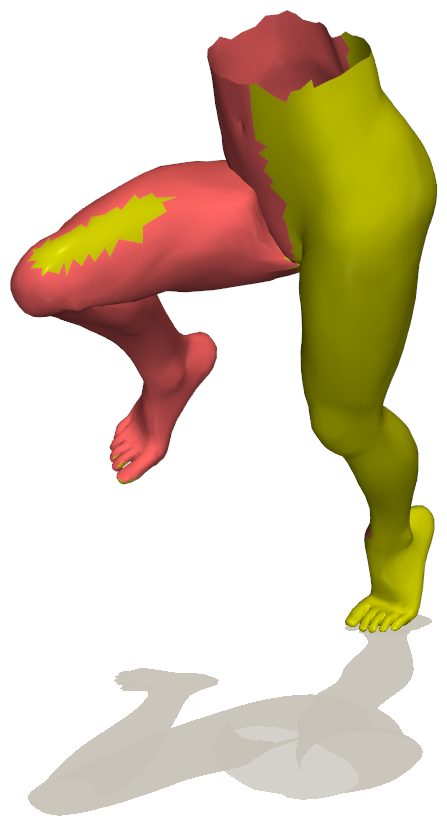}} & 
\imagespacingt
\adjustbox{valign=m}{\includegraphics[height=\imageheightat]{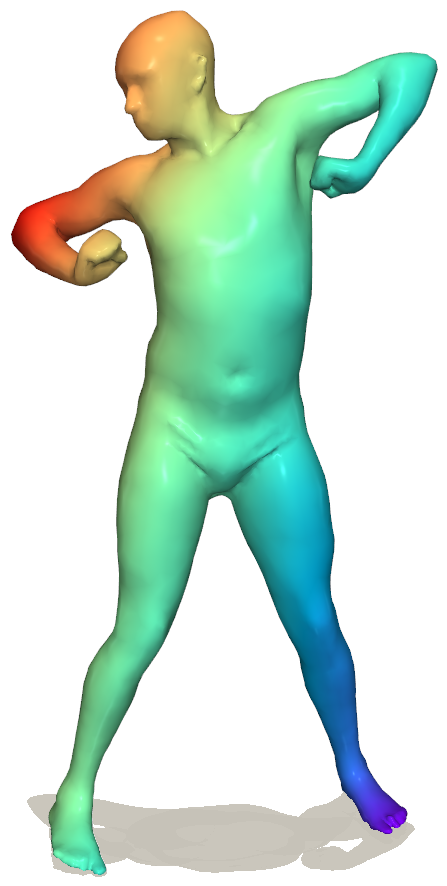}} & 
\imagespacingt
\adjustbox{valign=m}{\includegraphics[height=\imageheightat]{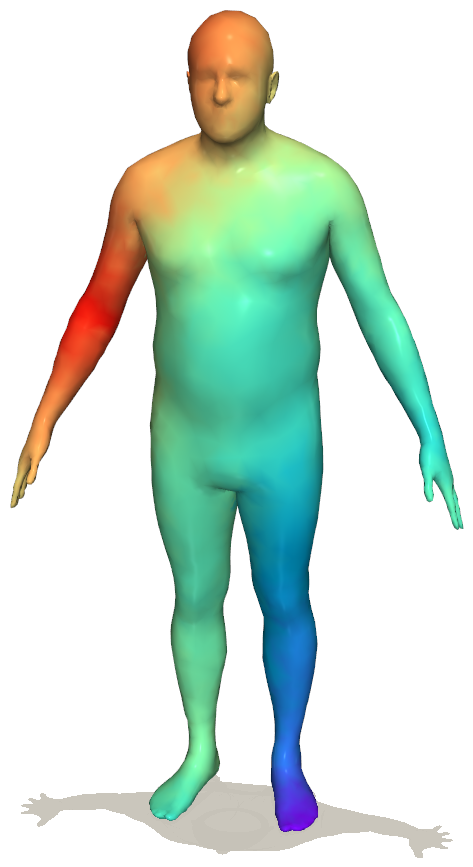}} &
\imagespacingt
\adjustbox{valign=m}{\includegraphics[height=\imageheightat]{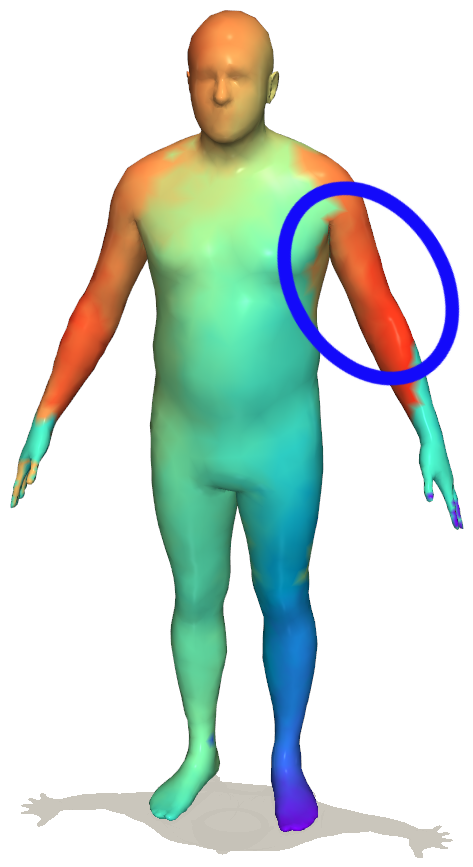}} &
\imagespacingt
\adjustbox{valign=m}{\includegraphics[height=\imageheightat]{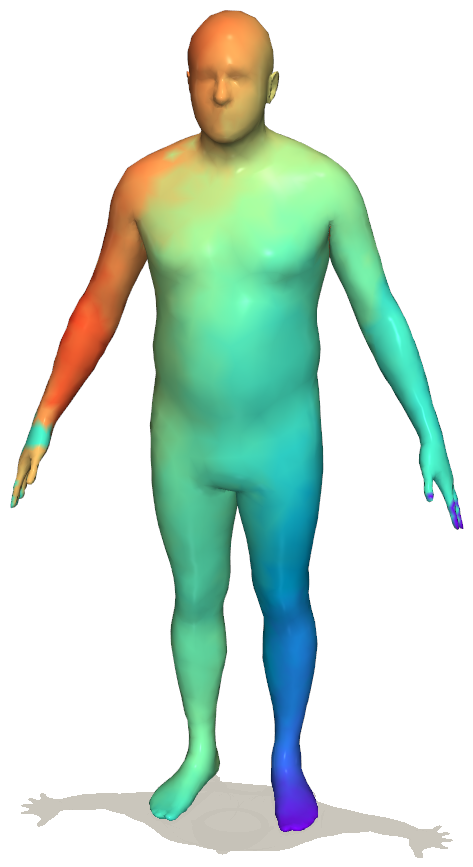}} \\

\adjustbox{valign=m}{\includegraphics[height=0.09\textheight]{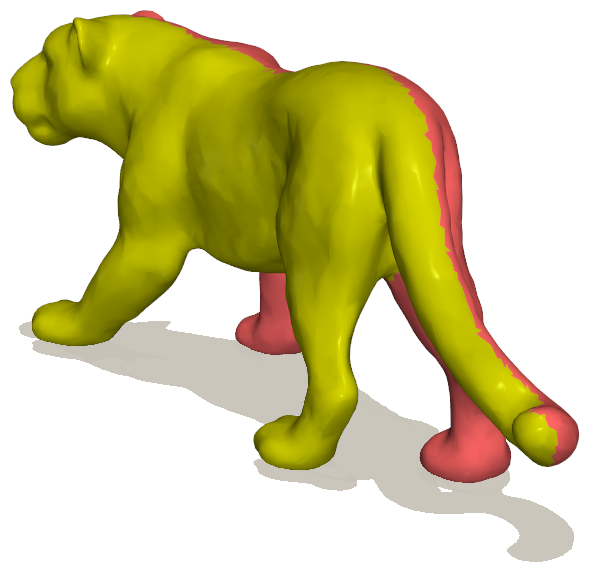}} &
\imagespacingt
\adjustbox{valign=m}{\includegraphics[height=0.09\textheight]{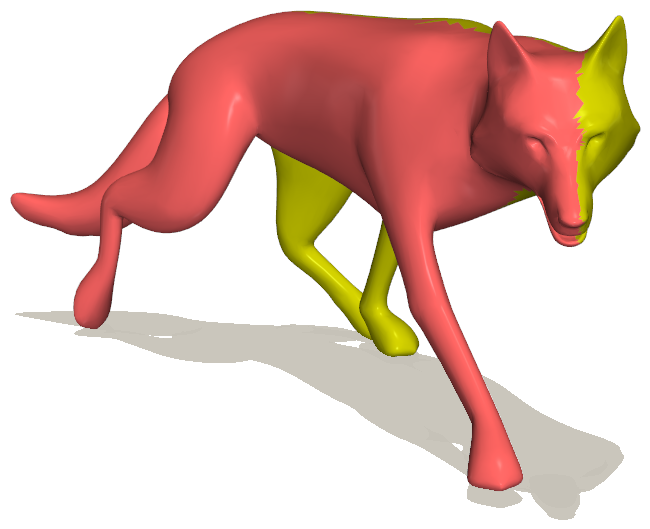}} &
\imagespacingt
\adjustbox{valign=m}{\includegraphics[height=0.09\textheight]{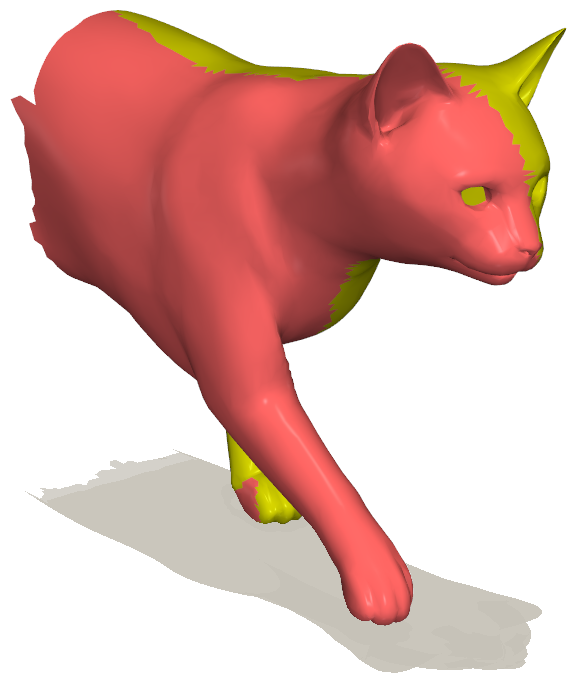}} & 
\imagespacingt
\adjustbox{valign=m}{\includegraphics[height=\imageheightbt]{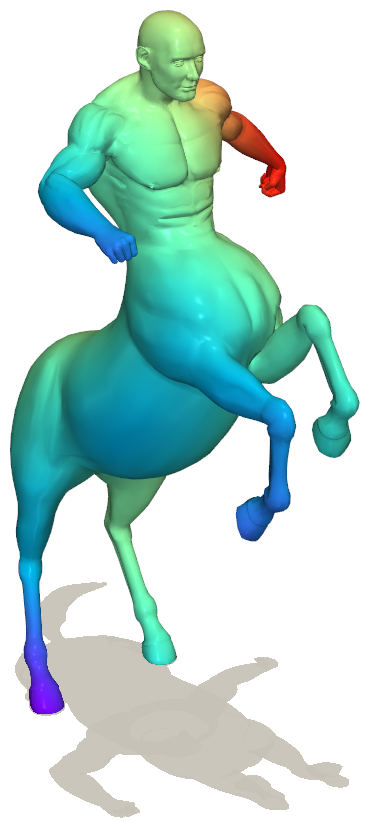}} & 
\imagespacingt
\adjustbox{valign=m}{\includegraphics[height=\imageheightbt]{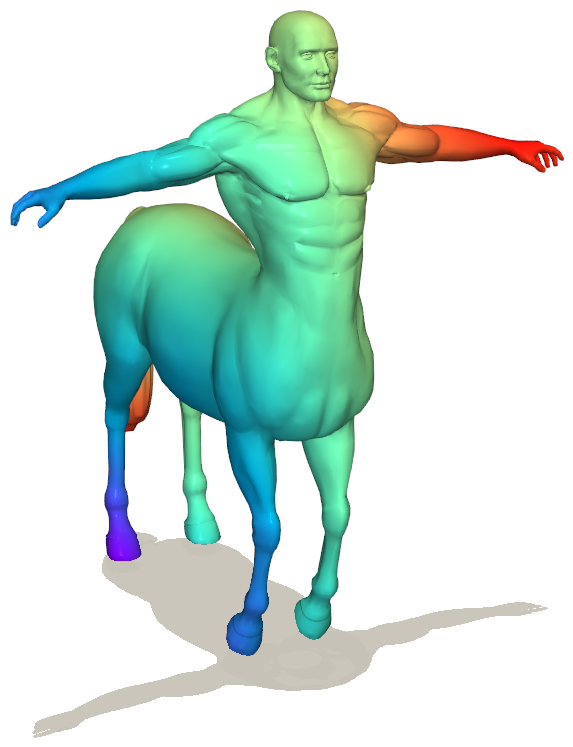}} &
\imagespacingt
\adjustbox{valign=m}{\includegraphics[height=\imageheightbt]{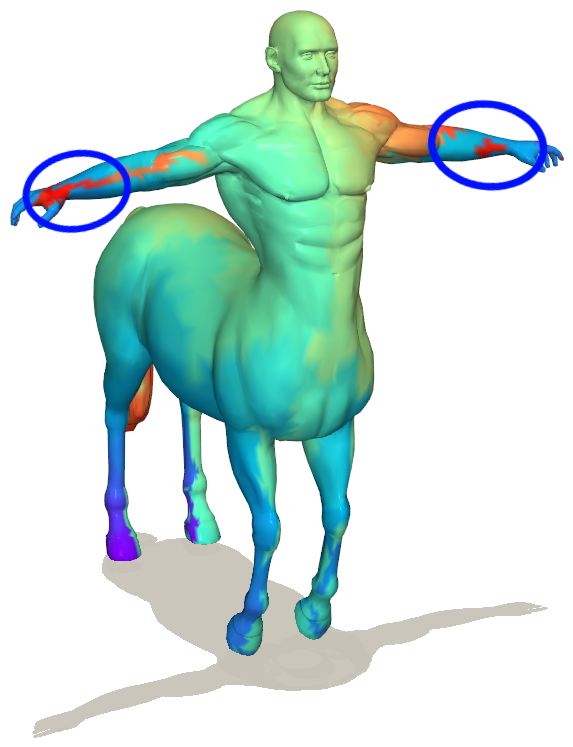}} & 
\imagespacingt
\adjustbox{valign=m}{\includegraphics[height=\imageheightbt]{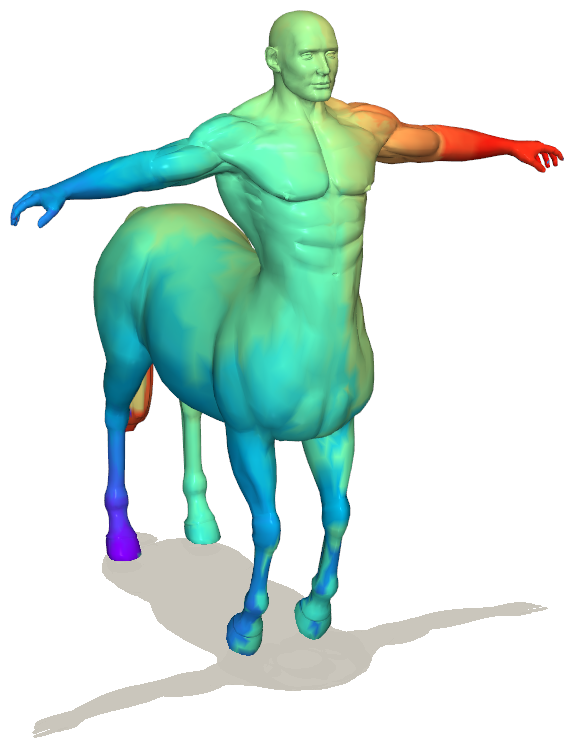}} \\

\end{tabular}

%% file: sec/0_abstract.tex
\begin{abstract}

    Chirality information (i.e.~information that allows distinguishing left from right) is ubiquitous for various data modes in computer vision, including images, videos, point clouds, and meshes. While chirality has been extensively studied in the image domain, its exploration in shape analysis (such as point clouds and meshes) remains underdeveloped. Although many shape vertex descriptors have shown appealing properties (e.g.~robustness to rigid-body transformations), they are often not able to disambiguate between left and right symmetric parts. Considering the ubiquity of chirality information in different shape analysis problems and the lack of chirality-aware features within current shape descriptors, developing a chirality feature extractor becomes necessary and urgent. Based on the recent Diff3F framework \cite{dutt2024diffusion}, we propose an unsupervised chirality feature extraction pipeline to decorate shape vertices with chirality-aware information, extracted from 2D foundation models. We evaluated the extracted chirality features through quantitative and qualitative experiments across diverse datasets. Results from downstream tasks including left-right disentanglement, shape matching, and part segmentation demonstrate their effectiveness and practical utility. Project page: \href{https://wei-kang-wang.github.io/chirality/}{https://wei-kang-wang.github.io/chirality/}

\end{abstract}

%% file: sec/1_introduction.tex
\section{Introduction}

Symmetry and chirality are two sides of the same coin: Symmetry highlights the similarities between two parts, whereas chirality focuses on the differences between them. As a broadly applicable and intuitive assumption for many objects in the physical world, symmetry has been extensively studied in various computer vision domains for a long time, including 3D reconstruction \cite{wu2020unsupervised, wu2023magicpony}, pose estimation \cite{corona2018pose, pitteri2019object, zhao2023learning, merrill2022symmetry}, and generative models \cite{yang2023generative, allingham2024generative}.  However, chirality, despite its close relationship to symmetry, remains less explored and has just re-attracted researchers' attention in recent years \cite{lin2020visual, lo2022facial, tan2022mirror, zheng2021visual, yeh2019chirality}.

\input{tables/motivation}

In the field of shape analysis, symmetry and chirality also play an important role in many problems, including matching \cite{liu2012finding, yoshiyasu2014symmetry, zhang2013symmetry, cosmo2017consistent}, deformation \cite{zheng2015skeleton}, symmetry plane detection \cite{podolak2006planar}, etc. For many shape analysis problems, vertex feature descriptors play a central role, and various methods to generate feature descriptors have been proposed \cite{jian2009heat, mathieu2011wave, cao2022unsupervised}. Recently, Diff3F \cite{dutt2024diffusion} was introduced as a novel framework for generating vertex feature descriptors. It aggregates 2D foundation model features from rendered multi-view images of a shape to create vertex feature descriptors. While recent features have shown semantic and geometric robustness, none of them are able to disambiguate between left-and-right symmetric shape parts, which could cause severe problems, as shown in Fig.~\ref{fig:motivation}.

In order to fill this gap, we propose an unsupervised method based on Diff3F \cite{dutt2024diffusion} that disentangles left-and-right (chirality) information to decorate shapes with chirality-aware vertex descriptors using features aggregated from 2D foundation models (such as DINO-V2 \cite{oquabdinov2} and StableDiffusion \cite{rombach2022high}). We conduct experiments on various datasets and show that our extracted chirality feature provides left-and-right information of a shape. 
We show that left-and-right ambiguities, which are a common issue of state-of-the-art shape matching methods \cite{cao2022unsupervised, dutt2024diffusion}, can be effectively mitigated by combining our chirality feature with other vertex feature descriptors.
The generalization experiments of our pre-trained model on partial shapes and anisotropic shapes additionally prove the robustness of our chirality feature extractor.\\

\noindent To summarize, our contributions 
are as follows:
\begin{itemize}
    \item We propose an unsupervised method to extract left-and-right (chirality) information for decorating shape vertex descriptors. Left-and-right disentanglement experiments on various datasets show the validity of our extracted chirality features quantitatively and qualitatively.
    \item For shape matching and part segmentation, we demonstrate that augmenting standard vertex descriptors with our chirality features effectively resolves left-right ambiguities across multiple datasets.
    \item Finally, cross-dataset and cross-category generalisation experiments, particularly those conducted on partial and anisotropic shapes, demonstrate the robustness of our trained chirality feature extractor.
\end{itemize}

\label{sec:intro}

%% file: tables/motivation.tex
\begin{figure}[!ht]
    \centering
    \begin{tabular}{c?ccc}
         SD+DINO & Source &  Target & Diff3F \\

         \adjustbox{valign=m}{\includegraphics[height=0.09\textheight]{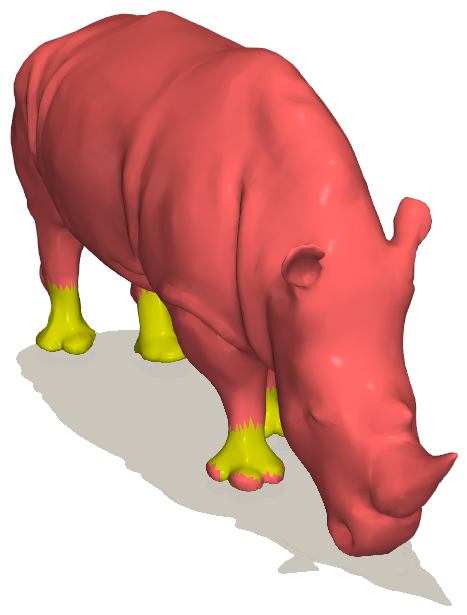}} &
         \adjustbox{valign=m}{\includegraphics[height=0.09\textheight]{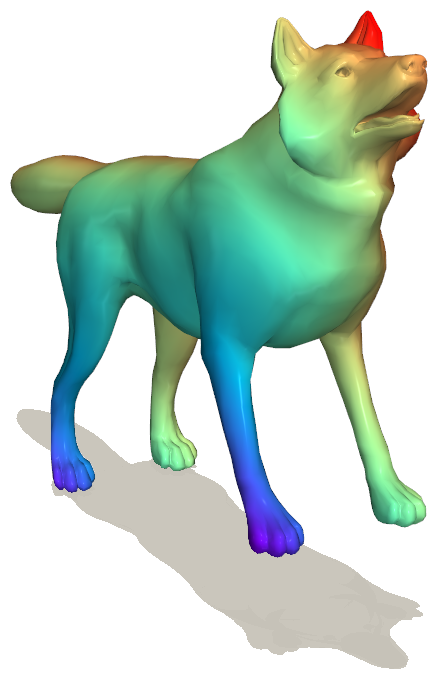}} & 
         \adjustbox{valign=m}{\includegraphics[height=0.085\textheight]{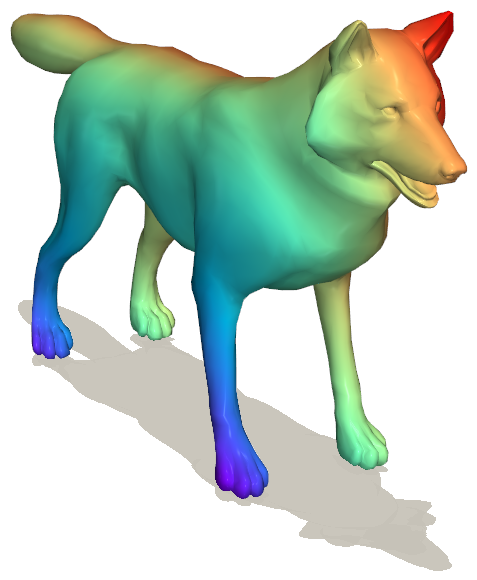}} & 
         \adjustbox{valign=m}{\includegraphics[height=0.085\textheight]{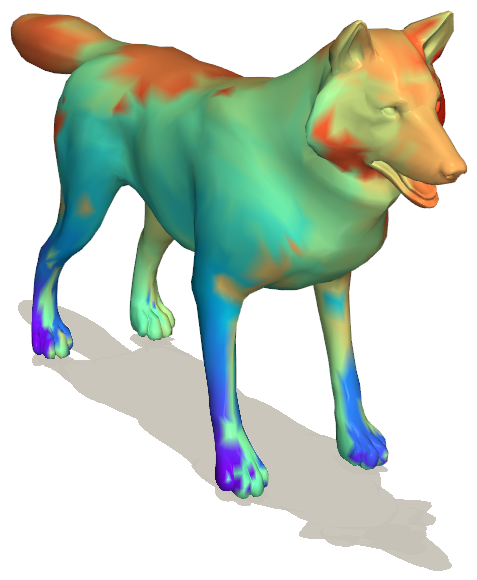}}\\ 

         \textbf{(a) Left/Right} &  \multicolumn{3}{c}{\textbf{(b) Shape matching}}

    \end{tabular}
    
    \caption{\textbf{Left}: The 2-center clustering of vertex features aggregated from StableDiffusion and DINO \cite{rombach2022high, oquabdinov2} features does not lead to the desired chirality indicator. \textbf{Right}: Shapes matched using Diff3F \cite{dutt2024diffusion} display a lack of chirality awareness.}
    \label{fig:motivation}
\end{figure}

%% file: sec/2_related_works.tex
\section{Related work}
\label{sec:related}

We briefly review related work on chirality (Sec.~\ref{subsec:chirality}), feature extraction (Sec.~\ref{subsec:features}), and shape matching (Sec.~\ref{subsec:shape_matching}).
\subsection{Chirality in visual computing}
\label{subsec:chirality}

In the context of studies on images, \citet{lin2020visual} explored how the statistics of visual data change under reflection and coined the term visual chirality. \citet{yeh2019chirality} proposed a network whose layers can be mirror-flipped, using it for human pose estimation by resolving left/right ambiguities in the human body. \citet{zheng2021visual} investigated the vertical flipping associated with visual chirality in freehand sketches. Moreover, \citet{tan2022mirror} leveraged the property of image-level visual chirality and reformulated it as a learnable pixel-level cue for mirror detection. In facial analysis, \citet{lo2022facial} employed chirality information from the left and right halves of faces to learn robust facial embeddings for expression recognition. Recently, to obtain geometry-aware features, \citet{zhang2024telling} fine-tuned a model on aggregated DINO-V2 \cite{oquabdinov2}+SD \cite{rombach2022high} features with labelled keypoint correspondences between image pairs, inherently capturing chirality information. 

In the field of shape analysis, chirality extraction is closely related to the detection of intrinsic or extrinsic shape symmetries. \citet{tevs2014relating} showed that finding correspondences between shapes of widely varying geometry could benefit from extrinsic symmetry. \citet{je2024robust} applied Langevin dynamics within a redefined symmetry space to enhance robustness of extrinsic symmetry detection against noise. E3Sym \cite{li2023e3sym} established robust point correspondences using E(3)-invariant features extracted from a lightweight neural network, enabling dense symmetry predictions. \citet{ovsjanikov2008global} utilized Euclidean symmetries in the signature space defined by the eigenfunctions of the Laplace-Beltrami operator to identify intrinsic symmetries. \citet{kim2010mobius} modelled intrinsic symmetry as Möbius transformations derived from critical points of the Average Geodesic Distance (AGD) function. \citet{liu2012finding} detected intrinsic symmetry on genus-zero mesh surfaces by extracting closed curves through conformal maps based on triplets of extrema points identified via the AGD function. Additionally, \citet{nagar2018fast} hypothesized that if a shape was intrinsically symmetric, the shortest geodesic between two symmetric points was also intrinsically symmetric, facilitating the extraction of intrinsic correspondences.

However, to our best knowledge, there exist no methods in the shape analysis field to extract chirality-aware vertex feature descriptors. Based on the recent paper Diff3F \cite{dutt2024diffusion}, which aggregated features from 2D foundation models (DINO-V2 \cite{oquabdinov2} and StableDiffusion \cite{rombach2022high}) to obtain vertex descriptors, we propose an unsupervised pipeline to distil chirality information from features extracted by 2D foundation models to decorate vertex feature descriptors.

\subsection{Feature descriptors from 2D foundation models}
\label{subsec:features}
In recent years, various 2D foundation models have been proposed, including DINO \cite{caron2021emerging}, DINO-V2 \cite{oquabdinov2}, CLIP \cite{radford2021learning}, StableDiffusion (SD) \cite{rombach2022high}, etc. They have been used in various 2D/3D areas as feature descriptors and have surpassed both handcrafted features and other deep features, due to their rich semantic and geometric information obtained from large-scale and/or multi-modal datasets. \citet{hedlin2023unsupervised} treated SD features as pixel-level feature descriptors to do semantic correspondence, while \citet{hedlin2024unsupervised} leveraged SD features as an unsupervised 2D keypoint detector. FeatureNeRF \cite{ye2023featurenerf} learned generalizable NeRFs by distilling pre-trained vision foundation models to perform on other downstream tasks beyond synthesis, such as semantic understanding and parsing. In this paper, similar to the setting in Diff3F \cite{dutt2024diffusion} (see Sec.~\ref{diff3f} for more details), we also aggregate per vertex features from 2D foundation model features of multiple views. However, we additionally derive chirality information embedded in the 2D foundation model features to augment vertex descriptors, enabling them to be left-right aware and consistent across shapes.

\subsection{Shape matching}
\label{subsec:shape_matching}
Shape matching is a well-studied problem in computer vision and graphics \cite{tam2013registraion, kaik2011survey}, aimed at finding correspondences between pairs of shapes. Classical approaches solve this problem by establishing correspondences based on geometric relations \cite{holzschuh2020simulated, roetzer2022scalable}, while other methods rely on non-rigid shape registration \cite{bernard2020mina, eisenberger2019divergence, huang2008non}. 
One prominent mesh-based approach, the functional map framework \cite{ovsjanikov2012functional}, encodes shape correspondences into a compact matrix using truncated basis functions, usually the first $k$ Laplacian eigenfunctions \cite{levy2006lbo}. Although it has been extended to handle non-isometries \cite{nogneng2017informative, ren2019structured}, partial shapes \cite{attaiki2021dpfm, xie2025echomatch} or multi-shape matching \cite{cao2022unsupervised, gao2021multi}, its reliance on spectral basis calculation involves high computational costs, pre-processing, mesh connectivity, and lacks semantic consideration.

An alternative line of works focuses on spatial approaches and operates on point clouds directly, such as 3D-Coded \cite{groueix20183d} and \citet{deprelle2019elementary}, which deform a template shape to find correspondences. Other methods use supervised learning for point-to-point correspondences \cite{donati2020deep, yew2020RPMNet, Wang_2019_ICCV}, while recent unsupervised approaches for point clouds, such as SE-ORNet \cite{Deng2023seornet}, first align point clouds and then use a teacher-student network with DGCNN backbone \cite{wang2019dgcnn} to learn point-wise embeddings. DPC \cite{lang2021dpc} leverages self- and cross-attention to learn discriminative per-point features for smooth mappings. While both mesh- and point cloud-based methods achieve good matching performance, they primarily focus on geometric features and overlook semantic information. To address this, Diff3F \cite{dutt2024diffusion} leverages foundation models to exctract semantic vertex features. Our method builds on Diff3F \cite{dutt2024diffusion} by incorporating chirality information, which makes features left/right aware and improves the quality of shape matching.

%% file: sec/3_chirality_optimizations.tex
\section{Chirality feature optimization}
\label{optimization}

\begin{figure*}[ht!]
    \centering
    \includegraphics[width=\textwidth]{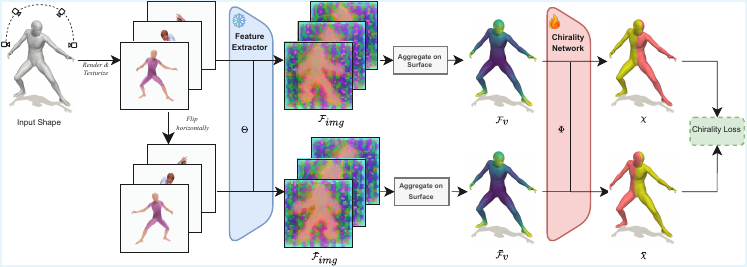}
    \caption{Overview of the pipeline of our method. We consider a single 3D mesh for which we generate $N$ textured images following Diff3F~\cite{dutt2024diffusion}. We flip the images horizontally to receive $N$ pairs of textured images. The images are then independently processed by a frozen feature extractor, supplying us with features $F_{img}$ and $\bar{F}_{img}$. After reprojecting onto the shape, we aggregate the features to receive $\mathcal{F}_{v}$ and $\bar{\mathcal{F}}_{v}$. Finally, we train a chirality network $\tilde{g}_{\Phi}$ on $\mathcal{F}_{v}$ and $\bar{\mathcal{F}}_{v}$ to learn chirality features $\chi$ and $\bar{\chi}$ that tells left from right.}
    \label{fig:method_overview}
\end{figure*}

In this section, we introduce our unsupervised pipeline to decorate an untextured shape with chirality-aware features.  Our method is based on Diff3F \citep{dutt2024diffusion}, a recently proposed method that utilizes in-painting diffusion \cite{rombach2022high} along with DINO-V2 \cite{oquabdinov2} features to decorate untextured shapes with semantical aware features. We briefly introduce Diff3F \cite{dutt2024diffusion} in Sec.~\ref{diff3f}. In Sec.~\ref{imgflip}, we introduce the generation of chiral image pairs and their features, and also the aggregation from image features to per-vertex chiral feature pairs, which are the core idea of our method. Finally, in Sec.~\ref{chiralityDisentanglement}, we introduce our model architecture and our unsupervised losses.

\subsection{Background: Diff3F}
\label{diff3f}
In this subsection, we provide a summary of the Diff3F \cite{dutt2024diffusion} method. We consider a 3D shape $\mathcal{X}$ represented by a vertex set $V \in \mathbb{R}^{|V| \times 3}$ and an edge set $E \in \mathbb{N}^{|E| \times 2}$. Diff3F \cite{dutt2024diffusion} generates vertex-wise features as follows: 
\begin{enumerate}
    \item Rendering from $N$ different camera poses of shape $\mathcal{X}$ results in $N$ images $I_j^S \in \mathbb{R}^{H \times W \times 3}$ , corresponding depth map $D_j \in \mathbb{R}^{H \times W \times 1}$ and normal map $N_j \in \mathbb{R}^{H \times W \times 3}$, where $j \in \{0, \dots, N-1\}$ and $H$ and $W$ denote the height and width, respectively. 
    \item Using a pre-trained StableDiffusion model \cite{rombach2022high} with ControlNet \cite{zhang2023adding}, the images $I_j^S$ are textured with the guidance of depth and normal maps and a category prompt of the shape to produce  $I_j^{\text{tex}} \in \mathbb{R}^{H \times W \times 3}$.
    \item During the texturing process, diffusion features from multiple layers are extracted and aggregated over multiple time steps of the diffusion process to form a set of features for each image, $F_j^{\text{SD}} \in \mathbb{R}^{H \times W \times 1280}$.
    \item The textured images $I_j^{\text{tex}}$ are fed through DINO-V2 \cite{oquabdinov2} to extract features $F_j^{\text{DINO}} \in \mathbb{R}^{H \times W \times 768}$.
    \item Diffusion ($F_j^{\text{SD}}$) and DINO-V2 ($F_j^{\text{DINO}}$) features are then normalised and concatenated to form the final image features $F_j^{\text{img}} \in \mathbb{R}^{H \times W \times 2048}$.
    \item Using the camera poses, each $F_j^{\text{img}}$ is projected back onto the vertices of $\mathcal{X}$ and the final feature $\mathcal{F}_{v}$ for vertex $v \in V$ is the average of aggregated features from all $N$ camera poses.
\end{enumerate}

\noindent For convenience, we refer to StableDiffusion as SD throughout the remainder of this paper. Similarly, we abbreviate DINO-V2 as DINO, as it is the only version used in this work.

\subsection{Generating chiral pairs}
\label{imgflip}

In our method, the input is an untextured mesh $\mathcal{X}$, defined by a vertex set $V \in \mathbb{R}^{|V| \times 3}$ and an edge set $E \in \mathbb{N}^{|E| \times 2}$, without any feature descriptors but accompanied by a category label used for diffusion guidance. 

Following Diff3F \cite{dutt2024diffusion}, we get the textured images $\lbrace I_j \rbrace_{j=1}^N$. Then, we flip these textured images $\lbrace I_j \rbrace_{j=1}^N$ horizontally to get a set of flipped textured images $\lbrace \bar{I}_j \rbrace_{j=1}^N$. Note that flipping the image vertically is equivalent to flipping the image horizontally together with a 180 degree in-plane rotation, and that in-plane rotations will not change the chirality of objects within an image.

After that, we feed both the textured image set $\lbrace I_j \rbrace_{j=1}^N$ and the flipped textured image set $\lbrace \bar{I}_j \rbrace_{j=1}^N$ into DINO \cite{oquabdinov2} and StableDiffusion \cite{rombach2022high} to get the feature sets $\lbrace (F^{\text{SD}}_j, F^{\text{DINO}}_j) \rbrace_{j=1}^N$, $\lbrace (\bar{F}^{\text{SD}}_j, \bar{F}^{\text{DINO}}_j) \rbrace_{j=1}^N$.\footnote{Note that Diff3F uses ControlNet \cite{zhang2023adding} to add textures to untextured images, and the texture adding is done on the latent feature space obtained from the untextured image using the encoder. Due to the nonlinearity of the encoder and the randomness involved in the process of obtaining SD features, it is impossible to get SD features of an image and its flipped counterpart with the same texture generated by ControlNet \cite{zhang2023adding}, which contradicts our requirements. Thus we follow \citet{zhang2024telling} to get SD features for both textured image and flipped textured image to fulfil the consistency of feature semantics.}
We adopt a similar approach to \cite{dutt2024diffusion, zhang2024telling} for concatenating features from each DINO and SD feature pair: each feature is first normalized individually, then concatenated along the feature dimension, followed by normalization of the combined feature vector. 
This process yields $\lbrace F_j^{\text{img}} \rbrace_{j=1}^N$ and $\lbrace \bar{F}_j^{\text{img}} \rbrace_{j=1}^N$ corresponding to the textured images and their flipped versions, respectively. Next, we flip $\lbrace \bar{F}_j^{\text{img}} \rbrace_{j=1}^N$ horizontally to get $\lbrace \hat{F}_j^{\text{img}} \rbrace_{j=1}^N$, which is spatially aligned with the original features $\lbrace F_j^{\text{img}} \rbrace_{j=1}^N$.

Finally, for each vertex $v \in V$ of shape $\mathcal{X}$, we utilize the camera information to locate its corresponding pixels in each rendered view. The vertex feature $\mathcal{F}_v$ is given by averaging the features of these pixels. Similarly, per vertex feature $\bar{\mathcal{F}}_v$ are obtained from $\hat{F}^{\text{img}}$. This process results in the chirality vertex feature pair $(\mathcal{F}_v,\bar{\mathcal{F}}_v)$.

Using this construction, $\mathcal{F}_v$ aggregates semantic and geometric information from the image foundation model features, while its counterpart $\bar{\mathcal{F}}_v$ aggregates information differing only in left-and-right (chirality) information. For example, if $v$ denotes the keypoint at the left eye in shape $\mathcal{X}$, then $\mathcal{F}_v$ will average {\bf left} eye image foundation model features from different views, while $\bar{\mathcal{F}}_v$ gathers {\bf right} eye features from different views (note that $v$ still denotes the left eye keypoint of $\mathcal{X}$, and these right eye features are virtually constructed and gathered by our flipping process).

\subsection{Chirality feature extraction}
\label{chiralityDisentanglement}

In this section, we describe the extraction of our chirality features $\chi, \bar{\chi}$ from the generated pairs of vertex features $\lbrace (\mathcal{F}_v, \bar{\mathcal{F}}_v) \rbrace_{v \in V}$, where $\mathcal{F}_v, \bar{\mathcal{F}}_v \in \mathbb{R}^{D}$.
We denote the stacked vertex features of $\lbrace \mathcal{F}_v \rbrace_{v \in V}$ and $\lbrace \bar{\mathcal{F}}_v \rbrace_{v \in V}$ as $\mathcal{F} \in \mathbb{R}^{\vert V \vert \times D}$ and $\bar{\mathcal{F}} \in \mathbb{R}^{\vert V \vert \times D}$, respectively.

To this end, we use an encoder $g_{\Phi}: \mathbb{R}^D \to \mathbb{R}^D$ together with a linear projection layer $A \in \mathbb{R}^{D \times D}$ as our chirality feature extractor denoted as $\tilde{g}(\cdot)=A g_{\Phi}(\cdot)$, with $\Phi$ and $A$ being the optimised parameters. A decoder $h_{\Psi}: \mathbb{R}^D \to \mathbb{R}^D$ is included to avoid trivial solutions.

The chirality feature vector $\chi := (\chi_{v_1},\cdots,\chi_{v_{|V|}})^{\top} \in \mathbb{R}^{\vert V \vert \times 1}$ of all vertices of shape $\mathcal{X}$ is given by

\begin{equation}
    \chi_v := \frac{[\tilde{g}(\mathcal{F}_v)]_1}{\|\tilde{g}(\mathcal{F}_v)\|_2} = \frac{[Ag(\mathcal{F}_v)]_1}{\|Ag(\mathcal{F}_v)\|_2},
\end{equation}
where $[ \cdot ]_1$ denotes taking the first entry of the feature vector. This indexing and normalisation will ensure that chirality features $\chi_{v}$ are in the range $[-1, 1]$. Similarly, we obtain $\bar{\chi}$ from $\bar{\mathcal{F}}$ using the model $\tilde{g}$. More details about the architecture of our feature extractor are included in Sec.~\textcolor{iccvblue}{B} in the supplementary material. 
To train our model, we employ the following losses:
\paragraph{Dissimilarity loss $\mathcal{L}_{\text{dis}}$.} The dissimilarity loss is defined as
\begin{equation}
    \mathcal{L}_{\text{dis}} = - \frac{1}{\sqrt{\vert V \vert}} \lVert \chi - \bar{\chi} \rVert_2.
\end{equation}
It serves as the main loss, since it maximises the difference between the chirality feature obtained from $\mathcal{F}_v$ and $\bar{\mathcal{F}_v}$, which should only differ in left-and-right information.

\paragraph{Invertibility loss $\mathcal{L}_{\text{inv}}$.}
To keep our encoder $g$ from learning trivial solutions, we introduce a decoder $h_{\Psi}$, where $\Psi$ is the learnable parameter. The invertibility loss
\begin{equation}
    \mathcal{L}_{\text{inv}} = \frac{1}{\sqrt{|V|}}\lVert [\mathcal{F}^\top \;\; \bar{\mathcal{F}}^\top]^\top - h(g([\mathcal{F}^\top \;\; \bar{\mathcal{F}}^\top]^\top)) \rVert_F,
\end{equation} 
ensures that $h$ is able to reconstruct the inputs $\mathcal{F}$ and $\bar{\mathcal{F}}$ from the outputs $g(\mathcal{F})$ and $g(\bar{\mathcal{F}})$, respectively.
Here, $[\mathcal{F}^\top \;\; \bar{\mathcal{F}}^\top]^\top$ corresponds to stacking the rows of $\mathcal{F}$ and $\bar{\mathcal{F}}$, and $g$ and $h$ are applied row-wise.
\paragraph{Total variation loss $\mathcal{L}_{\text{var}}$.}
To achieve spatial smoothness, we add a total variation loss
\begin{equation}
\begin{aligned}
    \mathcal{L}_{\text{var}} = \frac{1}{\vert E \vert} \sum_{(u, v) \in E} & \lVert \chi_u - \chi_v \rVert_1 + \lVert \bar{\chi}_u - \bar{\chi}_v \rVert_1,
    \end{aligned}
\end{equation}
where $\vert E \vert$ is the number of edges inside the mesh.

\paragraph{Fifty-fifty loss $\mathcal{L}_{\text{fif}}$.} %
The total variation loss $\mathcal{L}_{\text{var}}$ introduces a bias towards solutions with small boundary length. We counteract this by introducing

\begin{equation}
    \mathcal{L}_{\text{fif}} = \frac{1}{\vert V \vert} \left(\frac{\vert \chi^{\top} \mathbf{1}_{|V|} \vert}{\lVert \chi \rVert_{\infty}} + \frac{\vert \bar{\chi}^{\top} \mathbf{1}_{|V|} \vert}{\lVert \bar{\chi} \rVert_{\infty}} \right),
\end{equation}
which penalises solutions which assign more vertices to one of the two halves of each shape.

\noindent The overall training loss is the linear combination of the above losses, i.e.
\begin{equation}
    \mathcal{L} = \mathcal{L}_{\text{dis}} + \lambda_1 \mathcal{L}_{\text{inv}} + \lambda_2 \mathcal{L}_{\text{var}} + \lambda_3 \mathcal{L}_{\text{fif}}.
\end{equation}

%% file: sec/4_experiments.tex
\section{Experiments}
\label{sec:experiments}

\input{tables/left_right_results}

\input{tables/qualitative_left_right}

We conduct various experiments to show the effectiveness of our chirality feature. In Sec.~\ref{sec: left-right}, we compare the left-and-right disentanglement performance using our chirality feature compared with other shape feature descriptors and methods. Then, in Sec.~\ref{sec: matching} and Sec.~\ref{sec: segmentation}, we show the effectiveness of our chirality feature in various shape analysis tasks, including shape matching and shape segmentation. To evaluate the generalisation ability of our method, we test on partial and anisotropic shapes in Sec.~\ref{sec:partial_shapes} and Sec.~\ref{sec:anisotropic_shapes}, respectively. Finally, the ablation study of losses and 2D foundation features are conducted in Sec.~\ref{sec: ablation}. In each experiment section, we present the datasets, baselines and evaluation metrics used. For more details about the implementation, see Sec.~\textcolor{iccvblue}{E} in the supplementary material.

\subsection{Left-right disentanglement}
\label{sec: left-right}

\paragraph{Datasets.} To evaluate our approach's ability to distinguish between left and right of 3D shapes, we use datasets where left/right annotations are available. To this end, we use BeCoS \cite{ehm2025becos}, a recently proposed framework to generate shape matching datasets with dense correspondences and left/right annotations. This framework combines shapes from multiple shape matching datasets and connects them with cross-category and cross-class dense correspondences. Using this framework, we generate multiple versions to evaluate the performance and generalisation of our approach. 
\begin{itemize}
    \item \textbf{BeCoS} consists of humanoid and four-legged animals with 1980/284/274 train/test/validation split, generated from 7 different shape datasets, namely TOSCA \cite{bronstein2008numerical}, FAUST \cite{bogo2014faust}, SCAPE \cite{anguelov2005scape}, KIDS \cite{rodola2014dense}, DT4D \cite{magnet2022smooth}, SMAL \cite{Zuffi_CVPR_2017} and SHREC’20 \cite{dyke2020shrec}. See \cite{ehm2025becos} for more details. 
    \item \textbf{BeCoS\textsubscript{-h}} consists of only humanoid shapes from  TOSCA \cite{bronstein2008numerical}, FAUST \cite{bogo2014faust}, SCAPE \cite{anguelov2005scape}, KIDS \cite{rodola2014dense}, DT4D \cite{magnet2022smooth} with 366/64/58 train/test/validation split. 
    \item \textbf{BeCoS\textsubscript{-a}} consists of only four-legged animals from TOSCA \cite{bronstein2008numerical}, DT4D \cite{magnet2022smooth}, SMAL \cite{Zuffi_CVPR_2017} and SHREC’20 \cite{dyke2020shrec} with 1614/220/216 train/test/validation split. 
\end{itemize}
For more details about the data generation please refer to Sec.~\textcolor{iccvblue}{A} in the supplementary material.
Additionally, we use well-established shape matching datasets to evaluate our method, namely FAUST \cite{bogo2014faust}, SCAPE \cite{anguelov2005scape}, SMAL \cite{zheng2015skeleton}, and TOSCA \cite{bronstein2008numerical}. We use the same train/test/validation splits given by BeCoS \cite{ehm2025becos}. Since these datasets do not provide left/right annotations, we use the BeCoS \cite{ehm2025becos} framework to get these annotations. 

\paragraph{Baselines.}
We compare the performance of our chirality features against other feature descriptors, including Diff3F \cite{dutt2024diffusion}, DINO$+$SD \cite{oquabdinov2, rombach2022high}, and DINO$+$SD fine-tuned with \citet{zhang2024telling}. We also include an axiomatic method \citet{liu2012finding} that extracts closed intrinsic/extrinsic symmetric curves on surfaces of a genus-0 mesh that splits the mesh into left and right parts.

\paragraph{Evaluation metrics.}
Given a set of shapes $X = \lbrace \mathcal{X}_1, ...,\mathcal{X}_N \rbrace$, where each shape $\mathcal{X}_n$ has a vertex set $V_{\mathcal{X}_n}$, the chirality accuracy $\text{acc}_{\chi}$ is defined as
\begin{equation}
    \text{acc}_{\chi} = \max \lbrace \text{acc}, 1-\text{acc} \rbrace,
    \label{eq:acc_chi}
\end{equation}
where 
\begin{equation}
    \text{acc} = \frac{1}{N} \sum_{n=1}^N \frac{1}{\vert V_{\mathcal{X}_n} \vert} \sum_{v \in V_{\mathcal{X}_n}} \mathbbm{1}(\text{sign}(\chi_{v}) = \chi_{v}^{gt}).
    \label{eq:_acc}
\end{equation}
$\chi_{v}$ is our chirality features for vertex $v \in V_{\mathcal{X}_n}$ of each shape $\mathcal{X}_n$, $\mathbbm{1}$ is the indicator function, and $\chi_{v}^{gt}$ is the ground truth of left/right annotation of vertex $v \in V_{\mathcal{X}_n}$.

Note that in Eq.~\ref{eq:acc_chi}, we use the maximum of $\text{acc}$ and $1-\text{acc}$ from Eq.~\ref{eq:_acc}, because there is no inherent assignment of whether $\chi_v > 0$ or $\chi_v < 0$ corresponds to left or right.

\paragraph{Experiment results.} The results are summarised in Tab.~\ref{tab:chirality_accuracy}. Our method substantially outperforms all baselines in the context of distinguishing left/right of 3D shapes. Additionally, our method achieves good cross-dataset and cross-category generalisation ability. Fig.~\ref{fig:qualitative_left_right}, provides qualitative results of our method compared to baselines. 

\subsection{Shape matching}
\label{sec: matching}
In this section, we evaluate the performance of our chirality features on shape matching. To ensure a fair comparison with prior work, we adopt a similar experimental setup as in DPC \cite{lang2021dpc}, SE-ORNet \cite{deng2023se} and Diff3F \cite{dutt2024diffusion}.

\input{tables/shape_matching_results}
\input{tables/pck_curves_fig}
\input{tables/qualitative_matching}

\paragraph{Datasets.} We test our features on both human and animal shapes. For human shapes, we evaluate our method on SHREC'19 \cite{melzi2019shrec}. This dataset comprises of 44 human scans with a test set of 430 shape pairs. We use the more challenging re-meshed version from \cite{donati2020deepGeoMaps}. For the non-human shapes, we use only animal the 41 shapes from TOSCA \cite{bronstein2008numerical} (noted as TOSCA-a) and pair shapes from the same category to create a testing set of 286 pairs. 

\paragraph{Baselines.} We compare our approach to recente state-of-the-art shape matching methods. One supervised method (3D-CODED \cite{groueix20183d}) that we only have access to its model trained on humans. And two unsupervised methods, namely DPC \cite{lang2021dpc} and SR-ORNET \cite{deng2023se}, that were trained on human and animal datasets separately. We also compare our method to Diff3F \cite{dutt2024diffusion} that extracts vertex features of each shape and requires no training.    

\paragraph{Evaluation metrics.} \changed{Following Diff3F \cite{dutt2024diffusion}, we use the average matching error and the matching accuracy as our evaluation metrics. For a source shape $\mathcal{X}$ and a target shape $\mathcal{Y}$, represented with a vertex sets $V_{\mathcal{X}} \in \mathbb{R}^{|V_{\mathcal{X}}| \times 3}$ and  $V_{\mathcal{Y}} \in \mathbb{R}^{|V_{\mathcal{Y}}| \times 3}$, respectively, the matching error is 
\begin{equation}
    err = \frac{1}{\vert V_{\mathcal{X}} \vert} \sum_{v \in V_{\mathcal{X}}} \lVert f(v) - y_{gt} \rVert_2, 
\end{equation}
where $f(v)$ is the predicted matching point of $v \in V_{\mathcal{X}}$ in $\mathcal{Y}$ and $y_{gt} \in V_{\mathcal{Y}}$ is the ground truth corresponding point of $v$. Furthermore, the matching accuracy is defined as 
\begin{equation}
    acc(\epsilon) = \frac{1}{\vert V_{\mathcal{X}} \vert}  \sum_{v \in V_{\mathcal{X}}}  \mathbbm{1} ( \lVert f(v) - y_{gt} \rVert_2 < \epsilon d),
\end{equation}
where $\mathbbm{1}( \cdot )$ is the indicator function, $d$ is the maximal Euclidean distance between points in $V_{\mathcal{Y}}$, and $\epsilon \in [0, 1]$ is the error tolerance.}

\paragraph{Experiment results.} The results are summarised in Tab.~\ref{tab:shape_matching}. Our chirality features enhance Diff3F \cite{dutt2024diffusion} performance in shape matching task since it resolves the left/right ambiguity. In Fig.~\ref{fig:quantitive-seg-pck}, we summarise the PCK curves of our method compared to the state-of-the-art method on both human and animal dataset. Fig.~\ref{fig:qualitative_matching} provides qualitative results of our method compared to baselines. 

\subsection{Part segmentation}
\label{sec: segmentation}
We apply $k$-means clustering to our chirality features combined with Diff3F \cite{dutt2024diffusion} features to segment shapes into $k$ parts. Our features enhance Diff3F \cite{dutt2024diffusion} features to be able to differentiate between left and right parts of shapes as seen in Fig.~\ref{fig:part_seg} (left and right legs are not clustered together; the part segmentation is still consistent across shapes). 
\input{tables/part_seg_fig}

\subsection{Partial shapes}
\label{sec:partial_shapes}
To showcase the potential of our method, we conduct a proof-of-concept experiment on partially observed meshes, which is a common real-world scenario for 3D shapes collected by scanning devices. To this end, we extract our chirality features from both human and animal shapes from CP2P dataset \cite{attaiki2021dpfm} and use a model trained on complete shapes from BeCoS \cite{ehm2025becos}. In Fig.~\ref{fig:partiality_fig}, we observe that our chirality features can distinguish between left and right parts of partial shapes even when most of the mesh is missing. 
\input{tables/partiality_fig}

\subsection{Anisotropic shapes}
\label{sec:anisotropic_shapes}
To further demonstrate the robustness of our method to different discretisation of shapes, we show its performance on anisotropic meshes. We use the anisotropic version of FAUST and SCAPE from \cite{donati2020deep}. These non-uniform meshes are often encountered in adaptive refinement and characterised by having a non-consistent discretisation granularity. Qualitative results of the model trained on isotropic BeCoS \cite{ehm2025becos} on anisotropic shapes are shown in Fig.~\ref{fig:anisotropic_shapes}. 

\input{tables/anisotropic_fig}

%% file: tables/left_right_results.tex
\begin{table*}[tbh]
\centering
\resizebox{0.95\textwidth}{!}{%
\setlength{\tabcolsep}{3.5pt}
  \centering
  \begin{tabular}{@{}lccccccccc@{}}
    \toprule
    Train & \textbf{BeCoS} & \multicolumn{2}{c}{\textbf{BeCoS\textsubscript{-h}}} & \multicolumn{2}{c}{\textbf{BeCoS\textsubscript{-a}}} & \multicolumn{2}{c}{\textbf{FAUST}} & \multicolumn{2}{c}{\textbf{SMAL}} \\
    \cmidrule(r){2-2}
    \cmidrule(r){3-4}
    \cmidrule(r){5-6}
    \cmidrule(r){7-8}
    \cmidrule(r){9-10}
    Test & \textbf{BeCoS} & \textbf{BeCoS\textsubscript{-h}} & \textbf{BeCoS\textsubscript{-a}} & \textbf{BeCoS\textsubscript{-h}} & \textbf{BeCoS\textsubscript{-a}} & \textbf{FAUST} & \textbf{SCAPE} & \textbf{SMAL} & \textbf{TOSCA} \\
    \midrule
    Diff3F \cite{dutt2024diffusion} & 50.87 & 54.43 & 50.23& 53.28 & 50.77 & 51.21 & 52.53 & 50.91& 51.48\\
    DINO+SD \cite{oquabdinov2, rombach2022high} & 51.16 & 54.31 & 50.30 & 52.68 & 50.96 & 51.05 & 52.55 & 50.80 & 51.42\\
    \citet{zhang2024telling} & 51.18& 54.16 & 50.78 & 52.83 & 51.02 & 50.90 & 51.47 & 50.41 & 50.97\\
    \citet{liu2012finding} & 79.98 & 79.83 & 80.46 & 79.83 & 80.46 & 90.45 & 80.84 & 75.71 & 72.88 \\
    $\chi_{\text{DINO+SD}}$ & \textbf{91.84} & \textbf{94.09} & \textbf{84.19} & \textbf{90.36} & \textbf{91.10} & \textbf{94.76} & \textbf{95.51} & \textbf{96.59} & \textbf{94.09}\\
    \bottomrule
  \end{tabular}
  }
  \caption{Left and right distinguishment accuracy ($acc_{\chi}\uparrow$) on BeCoS \cite{ehm2025becos} datasets for various methods. Our method outperforms all others across all datasets.}
  \label{tab:chirality_accuracy}
\end{table*}

%% file: tables/qualitative_left_right.tex
\newcommand{\imageheighta}{0.1\textheight}
\newcommand{\imageheightb}{0.09\textheight}
\newcommand{\imageheightc}{0.09\textheight}
\newcommand{\imageheightd}{0.085\textheight}
\newcommand{\imagespacing}{\hspace{0.5cm}}

\begin{figure*}[!ht]
    \centering
    \begin{tabular}{ccccccc}
         & \textbf{GT} &  \textbf{Zhang et al.} \cite{zhang2013symmetry} & \textbf{Liu et al.} \cite{liu2012finding} & \textbf{DINO + SD} \cite{oquabdinov2, rombach2022high} & \textbf{Diff3F} \cite{dutt2024diffusion} & \textbf{Ours} \\

          \rotatebox[origin=c]{90}{\textbf{SCAPE}} &
          \adjustbox{valign=m}{\includegraphics[height=\imageheighta]{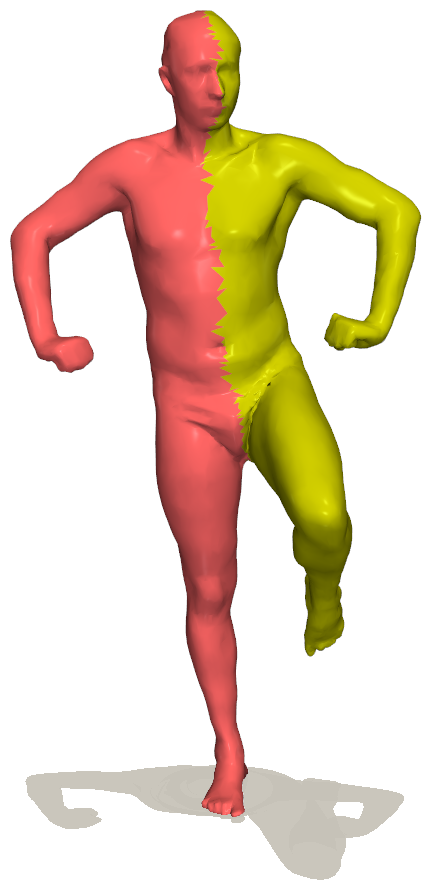}} &
          \imagespacing
         \adjustbox{valign=m}{\includegraphics[height=\imageheighta]{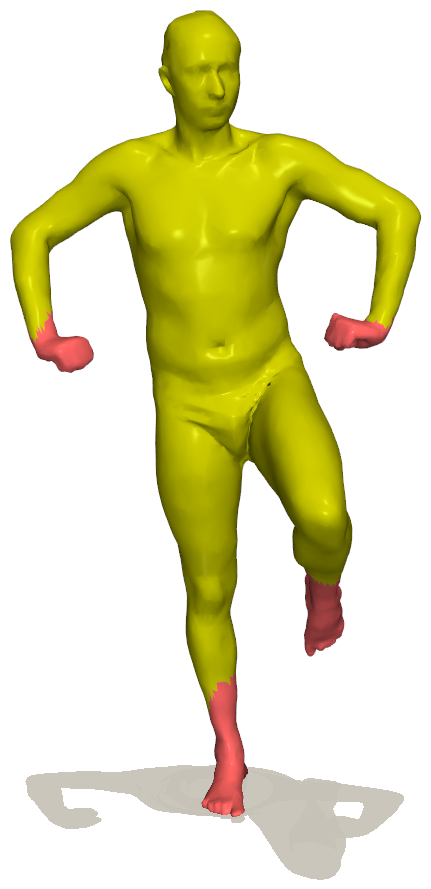}} &
         \imagespacing
         \adjustbox{valign=m}{\includegraphics[height=\imageheighta]{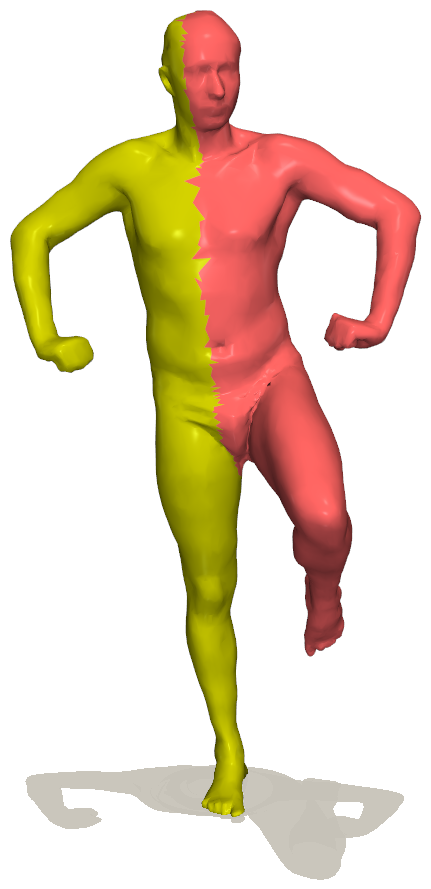}} &
         \imagespacing
         \adjustbox{valign=m}{\includegraphics[height=\imageheighta]{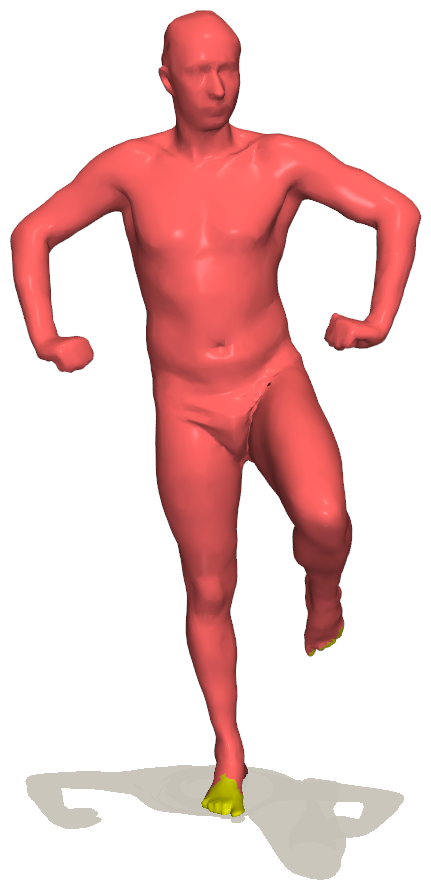}} &
         \imagespacing
         \adjustbox{valign=m}{\includegraphics[height=\imageheighta]{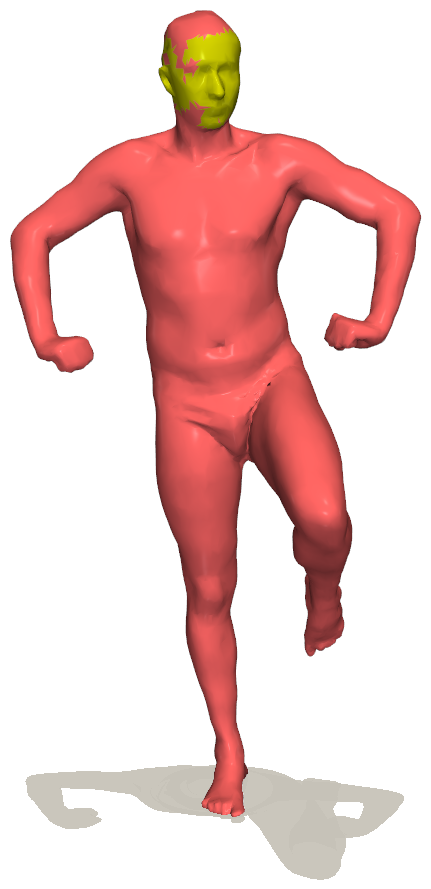}} &
         \imagespacing
         \adjustbox{valign=m}{\includegraphics[height=\imageheighta]{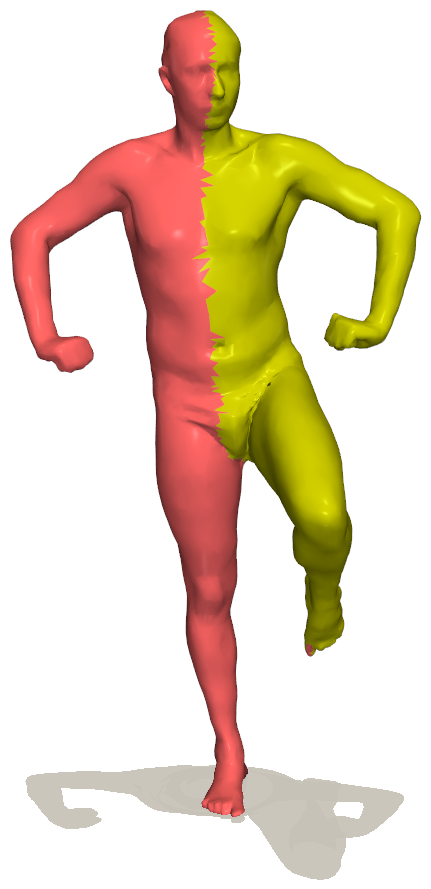}} \\
         
         \rotatebox[origin=c]{90}{\textbf{DT4D}} &
         \imagespacing
         \adjustbox{valign=m}{\includegraphics[height=\imageheightb]{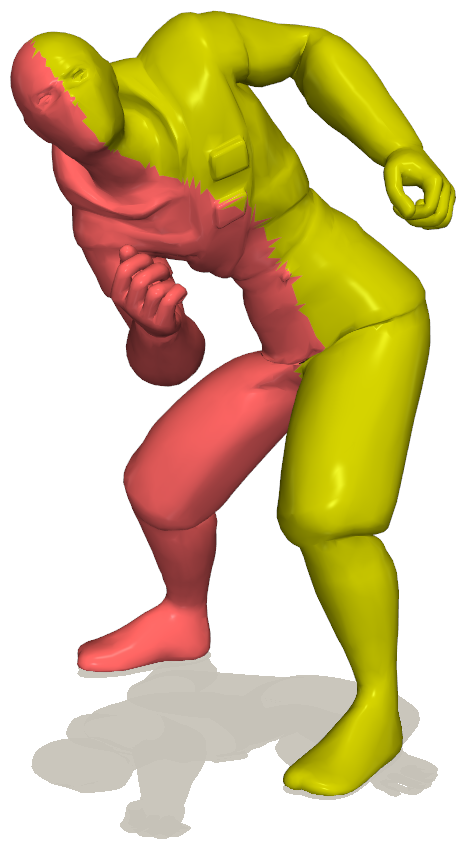}} &
         \imagespacing
         \adjustbox{valign=m}{\includegraphics[height=\imageheightb]{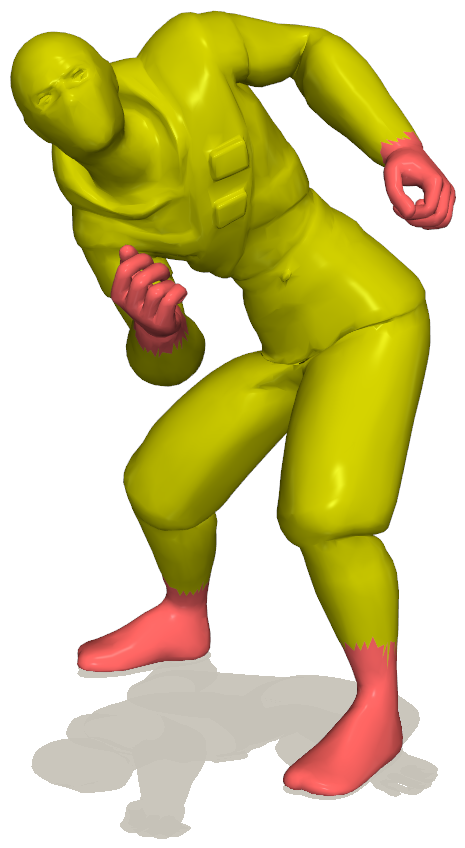}} &
         \imagespacing
         \adjustbox{valign=m}{\includegraphics[height=\imageheightb]{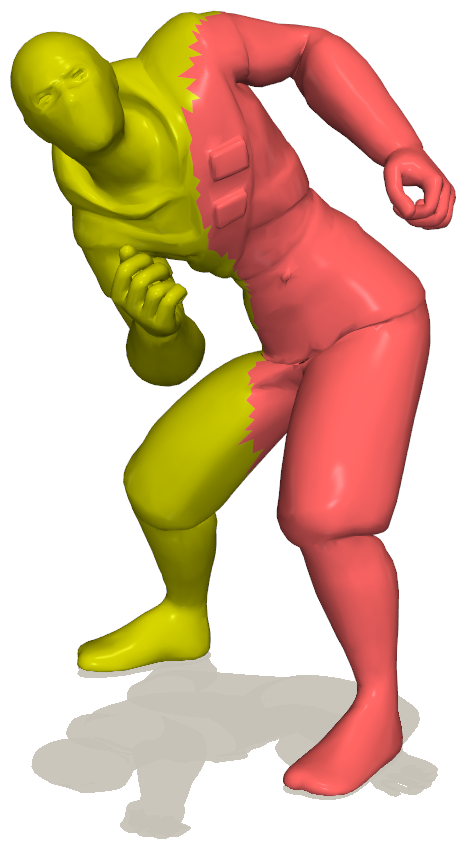}} &
         \imagespacing
         \adjustbox{valign=m}{\includegraphics[height=\imageheightb]{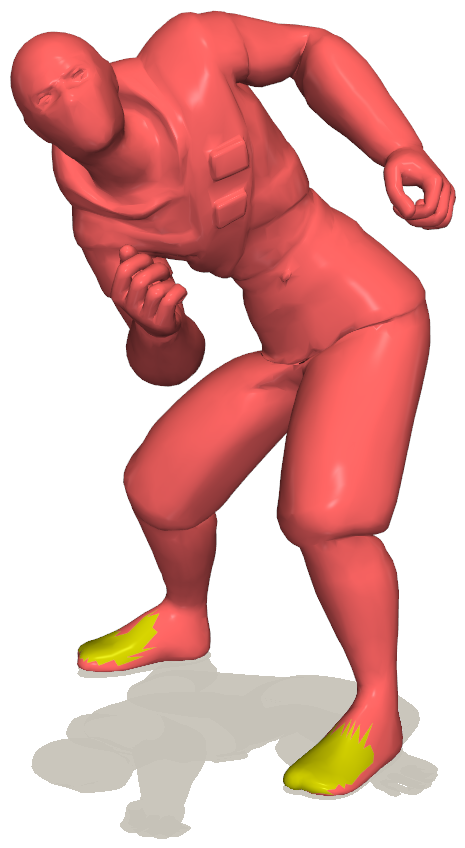}} &
         \imagespacing
         \adjustbox{valign=m}{\includegraphics[height=\imageheightb]{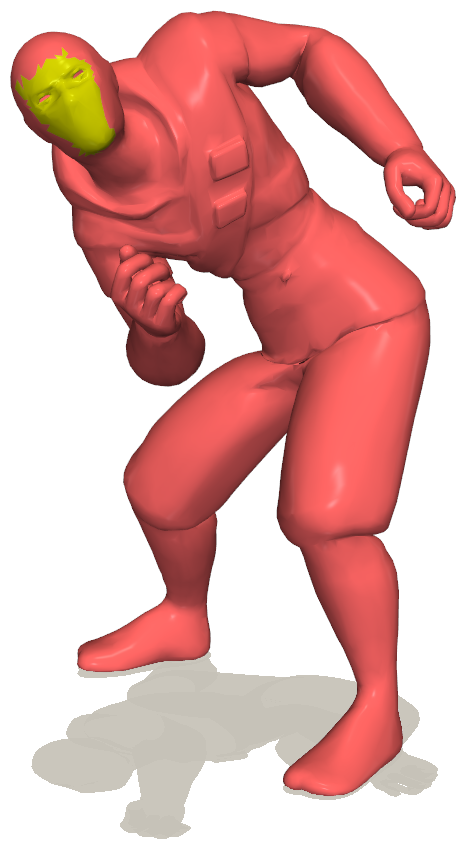}} &
         \imagespacing
         \adjustbox{valign=m}{\includegraphics[height=\imageheightb]{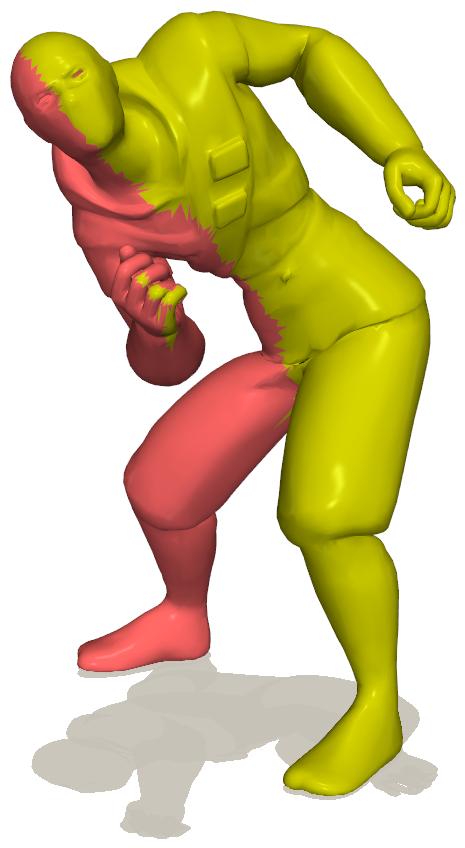}} \\

         \rotatebox[origin=c]{90}{\textbf{DT4D}} &
         \adjustbox{valign=m}{\includegraphics[height=\imageheightc]{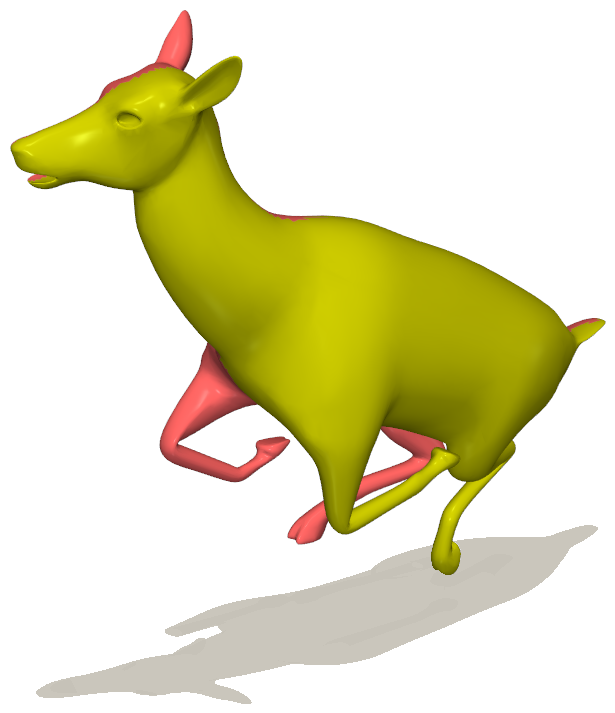}} &
         \imagespacing
         \adjustbox{valign=m}{\includegraphics[height=\imageheightc]{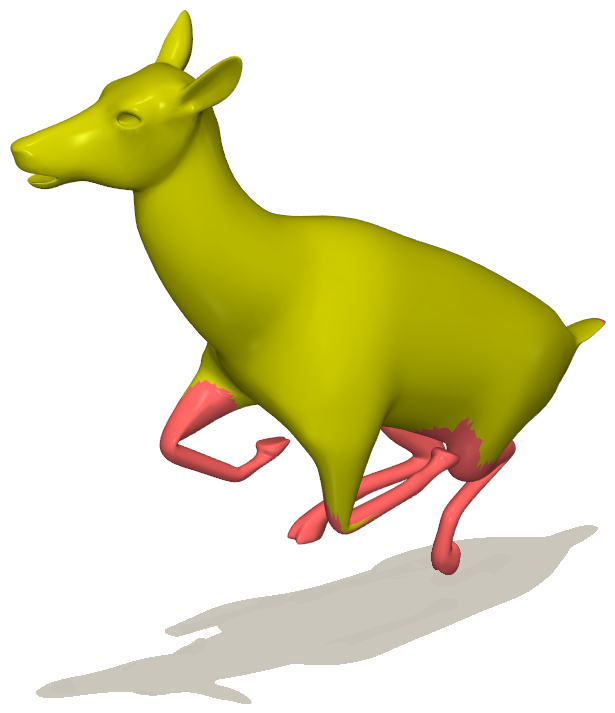}} &
         \imagespacing
         \adjustbox{valign=m}{\includegraphics[height=\imageheightc]{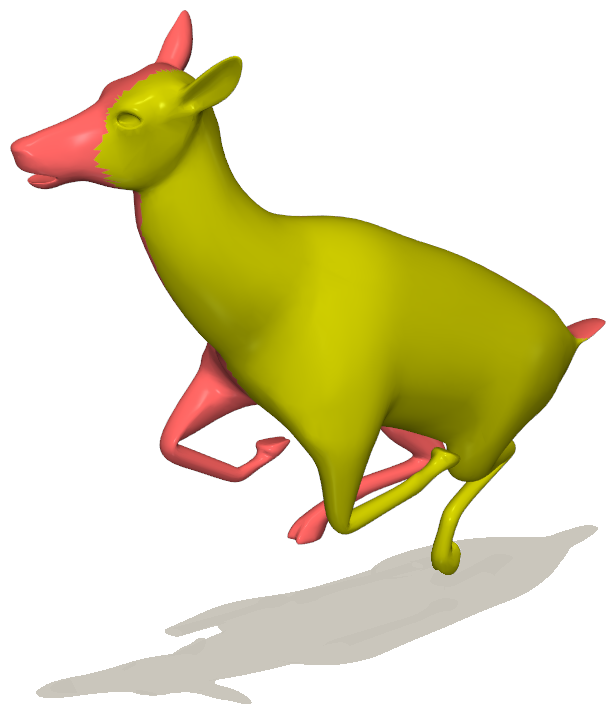}} &
         \imagespacing
         \adjustbox{valign=m}{\includegraphics[height=\imageheightc]{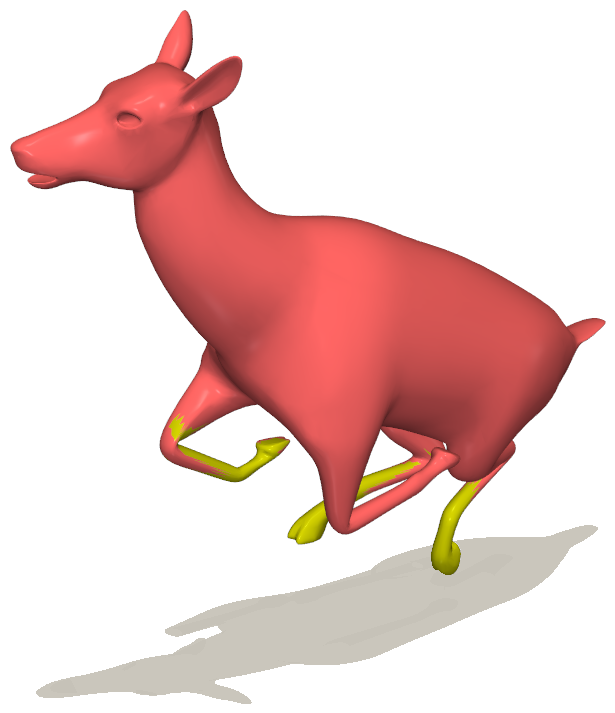}} &
         \imagespacing
         \adjustbox{valign=m}{\includegraphics[height=\imageheightc]{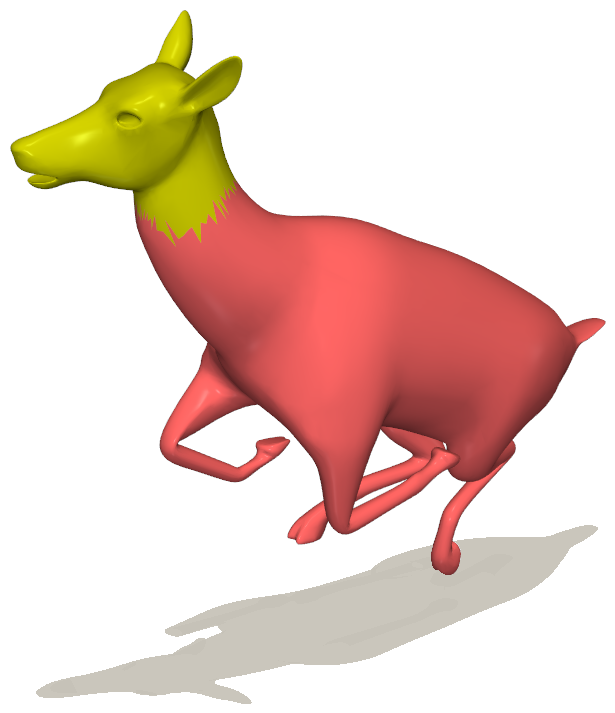}} &
         \imagespacing
         \adjustbox{valign=m}{\includegraphics[height=\imageheightc]{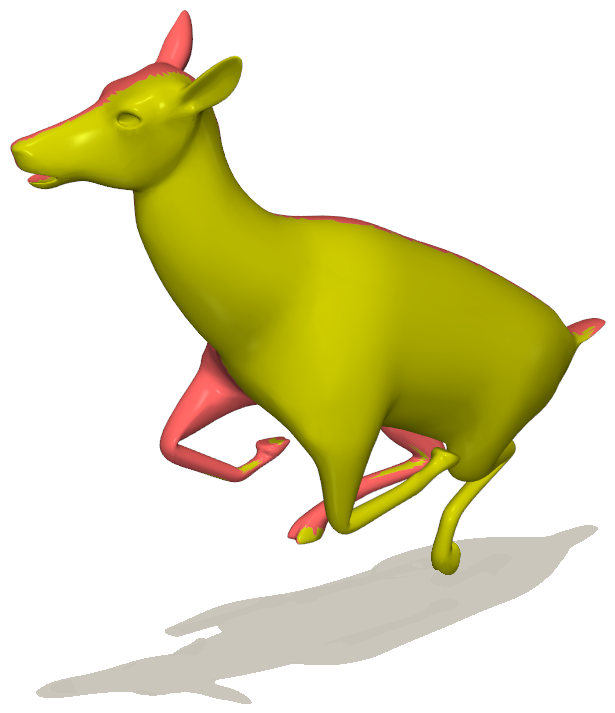}} \\

          \rotatebox[origin=c]{90}{\textbf{TOSCA}} &
          \adjustbox{valign=m}{\includegraphics[height=\imageheightd]{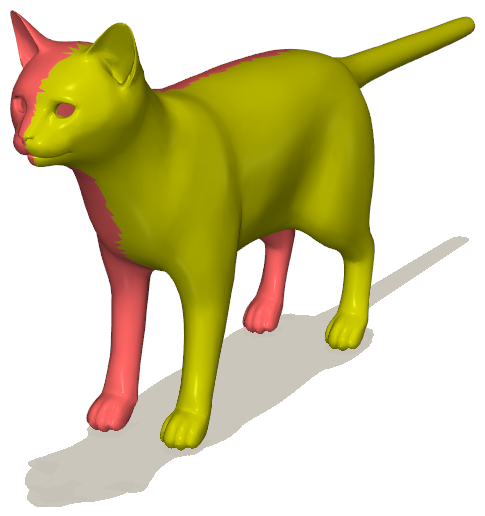}} &
          \imagespacing
         \adjustbox{valign=m}{\includegraphics[height=\imageheightd]{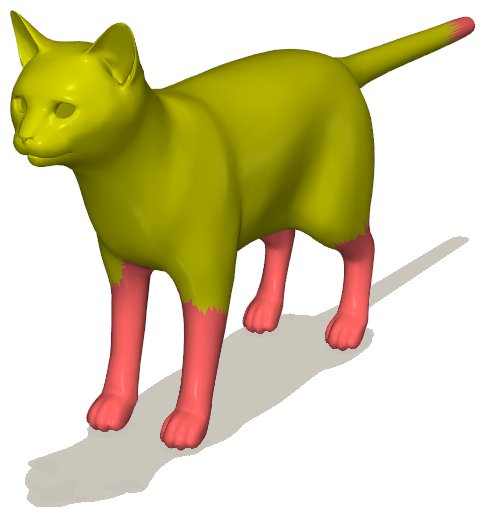}} &
         \imagespacing
         \adjustbox{valign=m}{\includegraphics[height=\imageheightd]{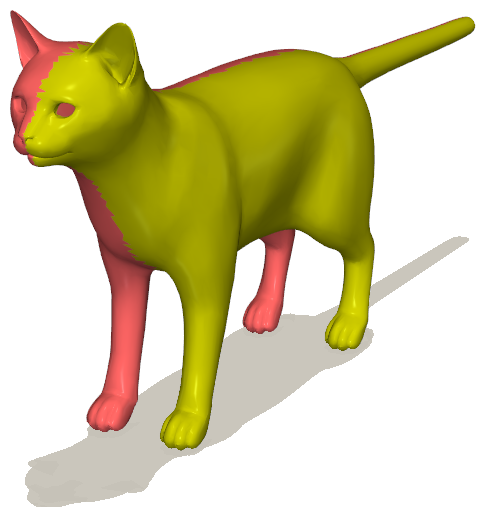}} &
         \imagespacing
         \adjustbox{valign=m}{\includegraphics[height=\imageheightd]{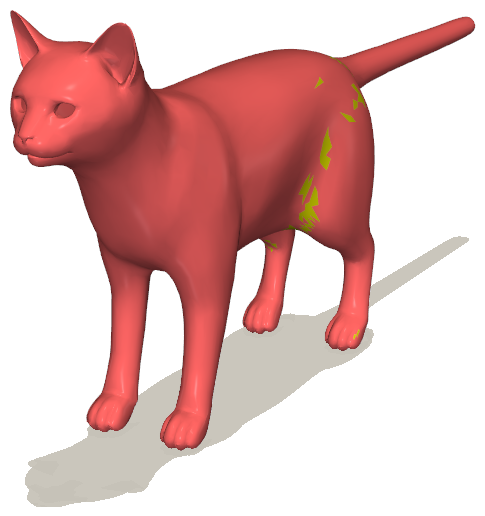}} &
         \imagespacing
         \adjustbox{valign=m}{\includegraphics[height=\imageheightd]{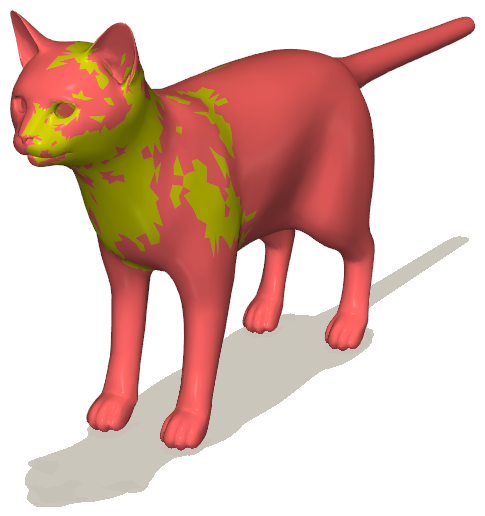}} &
         \imagespacing
         \adjustbox{valign=m}{\includegraphics[height=\imageheightd]{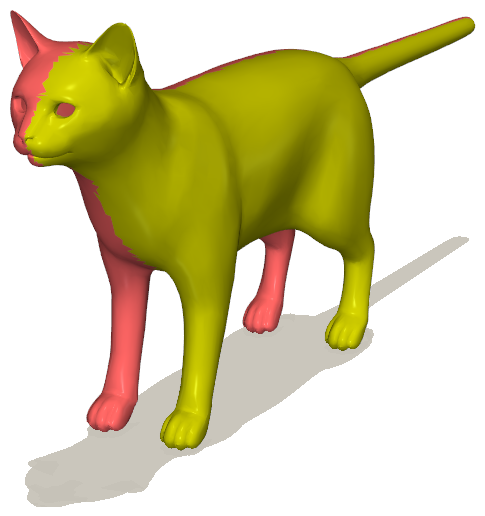}} \\

    \end{tabular}
    
    \caption{Qualitative results on left-right disentanglement of Zhang et al. \cite{zhang2013symmetry}, Liu et al. \cite{liu2012finding}, DINO + SD \cite{oquabdinov2, rombach2022high}, Diff3F \cite{dutt2024diffusion} and our features. Our method provides the only features that consistently and accurately distinguish between left and right of the object.}
    \label{fig:qualitative_left_right}
\end{figure*}

%% file: tables/shape_matching_results.tex
\begin{table}
 \setlength{\tabcolsep}{4.5pt}
  \centering
  \resizebox{0.9\columnwidth}{!}{%
  \begin{tabular}{lcccc}
    \toprule
      & \multicolumn{2}{c}{{\bf \textsc{TOSCA}-a}} & \multicolumn{2}{c}{{\bf \textsc{SHREC}'19}} \\
    \cmidrule(r){2-3}
    \cmidrule(r){4-5}
     & acc ($\uparrow$) & err ($\downarrow$) & acc ($\uparrow$) & err ($\downarrow$)\\
    \midrule
    DPC \cite{lang2021dpc} & 30.79 & 3.74 & 17.40 & 6.26\\
    SE-ORNet \cite{deng2023se} & 33.25 & 4.32 & 21.41 & 4.56\\
    3D-CODED \cite{groueix20183d} & - & - & 2.10 & 8.10 \\
    Diff3F \cite{dutt2024diffusion} & 20.25 & 5.44 & 26.40 & 1.69 \\
    Diff3F \cite{dutt2024diffusion} + $\chi_{\text{DINO+SD}}$ & 22.73 & 4.72 & 27.32 & 1.02 \\
    \bottomrule
  \end{tabular}
  }
  \caption{Shape matching results of \textsc{TOSCA}-a, \textsc{SHREC}'19. (The results of 
  \changed{\cite{lang2021dpc, deng2023se, groueix20183d} }
  are reported from \cite{dutt2024diffusion}).}
  \label{tab:shape_matching}
\end{table}

%% file: tables/pck_curves_fig.tex
\begin{figure}[t]
    
    \resizebox{\columnwidth}{!}{%
    \hspace{-0.9cm}
    \begin{tabular}{cc}
    \setlength{\tabcolsep}{0pt}
        \input{figures/pck_tosca.tikz} &
        \hspace{-1cm}
        \input{figures/pck_shrec19.tikz}
    \end{tabular}
    }
    \vspace{-0.3cm}
    \caption{\textbf{Quantitative comparison} \changed{(with (AUC x 100) shown in brackets)} on \textsc{TOSCA-a} and \textsc{SHREC}'19 datasets. 
    \changed{With our chirality features,}
    the vertex feature descriptor archieves better matching performance on both datasets. }
    \label{fig:quantitive-seg-pck}
\end{figure}
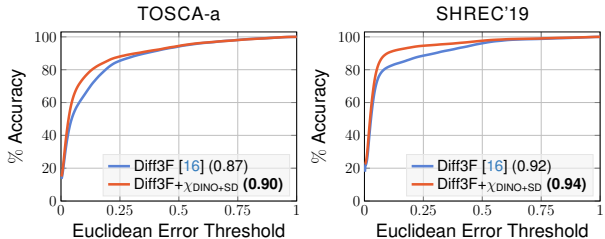

%% file: figures/pck_tosca.tikz
\newcommand{\pckLineWidth}{2pt}
\newcommand{\plotWidth}{0.97\columnwidth}
\newcommand{\plotHeight}{0.75\columnwidth}
\newcommand{\pckTitle}{\textsc{TOSCA}-a}

\pgfplotsset{
    every axis/.style={line width=0.01pt},
    label style = {font=\sffamily\Large},
    tick label style = {font=\sffamily\large},
    title style =  {font=\Large\sffamily},
    legend style={  fill= gray!10,
                    fill opacity=0.6, 
                    font=\sffamily\large,
                    draw=gray!20,
                    text opacity=1}
}
\begin{tikzpicture}[scale=0.5, transform shape]
	\begin{axis}[
		width=\plotWidth,
		height=\plotHeight,
		grid=major,
		title=\pckTitle,
		legend style={
			at={(0.97,0.03)},
			anchor=south east,
			legend columns=1},
		legend cell align={left},
		ylabel={{\sffamily\Large$\%$ Accuracy}},
        xlabel={Euclidean Error Threshold},
		xmin=0,
        xmax=1,
        ylabel near ticks,
        xtick={0, 0.25, 0.5, 0.75, 1},
		ymin=0,
        ymax=103,
        ytick={0, 20, 40, 60, 80, 100},
	]
	
	\addplot [color=mycolor1, smooth, line width=\pckLineWidth]
    table[row sep=crcr]{%
0.002 13.5332441 \\
0.004 14.52209899 \\
0.006 16.09859867 \\
0.008 18.05650404 \\
0.01 20.24898929 \\
0.012 22.55176464 \\
0.014 24.93444056 \\
0.016 27.24848394 \\
0.018 29.49389478 \\
0.02 31.66247815 \\
0.022 33.65999235 \\
0.024 35.60799552 \\
0.026 37.42966018 \\
0.028 39.03450066 \\
0.03 40.65812118 \\
0.032 42.19467056 \\
0.034 43.58234539 \\
0.036 44.91504589 \\
0.038 46.19721099 \\
0.04 47.32913571 \\
0.042 48.40779338 \\
0.044 49.38435588 \\
0.046 50.30355387 \\
0.048 51.19850852 \\
0.05 52.00331895 \\
0.052 52.7972028 \\
0.054 53.53303923 \\
0.056 54.21048678 \\
0.058 54.87769067 \\
0.06 55.5001639 \\
0.062 56.11375929 \\
0.064 56.69662369 \\
0.066 57.26548842 \\
0.068 57.74352601 \\
0.07 58.24580693 \\
0.072 58.72760052 \\
0.074 59.20290647 \\
0.076 59.67240767 \\
0.078 60.13644559 \\
0.08 60.57692308 \\
0.082 61.03071733 \\
0.084 61.48382867 \\
0.086 61.88913625 \\
0.088 62.31219952 \\
0.09 62.72228748 \\
0.092 63.15252131 \\
0.094 63.56807255 \\
0.096 63.98328234 \\
0.098 64.40122378 \\
0.1 64.80892155 \\
0.102 65.24154556 \\
0.104 65.65573099 \\
0.106 66.08562336 \\
0.108 66.51278409 \\
0.11 66.93448153 \\
0.112 67.35583752 \\
0.114 67.77787642 \\
0.116 68.18762292 \\
0.118 68.58644285 \\
0.12 68.99857955 \\
0.122 69.39842384 \\
0.124 69.80748743 \\
0.126 70.21074628 \\
0.128 70.60922476 \\
0.13 70.99097192 \\
0.132 71.36076814 \\
0.134 71.733296 \\
0.136 72.0733856 \\
0.138 72.42064576 \\
0.14 72.75390625 \\
0.142 73.0690696 \\
0.144 73.40198864 \\
0.146 73.74720007 \\
0.148 74.07909473 \\
0.15 74.40279447 \\
0.152 74.72444548 \\
0.154 75.03619427 \\
0.156 75.35784528 \\
0.158 75.67813046 \\
0.16 75.99704983 \\
0.162 76.2852382 \\
0.164 76.56728038 \\
0.166 76.85376147 \\
0.168 77.12658435 \\
0.17 77.41204108 \\
0.172 77.69135162 \\
0.174 77.95836976 \\
0.176 78.21719296 \\
0.178 78.49411331 \\
0.18 78.77035074 \\
0.182 79.04078344 \\
0.184 79.30780157 \\
0.186 79.56969788 \\
0.188 79.83500874 \\
0.19 80.09383195 \\
0.192 80.34855769 \\
0.194 80.60328344 \\
0.196 80.85493608 \\
0.198 81.0922476 \\
0.2 81.35550972 \\
0.202 81.5777972 \\
0.204 81.8075967 \\
0.206 82.0220307 \\
0.208 82.23748907 \\
0.21 82.44167941 \\
0.212 82.64621121 \\
0.214 82.84869427 \\
0.216 83.03376311 \\
0.218 83.2253196 \\
0.22 83.41414445 \\
0.222 83.58555507 \\
0.224 83.76038024 \\
0.226 83.91574246 \\
0.228 84.0632512 \\
0.23 84.22305234 \\
0.232 84.36543925 \\
0.234 84.49928977 \\
0.236 84.61606753 \\
0.238 84.75777153 \\
0.24 84.87181764 \\
0.242 84.99405868 \\
0.244 85.10742188 \\
0.246 85.22317526 \\
0.248 85.34268466 \\
0.25 85.45126748 \\
0.252 85.55814303 \\
0.254 85.65477491 \\
0.256 85.76301628 \\
0.258 85.87296493 \\
0.26 85.96823099 \\
0.262 86.07237489 \\
0.264 86.16627513 \\
0.266 86.26973612 \\
0.268 86.37046547 \\
0.27 86.46436571 \\
0.272 86.55621722 \\
0.274 86.63884943 \\
0.276 86.73274967 \\
0.278 86.83416193 \\
0.28 86.92089161 \\
0.282 87.00932856 \\
0.284 87.09947279 \\
0.286 87.18893411 \\
0.288 87.27224924 \\
0.29 87.36512511 \\
0.292 87.46073263 \\
0.294 87.54643794 \\
0.296 87.63794799 \\
0.298 87.73082386 \\
0.3 87.82369974 \\
0.302 87.91111233 \\
0.304 88.01491477 \\
0.306 88.10130299 \\
0.308 88.18564248 \\
0.31 88.27646962 \\
0.312 88.3601262 \\
0.314 88.43866095 \\
0.316 88.51719569 \\
0.318 88.58855988 \\
0.32 88.66982627 \\
0.322 88.74699519 \\
0.324 88.82757867 \\
0.326 88.90645487 \\
0.328 88.9805507 \\
0.33 89.05089052 \\
0.332 89.12123033 \\
0.334 89.19225306 \\
0.336 89.26498306 \\
0.338 89.33327415 \\
0.34 89.41146744 \\
0.342 89.47668542 \\
0.344 89.54497651 \\
0.346 89.61360905 \\
0.348 89.6829245 \\
0.35 89.75155704 \\
0.352 89.81950667 \\
0.354 89.88165155 \\
0.356 89.94379644 \\
0.358 90.00730715 \\
0.36 90.07286659 \\
0.362 90.13364565 \\
0.364 90.19818073 \\
0.366 90.26647181 \\
0.368 90.33203125 \\
0.37 90.40032233 \\
0.372 90.4641745 \\
0.374 90.51983173 \\
0.376 90.58436681 \\
0.378 90.65060916 \\
0.38 90.71855878 \\
0.382 90.77763057 \\
0.384 90.84489729 \\
0.386 90.91011528 \\
0.388 90.9650896 \\
0.39 91.02928322 \\
0.392 91.0945012 \\
0.394 91.15869482 \\
0.396 91.22630299 \\
0.398 91.28264314 \\
0.4 91.34512948 \\
0.402 91.4001038 \\
0.404 91.46224869 \\
0.406 91.52644231 \\
0.408 91.58585555 \\
0.41 91.64458588 \\
0.412 91.70673077 \\
0.414 91.7808266 \\
0.416 91.85014205 \\
0.418 91.92662806 \\
0.42 91.99048022 \\
0.422 92.05569821 \\
0.424 92.12808676 \\
0.426 92.18579272 \\
0.428 92.25374235 \\
0.43 92.31349705 \\
0.432 92.37325175 \\
0.434 92.4302748 \\
0.436 92.48968805 \\
0.438 92.54944274 \\
0.44 92.61158763 \\
0.442 92.67031796 \\
0.444 92.72426792 \\
0.446 92.77138877 \\
0.448 92.81714379 \\
0.45 92.87177666 \\
0.452 92.92879972 \\
0.454 92.98240822 \\
0.456 93.03806545 \\
0.458 93.09406414 \\
0.46 93.14528245 \\
0.462 93.20367133 \\
0.464 93.24976781 \\
0.466 93.29449847 \\
0.468 93.33479021 \\
0.47 93.38020378 \\
0.472 93.42049552 \\
0.474 93.46317745 \\
0.476 93.5072252 \\
0.478 93.55741914 \\
0.48 93.59771088 \\
0.482 93.64927065 \\
0.484 93.69707441 \\
0.486 93.74760981 \\
0.488 93.8097547 \\
0.49 93.86370465 \\
0.492 93.91560588 \\
0.494 93.98218969 \\
0.496 94.03101781 \\
0.498 94.08974814 \\
0.5 94.14028354 \\
0.502 94.20618444 \\
0.504 94.26389041 \\
0.506 94.31954764 \\
0.508 94.38578999 \\
0.51 94.44042286 \\
0.512 94.48617788 \\
0.514 94.53876202 \\
0.516 94.58656578 \\
0.518 94.63880846 \\
0.52 94.68388057 \\
0.522 94.72724541 \\
0.524 94.77197607 \\
0.526 94.81260927 \\
0.528 94.84948645 \\
0.53 94.89319274 \\
0.532 94.93177721 \\
0.534 94.96831294 \\
0.536 95.01440942 \\
0.538 95.05196951 \\
0.54 95.09226125 \\
0.542 95.13494318 \\
0.544 95.17625929 \\
0.546 95.21416084 \\
0.548 95.25479403 \\
0.55 95.30157343 \\
0.552 95.34186517 \\
0.554 95.38659583 \\
0.556 95.42005846 \\
0.558 95.45864292 \\
0.56 95.49415428 \\
0.562 95.53547039 \\
0.564 95.57268903 \\
0.566 95.61263931 \\
0.568 95.64610194 \\
0.57 95.68502786 \\
0.572 95.72190505 \\
0.574 95.76014806 \\
0.576 95.79702524 \\
0.578 95.83390243 \\
0.58 95.88034036 \\
0.582 95.92541248 \\
0.584 95.96638713 \\
0.586 96.00599596 \\
0.588 96.04901934 \\
0.59 96.08214052 \\
0.592 96.11389587 \\
0.594 96.14360249 \\
0.596 96.1726262 \\
0.598 96.20028409 \\
0.6 96.22760052 \\
0.602 96.24945367 \\
0.604 96.27711156 \\
0.606 96.30033053 \\
0.608 96.32935424 \\
0.61 96.35598776 \\
0.612 96.37886528 \\
0.614 96.40003551 \\
0.616 96.42018138 \\
0.618 96.44749781 \\
0.62 96.46457059 \\
0.622 96.4935943 \\
0.624 96.51749618 \\
0.626 96.54242242 \\
0.628 96.56427557 \\
0.63 96.58442144 \\
0.632 96.60661604 \\
0.634 96.63359102 \\
0.636 96.66193182 \\
0.638 96.68685806 \\
0.64 96.71041849 \\
0.642 96.73671056 \\
0.644 96.76266117 \\
0.646 96.78895323 \\
0.648 96.81865986 \\
0.65 96.85007375 \\
0.652 96.88626803 \\
0.654 96.91495028 \\
0.656 96.94192526 \\
0.658 96.97846099 \\
0.66 97.01431381 \\
0.662 97.04675208 \\
0.664 97.08158053 \\
0.666 97.11401879 \\
0.668 97.14270105 \\
0.67 97.17718805 \\
0.672 97.20996777 \\
0.674 97.23830857 \\
0.676 97.26186899 \\
0.678 97.29055125 \\
0.68 97.31889205 \\
0.682 97.34279392 \\
0.684 97.3697689 \\
0.686 97.39913407 \\
0.688 97.42884069 \\
0.69 97.45205966 \\
0.692 97.47937609 \\
0.694 97.50259506 \\
0.696 97.52752131 \\
0.698 97.54527699 \\
0.7 97.56371558 \\
0.702 97.57908108 \\
0.704 97.5964953 \\
0.706 97.61425098 \\
0.708 97.63268958 \\
0.71 97.64976235 \\
0.712 97.67024967 \\
0.714 97.68834681 \\
0.716 97.70644395 \\
0.718 97.72351672 \\
0.72 97.74161385 \\
0.722 97.76175972 \\
0.724 97.78634451 \\
0.726 97.81058785 \\
0.728 97.83756283 \\
0.73 97.86590363 \\
0.732 97.88980551 \\
0.734 97.9113172 \\
0.736 97.93624344 \\
0.738 97.95980387 \\
0.74 97.97994974 \\
0.742 98.00658326 \\
0.744 98.02843641 \\
0.746 98.05199683 \\
0.748 98.0782889 \\
0.75 98.10150787 \\
0.752 98.12609266 \\
0.754 98.15614073 \\
0.756 98.17696951 \\
0.758 98.19984703 \\
0.76 98.22238309 \\
0.762 98.24662642 \\
0.764 98.26984539 \\
0.766 98.29477163 \\
0.768 98.32174661 \\
0.77 98.34769722 \\
0.772 98.37569657 \\
0.774 98.39857408 \\
0.776 98.42486615 \\
0.778 98.44842657 \\
0.78 98.47745028 \\
0.782 98.49623033 \\
0.784 98.51808348 \\
0.786 98.54027808 \\
0.788 98.56213123 \\
0.79 98.58193564 \\
0.792 98.60481316 \\
0.794 98.62632485 \\
0.796 98.64715363 \\
0.798 98.66217767 \\
0.8 98.67890898 \\
0.802 98.69803049 \\
0.804 98.71339598 \\
0.806 98.73081021 \\
0.808 98.75027316 \\
0.81 98.76734594 \\
0.812 98.7806627 \\
0.814 98.79773547 \\
0.816 98.81514969 \\
0.818 98.82778354 \\
0.82 98.85100251 \\
0.822 98.86739237 \\
0.824 98.88924552 \\
0.826 98.90153792 \\
0.828 98.91758632 \\
0.83 98.93124454 \\
0.832 98.94285402 \\
0.834 98.95821951 \\
0.836 98.96948754 \\
0.838 98.98382867 \\
0.84 98.99748689 \\
0.842 99.01763276 \\
0.844 99.02514478 \\
0.846 99.03675426 \\
0.848 99.04904666 \\
0.85 99.05963177 \\
0.852 99.07943619 \\
0.854 99.09514314 \\
0.856 99.10709408 \\
0.858 99.11972793 \\
0.86 99.13236178 \\
0.862 99.14397126 \\
0.864 99.15967821 \\
0.866 99.17640953 \\
0.868 99.18972629 \\
0.87 99.20167723 \\
0.872 99.21840854 \\
0.874 99.2310424 \\
0.876 99.24470061 \\
0.878 99.25665155 \\
0.88 99.27201705 \\
0.882 99.28601672 \\
0.884 99.30343094 \\
0.886 99.31913789 \\
0.888 99.33552775 \\
0.89 99.34986888 \\
0.892 99.36113691 \\
0.894 99.37240494 \\
0.896 99.39084353 \\
0.898 99.40825776 \\
0.9 99.4263549 \\
0.902 99.44274476 \\
0.904 99.4615248 \\
0.906 99.47825612 \\
0.908 99.49737762 \\
0.91 99.51240166 \\
0.912 99.52674279 \\
0.914 99.5434741 \\
0.916 99.56361997 \\
0.918 99.58240002 \\
0.92 99.60083861 \\
0.922 99.62200885 \\
0.924 99.63669143 \\
0.926 99.65034965 \\
0.928 99.66844679 \\
0.93 99.68995848 \\
0.932 99.70737271 \\
0.934 99.7288844 \\
0.936 99.74390844 \\
0.938 99.76166412 \\
0.94 99.78010271 \\
0.942 99.79785839 \\
0.944 99.81219952 \\
0.946 99.82585774 \\
0.948 99.83780868 \\
0.95 99.85010107 \\
0.952 99.86171056 \\
0.954 99.87468586 \\
0.956 99.88458807 \\
0.958 99.89380736 \\
0.96 99.90063647 \\
0.962 99.90644122 \\
0.964 99.91736779 \\
0.966 99.92351399 \\
0.968 99.92966018 \\
0.97 99.93478201 \\
0.972 99.94126967 \\
0.974 99.94673295 \\
0.976 99.95117188 \\
0.978 99.95492788 \\
0.98 99.95902535 \\
0.982 99.96346427 \\
0.984 99.97097629 \\
0.986 99.97643958 \\
0.988 99.97951267 \\
0.99 99.98668324 \\
0.992 99.99146361 \\
0.994 99.99419526 \\
0.996 99.99658545 \\
0.998 99.99897563 \\
1.0 100.0 \\
    };
    \addlegendentry{\textcolor{black}{Diff3F \cite{dutt2024diffusion} (0.87)}}
    
     \addplot [color=mycolor3, smooth, line width=\pckLineWidth]
           table[row sep=crcr]{%
0.002 14.9349186 \\
0.004 16.06103857 \\
0.006 17.83558239 \\
0.008 20.09396853 \\
0.01 22.73137019 \\
0.012 25.47906195 \\
0.014 28.37221372 \\
0.016 31.20083042 \\
0.018 33.97276552 \\
0.02 36.65762948 \\
0.022 39.13386418 \\
0.024 41.5073208 \\
0.026 43.77697498 \\
0.028 45.91585173 \\
0.03 47.95160894 \\
0.032 49.8207359 \\
0.034 51.65230278 \\
0.036 53.37870138 \\
0.038 55.02554087 \\
0.04 56.51189631 \\
0.042 57.84903573 \\
0.044 59.10866477 \\
0.046 60.28702743 \\
0.048 61.3656851 \\
0.05 62.38800262 \\
0.052 63.31539554 \\
0.054 64.21308184 \\
0.056 65.02233118 \\
0.058 65.78070367 \\
0.06 66.49707714 \\
0.062 67.18647563 \\
0.064 67.82055835 \\
0.066 68.43756829 \\
0.068 68.99789663 \\
0.07 69.53466455 \\
0.072 70.05538407 \\
0.074 70.5330802 \\
0.076 71.00326431 \\
0.078 71.42325448 \\
0.08 71.86031742 \\
0.082 72.2594788 \\
0.084 72.65829873 \\
0.086 73.06872815 \\
0.088 73.45013385 \\
0.09 73.81753988 \\
0.092 74.19416521 \\
0.094 74.55201049 \\
0.096 74.89551464 \\
0.098 75.22809222 \\
0.1 75.56818182 \\
0.102 75.91236888 \\
0.104 76.22172749 \\
0.106 76.53825667 \\
0.108 76.84317635 \\
0.11 77.14809604 \\
0.112 77.44516226 \\
0.114 77.7374481 \\
0.116 78.03451431 \\
0.118 78.31075175 \\
0.12 78.59825721 \\
0.122 78.85912915 \\
0.124 79.12648875 \\
0.126 79.38360468 \\
0.128 79.61750164 \\
0.13 79.85583752 \\
0.132 80.09178322 \\
0.134 80.31338778 \\
0.136 80.52645597 \\
0.138 80.73440232 \\
0.14 80.92698317 \\
0.142 81.12468586 \\
0.144 81.31351071 \\
0.146 81.50745739 \\
0.148 81.69662369 \\
0.15 81.88510708 \\
0.152 82.06198099 \\
0.154 82.22451377 \\
0.156 82.39558293 \\
0.158 82.5697252 \\
0.16 82.72030704 \\
0.162 82.87601071 \\
0.164 83.03786058 \\
0.166 83.17854021 \\
0.168 83.32195149 \\
0.17 83.46160675 \\
0.172 83.61321296 \\
0.174 83.75081949 \\
0.176 83.89286495 \\
0.178 84.02398383 \\
0.18 84.160566 \\
0.182 84.30875765 \\
0.184 84.43987653 \\
0.186 84.58226344 \\
0.188 84.72328453 \\
0.19 84.86293979 \\
0.192 85.00430234 \\
0.194 85.15181108 \\
0.196 85.29112489 \\
0.198 85.42053649 \\
0.2 85.56872815 \\
0.202 85.7107736 \\
0.204 85.85794089 \\
0.206 85.99691324 \\
0.208 86.13486123 \\
0.21 86.25641936 \\
0.212 86.38890406 \\
0.214 86.49953562 \\
0.216 86.60743553 \\
0.218 86.71875 \\
0.22 86.8129917 \\
0.222 86.90859921 \\
0.224 87.00898711 \\
0.226 87.10698481 \\
0.228 87.19917778 \\
0.23 87.28112708 \\
0.232 87.36068619 \\
0.234 87.43580638 \\
0.236 87.51024366 \\
0.238 87.59082714 \\
0.24 87.66799607 \\
0.242 87.74311626 \\
0.244 87.82130955 \\
0.246 87.89642974 \\
0.248 87.96984266 \\
0.25 88.04325557 \\
0.252 88.11257102 \\
0.254 88.17778901 \\
0.256 88.24539718 \\
0.258 88.31334681 \\
0.26 88.37788188 \\
0.262 88.43558785 \\
0.264 88.49226945 \\
0.266 88.54485358 \\
0.268 88.61075448 \\
0.27 88.68075284 \\
0.272 88.75723885 \\
0.274 88.82074956 \\
0.276 88.88699191 \\
0.278 88.94845389 \\
0.28 89.0119646 \\
0.282 89.0717193 \\
0.284 89.12942526 \\
0.286 89.18713123 \\
0.288 89.2356179 \\
0.29 89.28683621 \\
0.292 89.33771307 \\
0.294 89.38756556 \\
0.296 89.44117406 \\
0.298 89.49580693 \\
0.3 89.55282998 \\
0.302 89.60507266 \\
0.304 89.65253497 \\
0.306 89.70989948 \\
0.308 89.75667887 \\
0.31 89.80789718 \\
0.312 89.85706676 \\
0.314 89.90316324 \\
0.316 89.95335719 \\
0.318 89.99399038 \\
0.32 90.03701377 \\
0.322 90.08242734 \\
0.324 90.12647509 \\
0.326 90.16983993 \\
0.328 90.21013167 \\
0.33 90.25554524 \\
0.332 90.30095881 \\
0.334 90.34978693 \\
0.336 90.39144449 \\
0.338 90.44812609 \\
0.34 90.50446624 \\
0.342 90.56592821 \\
0.344 90.62397563 \\
0.346 90.67758413 \\
0.348 90.73768029 \\
0.35 90.79231316 \\
0.352 90.8486533 \\
0.354 90.90123743 \\
0.356 90.95860194 \\
0.358 91.01664937 \\
0.36 91.05967275 \\
0.362 91.10440341 \\
0.364 91.15289008 \\
0.366 91.2047913 \\
0.368 91.25805835 \\
0.37 91.31234976 \\
0.372 91.35605605 \\
0.374 91.40795728 \\
0.376 91.45302939 \\
0.378 91.50083315 \\
0.38 91.54522236 \\
0.382 91.59268466 \\
0.384 91.64834189 \\
0.386 91.69136528 \\
0.388 91.73643739 \\
0.39 91.77809495 \\
0.392 91.81565505 \\
0.394 91.86311735 \\
0.396 91.9071651 \\
0.398 91.94813975 \\
0.4 91.98433403 \\
0.402 92.0236014 \\
0.404 92.06969788 \\
0.406 92.11306272 \\
0.408 92.15437882 \\
0.41 92.19706075 \\
0.412 92.23735249 \\
0.414 92.27696132 \\
0.416 92.32203344 \\
0.418 92.36847137 \\
0.42 92.41934823 \\
0.422 92.46612762 \\
0.424 92.52690669 \\
0.426 92.58495411 \\
0.428 92.64607463 \\
0.43 92.70514642 \\
0.432 92.76694985 \\
0.434 92.82909473 \\
0.436 92.89328835 \\
0.438 92.94382375 \\
0.44 93.00118826 \\
0.442 93.06026005 \\
0.444 93.11933184 \\
0.446 93.17157452 \\
0.448 93.22586593 \\
0.45 93.27913298 \\
0.452 93.33205857 \\
0.454 93.38395979 \\
0.456 93.43995848 \\
0.458 93.50005463 \\
0.46 93.5485413 \\
0.462 93.59395487 \\
0.464 93.64380736 \\
0.466 93.68546493 \\
0.468 93.73019559 \\
0.47 93.77356042 \\
0.472 93.81248634 \\
0.474 93.85311954 \\
0.476 93.891704 \\
0.478 93.92755682 \\
0.48 93.97501912 \\
0.482 94.02077415 \\
0.484 94.07540701 \\
0.486 94.11740603 \\
0.488 94.16828289 \\
0.49 94.21847684 \\
0.492 94.27447552 \\
0.494 94.32398656 \\
0.496 94.37179032 \\
0.498 94.42505736 \\
0.5 94.47490986 \\
0.502 94.53159146 \\
0.504 94.58622432 \\
0.506 94.63710118 \\
0.508 94.68627076 \\
0.51 94.73441597 \\
0.512 94.78426847 \\
0.514 94.83241368 \\
0.516 94.88636364 \\
0.518 94.93246012 \\
0.52 94.97480059 \\
0.522 95.0130436 \\
0.524 95.05811571 \\
0.526 95.10079764 \\
0.528 95.14484539 \\
0.53 95.18138112 \\
0.532 95.22064849 \\
0.534 95.2619646 \\
0.536 95.29918324 \\
0.538 95.33401169 \\
0.54 95.37805944 \\
0.542 95.42039991 \\
0.544 95.4634233 \\
0.546 95.50610522 \\
0.548 95.54571405 \\
0.55 95.58293269 \\
0.552 95.61776115 \\
0.554 95.65907725 \\
0.556 95.69732026 \\
0.558 95.73556326 \\
0.56 95.76458698 \\
0.562 95.79565942 \\
0.564 95.82707332 \\
0.566 95.86907233 \\
0.568 95.90526661 \\
0.57 95.93975361 \\
0.572 95.97014314 \\
0.574 96.00326431 \\
0.576 96.03433676 \\
0.578 96.07121394 \\
0.58 96.09853038 \\
0.582 96.12243226 \\
0.584 96.15077306 \\
0.586 96.17433348 \\
0.588 96.19755245 \\
0.59 96.22452743 \\
0.592 96.24877076 \\
0.594 96.27472137 \\
0.596 96.30203781 \\
0.598 96.32798842 \\
0.6 96.35974377 \\
0.602 96.38945039 \\
0.604 96.40993772 \\
0.606 96.42974213 \\
0.608 96.45159528 \\
0.61 96.47925317 \\
0.612 96.49598448 \\
0.614 96.51988636 \\
0.616 96.5441297 \\
0.618 96.56734867 \\
0.62 96.59364073 \\
0.622 96.61890844 \\
0.624 96.64076158 \\
0.626 96.66534637 \\
0.628 96.68788243 \\
0.63 96.70939412 \\
0.632 96.73705201 \\
0.634 96.7616368 \\
0.636 96.78724596 \\
0.638 96.80807474 \\
0.64 96.82890352 \\
0.642 96.85519559 \\
0.644 96.87978038 \\
0.646 96.89890188 \\
0.648 96.9224623 \\
0.65 96.94124235 \\
0.652 96.96958315 \\
0.654 96.99621667 \\
0.656 97.0228502 \\
0.658 97.05153245 \\
0.66 97.07748306 \\
0.662 97.10548241 \\
0.664 97.13245739 \\
0.666 97.16182255 \\
0.668 97.18743171 \\
0.67 97.21406523 \\
0.672 97.23865002 \\
0.674 97.25913735 \\
0.676 97.28133195 \\
0.678 97.30147782 \\
0.68 97.32264806 \\
0.682 97.34381829 \\
0.684 97.36396416 \\
0.686 97.38001257 \\
0.688 97.40015844 \\
0.69 97.42064576 \\
0.692 97.43601125 \\
0.694 97.4517182 \\
0.696 97.46879097 \\
0.698 97.48791248 \\
0.7 97.51215581 \\
0.702 97.53400896 \\
0.704 97.549033 \\
0.706 97.56917887 \\
0.708 97.58830037 \\
0.71 97.60332441 \\
0.712 97.62142155 \\
0.714 97.64054305 \\
0.716 97.66000601 \\
0.718 97.67878606 \\
0.72 97.70098066 \\
0.722 97.71805343 \\
0.724 97.73580911 \\
0.726 97.75458916 \\
0.728 97.77644231 \\
0.73 97.8010271 \\
0.732 97.83346536 \\
0.734 97.86044034 \\
0.736 97.89048842 \\
0.738 97.91541466 \\
0.74 97.93897509 \\
0.742 97.96458424 \\
0.744 97.99599814 \\
0.746 98.02604622 \\
0.748 98.05131392 \\
0.75 98.07419143 \\
0.752 98.1004835 \\
0.754 98.12165374 \\
0.756 98.14453125 \\
0.758 98.16365275 \\
0.76 98.1824328 \\
0.762 98.20394449 \\
0.764 98.22477327 \\
0.766 98.24389478 \\
0.768 98.2674552 \\
0.77 98.28794253 \\
0.772 98.31252732 \\
0.774 98.3333561 \\
0.776 98.35657507 \\
0.778 98.38115986 \\
0.78 98.40130573 \\
0.782 98.41974432 \\
0.784 98.43681709 \\
0.786 98.4596946 \\
0.788 98.4818892 \\
0.79 98.50203507 \\
0.792 98.52286385 \\
0.794 98.53959517 \\
0.796 98.55769231 \\
0.798 98.57203344 \\
0.8 98.58910621 \\
0.802 98.6075448 \\
0.804 98.61983719 \\
0.806 98.63793433 \\
0.808 98.64988527 \\
0.81 98.66661659 \\
0.812 98.68539663 \\
0.814 98.70246941 \\
0.816 98.71817635 \\
0.818 98.73627349 \\
0.82 98.7523219 \\
0.822 98.76939467 \\
0.824 98.78885763 \\
0.826 98.80558894 \\
0.828 98.82471045 \\
0.83 98.84588068 \\
0.832 98.86090472 \\
0.834 98.87934331 \\
0.836 98.89675754 \\
0.838 98.91826923 \\
0.84 98.92782998 \\
0.842 98.94319548 \\
0.844 98.95309768 \\
0.846 98.96573153 \\
0.848 98.97563374 \\
0.85 98.98724323 \\
0.852 98.9981698 \\
0.854 99.0073891 \\
0.856 99.02173022 \\
0.858 99.03812008 \\
0.86 99.05143684 \\
0.862 99.06748525 \\
0.864 99.08080201 \\
0.866 99.10060642 \\
0.868 99.12075229 \\
0.87 99.14089816 \\
0.872 99.16138549 \\
0.874 99.17914117 \\
0.876 99.19928704 \\
0.878 99.2218231 \\
0.88 99.24470061 \\
0.882 99.26860249 \\
0.884 99.29216292 \\
0.886 99.31469897 \\
0.888 99.33450339 \\
0.89 99.35464926 \\
0.892 99.37069766 \\
0.894 99.38845334 \\
0.896 99.40825776 \\
0.898 99.42669635 \\
0.9 99.44342767 \\
0.902 99.46357354 \\
0.904 99.47757321 \\
0.906 99.49806053 \\
0.908 99.51649913 \\
0.91 99.5335719 \\
0.912 99.54996176 \\
0.914 99.56635162 \\
0.916 99.58171711 \\
0.918 99.59913134 \\
0.92 99.6124481 \\
0.922 99.62986233 \\
0.924 99.6469351 \\
0.926 99.66844679 \\
0.928 99.68927557 \\
0.93 99.70668979 \\
0.932 99.72717712 \\
0.934 99.74732299 \\
0.936 99.76883468 \\
0.938 99.78078562 \\
0.94 99.79444384 \\
0.942 99.80810205 \\
0.944 99.81937008 \\
0.946 99.83268684 \\
0.948 99.84566215 \\
0.95 99.85761309 \\
0.952 99.86853966 \\
0.954 99.88151497 \\
0.956 99.89141718 \\
0.958 99.90507539 \\
0.96 99.91361178 \\
0.962 99.92522126 \\
0.964 99.93614784 \\
0.966 99.94161112 \\
0.968 99.94912314 \\
0.97 99.95663516 \\
0.972 99.96278136 \\
0.974 99.9685861 \\
0.976 99.97404939 \\
0.978 99.97712249 \\
0.98 99.98258577 \\
0.982 99.98668324 \\
0.984 99.99009779 \\
0.986 99.99282944 \\
0.988 99.99487817 \\
0.99 99.99624399 \\
0.992 99.99760981 \\
0.994 99.99897563 \\
0.996 99.99897563 \\
0.998 99.99965854 \\
1.0 100.0 \\
         };
         \addlegendentry{\textcolor{black}{Diff3F+$\chi_{\text{DINO+SD}}$ \textbf{(0.90)}}}

	\end{axis}
\end{tikzpicture}

%% file: figures/pck_shrec19.tikz
\newcommand{\pckLineWidth}{2pt}
\newcommand{\plotWidth}{0.97\columnwidth}
\newcommand{\plotHeight}{0.75\columnwidth}
\newcommand{\pckTitle}{\textsc{SHREC}'19}

\pgfplotsset{
    every axis/.style={line width=0.01pt},
    label style = {font=\sffamily\Large},
    tick label style = {font=\sffamily\large},
    title style =  {font=\Large\sffamily},
    legend style={  fill= gray!10,
                    fill opacity=0.6, 
                    font=\sffamily\large,
                    draw=gray!20,
                    text opacity=1}
}
\begin{tikzpicture}[scale=0.5, transform shape]
	\begin{axis}[
		width=\plotWidth,
		height=\plotHeight,
		grid=major,
		title=\pckTitle,
		legend style={
			at={(0.97,0.03)},
			anchor=south east,
			legend columns=1},
		legend cell align={left},
		ylabel={{\sffamily\Large$\%$ Accuracy}},
        xlabel={Euclidean Error Threshold},
		xmin=0,
        xmax=1,
        ylabel near ticks,
        xtick={0, 0.25, 0.5, 0.75, 1},
		ymin=0,
        ymax=103,
        ytick={0, 20, 40, 60, 80, 100},
	]
	
	\addplot [color=mycolor1, smooth, line width=\pckLineWidth]
    table[row sep=crcr]{%
0.002 17.81295422 \\
0.004 21.82753452 \\
0.006 22.11800509 \\
0.008 22.98737282 \\
0.01 24.37908794 \\
0.012 26.39671148 \\
0.014 28.94394985 \\
0.016 31.97651708 \\
0.018 35.29319586 \\
0.02 38.54196948 \\
0.022 41.69263263 \\
0.024 44.75495094 \\
0.026 47.69122456 \\
0.028 50.47261083 \\
0.03 53.09252362 \\
0.032 55.59774709 \\
0.034 57.95807594 \\
0.036 60.11014717 \\
0.038 62.13004179 \\
0.04 64.06045603 \\
0.042 65.79987282 \\
0.044 67.35737645 \\
0.046 68.78270349 \\
0.048 70.0690407 \\
0.05 71.24432231 \\
0.052 72.27698038 \\
0.054 73.21061955 \\
0.056 74.05386991 \\
0.058 74.81604288 \\
0.06 75.53075036 \\
0.062 76.12895167 \\
0.064 76.67173874 \\
0.066 77.17410065 \\
0.068 77.63376635 \\
0.07 78.04460392 \\
0.072 78.43500182 \\
0.074 78.78883539 \\
0.076 79.09474927 \\
0.078 79.39339571 \\
0.08 79.68409339 \\
0.082 79.94186047 \\
0.084 80.17623547 \\
0.086 80.40493278 \\
0.088 80.60637718 \\
0.09 80.80327943 \\
0.092 80.98155887 \\
0.094 81.14871003 \\
0.096 81.30813953 \\
0.098 81.45507813 \\
0.1 81.60247093 \\
0.102 81.74373183 \\
0.104 81.87590843 \\
0.106 82.00763081 \\
0.108 82.12913336 \\
0.11 82.26244549 \\
0.112 82.37849746 \\
0.114 82.49227834 \\
0.116 82.60651344 \\
0.118 82.71961301 \\
0.12 82.84293241 \\
0.122 82.95285247 \\
0.124 83.06527071 \\
0.126 83.17405523 \\
0.128 83.27193859 \\
0.13 83.37368278 \\
0.132 83.46566134 \\
0.134 83.55695858 \\
0.136 83.64416788 \\
0.138 83.73024164 \\
0.14 83.81200036 \\
0.142 83.89534884 \\
0.144 83.98323946 \\
0.146 84.06590661 \\
0.148 84.14403161 \\
0.15 84.23396621 \\
0.152 84.32549055 \\
0.154 84.4140625 \\
0.156 84.50626817 \\
0.158 84.59211483 \\
0.16 84.67455487 \\
0.162 84.76403525 \\
0.164 84.86123728 \\
0.166 84.95162609 \\
0.168 85.04474019 \\
0.17 85.13331214 \\
0.172 85.22733467 \\
0.174 85.34225109 \\
0.176 85.43468387 \\
0.178 85.53574673 \\
0.18 85.65225291 \\
0.182 85.75399709 \\
0.184 85.84052507 \\
0.186 85.93295785 \\
0.188 86.01971294 \\
0.19 86.12304688 \\
0.192 86.2229742 \\
0.194 86.31586119 \\
0.196 86.41601563 \\
0.198 86.51685138 \\
0.2 86.62200218 \\
0.202 86.71875 \\
0.204 86.82117551 \\
0.206 86.91565225 \\
0.208 87.01103743 \\
0.21 87.10256177 \\
0.212 87.19136083 \\
0.214 87.28470203 \\
0.216 87.37350109 \\
0.218 87.4548056 \\
0.22 87.54087936 \\
0.222 87.61991279 \\
0.224 87.69485828 \\
0.226 87.77071221 \\
0.228 87.84906432 \\
0.23 87.92014898 \\
0.232 87.98396621 \\
0.234 88.05073583 \\
0.236 88.11659702 \\
0.238 88.18359375 \\
0.24 88.24900073 \\
0.242 88.30873001 \\
0.244 88.36936773 \\
0.246 88.43500182 \\
0.248 88.49404978 \\
0.25 88.55196221 \\
0.252 88.61078307 \\
0.254 88.67005814 \\
0.256 88.72229288 \\
0.258 88.77566315 \\
0.26 88.83312137 \\
0.262 88.89580305 \\
0.264 88.9543968 \\
0.266 89.01389898 \\
0.268 89.07249273 \\
0.27 89.13108648 \\
0.272 89.18718205 \\
0.274 89.24146076 \\
0.276 89.29710211 \\
0.278 89.36273619 \\
0.28 89.42360102 \\
0.282 89.48764535 \\
0.284 89.54578488 \\
0.286 89.609375 \\
0.288 89.67023983 \\
0.29 89.73791788 \\
0.292 89.79424055 \\
0.294 89.8587391 \\
0.296 89.921875 \\
0.298 89.98591933 \\
0.3 90.05427871 \\
0.302 90.11764172 \\
0.304 90.19258721 \\
0.306 90.25458757 \\
0.308 90.31931323 \\
0.31 90.38994368 \\
0.312 90.46080124 \\
0.314 90.53165879 \\
0.316 90.61228198 \\
0.318 90.69017987 \\
0.32 90.76875908 \\
0.322 90.83757267 \\
0.324 90.91229106 \\
0.326 90.98065044 \\
0.328 91.04492188 \\
0.33 91.10601381 \\
0.332 91.16188227 \\
0.334 91.21729651 \\
0.336 91.27498183 \\
0.338 91.33970749 \\
0.34 91.39466751 \\
0.342 91.44894622 \\
0.344 91.49800145 \\
0.346 91.54978198 \\
0.348 91.60474201 \\
0.35 91.65674964 \\
0.352 91.71352653 \\
0.354 91.76144622 \\
0.356 91.81413517 \\
0.358 91.8688681 \\
0.36 91.92291969 \\
0.362 91.97492733 \\
0.364 92.02875182 \\
0.366 92.09279615 \\
0.368 92.15116279 \\
0.37 92.20952943 \\
0.372 92.265625 \\
0.374 92.32376453 \\
0.376 92.37736192 \\
0.378 92.4400436 \\
0.38 92.50181686 \\
0.382 92.56472565 \\
0.384 92.6383085 \\
0.386 92.71075581 \\
0.388 92.77252907 \\
0.39 92.84202398 \\
0.392 92.91901344 \\
0.394 92.97579033 \\
0.396 93.04028888 \\
0.398 93.11023801 \\
0.4 93.16792333 \\
0.402 93.22651708 \\
0.404 93.28806323 \\
0.406 93.34552144 \\
0.408 93.3984375 \\
0.41 93.45089935 \\
0.412 93.49586664 \\
0.414 93.54810138 \\
0.416 93.59874637 \\
0.418 93.65120821 \\
0.42 93.70662246 \\
0.422 93.75863009 \\
0.424 93.81086483 \\
0.426 93.8701399 \\
0.428 93.92941497 \\
0.43 93.98551054 \\
0.432 94.04387718 \\
0.434 94.09906432 \\
0.436 94.15652253 \\
0.438 94.22442769 \\
0.44 94.28869913 \\
0.442 94.35887536 \\
0.444 94.42632631 \\
0.446 94.4951399 \\
0.448 94.55714026 \\
0.45 94.62118459 \\
0.452 94.68000545 \\
0.454 94.74972747 \\
0.456 94.82149346 \\
0.458 94.88962573 \\
0.46 94.96275436 \\
0.462 95.02384629 \\
0.464 95.09334121 \\
0.466 95.16011083 \\
0.468 95.22097565 \\
0.47 95.28683685 \\
0.472 95.34770167 \\
0.474 95.40743096 \\
0.476 95.46398074 \\
0.478 95.51939499 \\
0.48 95.57435501 \\
0.482 95.62568132 \\
0.484 95.67496366 \\
0.486 95.73287609 \\
0.488 95.78374818 \\
0.49 95.83303052 \\
0.492 95.88821766 \\
0.494 95.9436319 \\
0.496 95.98928052 \\
0.498 96.03833576 \\
0.5 96.08898074 \\
0.502 96.13962573 \\
0.504 96.20117188 \\
0.506 96.25545058 \\
0.508 96.30200763 \\
0.51 96.36196403 \\
0.512 96.40216206 \\
0.514 96.45235283 \\
0.516 96.50345203 \\
0.518 96.55750363 \\
0.52 96.61200945 \\
0.522 96.66061047 \\
0.524 96.70557776 \\
0.526 96.75554142 \\
0.528 96.80096294 \\
0.53 96.84479469 \\
0.532 96.88885356 \\
0.534 96.93313953 \\
0.536 96.97492733 \\
0.538 97.01557958 \\
0.54 97.05600472 \\
0.542 97.08961664 \\
0.544 97.13889898 \\
0.546 97.18000545 \\
0.548 97.21656977 \\
0.55 97.24700218 \\
0.552 97.28265807 \\
0.554 97.31490734 \\
0.556 97.34579397 \\
0.558 97.37736192 \\
0.56 97.40052689 \\
0.562 97.42573583 \\
0.564 97.45253452 \\
0.566 97.47978743 \\
0.568 97.50454215 \\
0.57 97.52861555 \\
0.572 97.55427871 \\
0.574 97.57948765 \\
0.576 97.60333394 \\
0.578 97.62899709 \\
0.58 97.65806686 \\
0.582 97.68531977 \\
0.584 97.71257267 \\
0.586 97.74550327 \\
0.588 97.77707122 \\
0.59 97.81159157 \\
0.592 97.84883721 \\
0.594 97.87472747 \\
0.596 97.90629542 \\
0.598 97.9387718 \\
0.6 97.96352653 \\
0.602 97.99305051 \\
0.604 98.02325581 \\
0.606 98.04846475 \\
0.608 98.07208394 \\
0.61 98.0979742 \\
0.612 98.11864099 \\
0.614 98.1340843 \\
0.616 98.14861919 \\
0.618 98.16769622 \\
0.62 98.18541061 \\
0.622 98.20267078 \\
0.624 98.21947674 \\
0.626 98.23151344 \\
0.628 98.24377725 \\
0.63 98.25922057 \\
0.632 98.2751181 \\
0.634 98.28692769 \\
0.636 98.30032703 \\
0.638 98.31622456 \\
0.64 98.32667151 \\
0.642 98.34007086 \\
0.644 98.35097202 \\
0.646 98.36300872 \\
0.648 98.37459121 \\
0.65 98.38367551 \\
0.652 98.39344113 \\
0.654 98.40547783 \\
0.656 98.42023983 \\
0.658 98.43432049 \\
0.66 98.44431323 \\
0.662 98.45407885 \\
0.664 98.46384448 \\
0.666 98.4736101 \\
0.668 98.48655523 \\
0.67 98.49404978 \\
0.672 98.50608648 \\
0.674 98.51358103 \\
0.676 98.52243823 \\
0.678 98.53197674 \\
0.68 98.5396984 \\
0.682 98.54696584 \\
0.684 98.55355196 \\
0.686 98.56240916 \\
0.688 98.56899528 \\
0.69 98.5755814 \\
0.692 98.58103198 \\
0.694 98.58602834 \\
0.696 98.59147892 \\
0.698 98.59806504 \\
0.7 98.60351563 \\
0.702 98.61010174 \\
0.704 98.61600654 \\
0.706 98.62304688 \\
0.708 98.62872456 \\
0.71 98.63394804 \\
0.712 98.63939862 \\
0.714 98.64598474 \\
0.716 98.65279797 \\
0.718 98.65870276 \\
0.72 98.66506177 \\
0.722 98.67210211 \\
0.724 98.67868823 \\
0.726 98.68459302 \\
0.728 98.69140625 \\
0.73 98.69731105 \\
0.732 98.70276163 \\
0.734 98.70889353 \\
0.736 98.71979469 \\
0.738 98.72683503 \\
0.74 98.73410247 \\
0.742 98.74205124 \\
0.744 98.75136265 \\
0.746 98.76339935 \\
0.748 98.77589026 \\
0.75 98.78792696 \\
0.752 98.7951944 \\
0.754 98.80382449 \\
0.756 98.81268169 \\
0.758 98.82358285 \\
0.76 98.83175872 \\
0.762 98.84243278 \\
0.764 98.85310683 \\
0.766 98.86287246 \\
0.768 98.87468205 \\
0.77 98.88467478 \\
0.772 98.89557594 \\
0.774 98.90488735 \\
0.776 98.91624273 \\
0.778 98.92623547 \\
0.78 98.93600109 \\
0.782 98.94690225 \\
0.784 98.96052871 \\
0.786 98.96915879 \\
0.788 98.97960574 \\
0.79 98.9905069 \\
0.792 99.00186228 \\
0.794 99.01367188 \\
0.796 99.02321039 \\
0.798 99.03479288 \\
0.8 99.04342297 \\
0.802 99.05841206 \\
0.804 99.06999455 \\
0.806 99.08134993 \\
0.808 99.09088844 \\
0.81 99.10292515 \\
0.812 99.11269077 \\
0.814 99.1226835 \\
0.816 99.1344931 \\
0.818 99.1435774 \\
0.82 99.15493278 \\
0.822 99.16401708 \\
0.824 99.17469113 \\
0.826 99.18150436 \\
0.828 99.19081577 \\
0.83 99.19831032 \\
0.832 99.20148983 \\
0.834 99.20898438 \\
0.836 99.21466206 \\
0.838 99.21988554 \\
0.84 99.22806141 \\
0.842 99.23646439 \\
0.844 99.24554869 \\
0.846 99.25145349 \\
0.848 99.25894804 \\
0.85 99.26666969 \\
0.852 99.27802507 \\
0.854 99.28211301 \\
0.856 99.28915334 \\
0.858 99.29528525 \\
0.86 99.30391533 \\
0.862 99.31209121 \\
0.864 99.32162972 \\
0.866 99.33275799 \\
0.868 99.34434048 \\
0.87 99.35501453 \\
0.872 99.36228198 \\
0.874 99.37363735 \\
0.876 99.3840843 \\
0.878 99.39725654 \\
0.88 99.40929324 \\
0.882 99.42064862 \\
0.884 99.43109557 \\
0.886 99.44245094 \\
0.888 99.45562318 \\
0.89 99.4651617 \\
0.892 99.47697129 \\
0.894 99.48741824 \\
0.896 99.49900073 \\
0.898 99.510129 \\
0.9 99.52307413 \\
0.902 99.53442951 \\
0.904 99.54419513 \\
0.906 99.56145531 \\
0.908 99.5732649 \\
0.91 99.58303052 \\
0.912 99.59506722 \\
0.914 99.60801235 \\
0.916 99.61891352 \\
0.918 99.62845203 \\
0.92 99.64094295 \\
0.922 99.65275254 \\
0.924 99.66183685 \\
0.926 99.67387355 \\
0.928 99.68386628 \\
0.93 99.69544876 \\
0.932 99.70816679 \\
0.934 99.71906795 \\
0.936 99.7326944 \\
0.938 99.74404978 \\
0.94 99.75744913 \\
0.942 99.76358103 \\
0.944 99.77516352 \\
0.946 99.78810865 \\
0.948 99.79469477 \\
0.95 99.80514172 \\
0.952 99.8196766 \\
0.954 99.82830669 \\
0.956 99.84102471 \\
0.958 99.85215298 \\
0.96 99.86237282 \\
0.962 99.87327398 \\
0.964 99.88531068 \\
0.966 99.89530342 \\
0.968 99.90484193 \\
0.97 99.9157431 \\
0.972 99.92823401 \\
0.974 99.93754542 \\
0.976 99.94549419 \\
0.978 99.95208031 \\
0.98 99.95843932 \\
0.982 99.9668423 \\
0.984 99.97251999 \\
0.986 99.97683503 \\
0.988 99.98160429 \\
0.99 99.98728198 \\
0.992 99.99000727 \\
0.994 99.99273256 \\
0.996 99.99500363 \\
0.998 99.9970476 \\
1.0 99.99909157 \\
    };
    \addlegendentry{\textcolor{black}{Diff3F \cite{dutt2024diffusion} (0.92)}}
    
     \addplot [color=mycolor3, smooth, line width=\pckLineWidth]
           table[row sep=crcr]{%
0.002 22.5420149 \\
0.004 22.84951853 \\
0.006 23.77906977 \\
0.008 25.24141533 \\
0.01 27.32035792 \\
0.012 30.05813953 \\
0.014 33.24309593 \\
0.016 36.70557776 \\
0.018 40.25572311 \\
0.02 43.68504724 \\
0.022 47.03306686 \\
0.024 50.25072674 \\
0.026 53.34824673 \\
0.028 56.3251726 \\
0.03 59.14902798 \\
0.032 61.78302144 \\
0.034 64.25849382 \\
0.036 66.58157703 \\
0.038 68.77293786 \\
0.04 70.84302326 \\
0.042 72.71802326 \\
0.044 74.43563772 \\
0.046 75.97769804 \\
0.048 77.40029978 \\
0.05 78.66846839 \\
0.052 79.80559593 \\
0.054 80.8502907 \\
0.056 81.78392987 \\
0.058 82.63126817 \\
0.06 83.40729469 \\
0.062 84.08611919 \\
0.064 84.71180051 \\
0.066 85.27003089 \\
0.068 85.79987282 \\
0.07 86.29632994 \\
0.072 86.73055959 \\
0.074 87.1302689 \\
0.076 87.50272529 \\
0.078 87.83884448 \\
0.08 88.12749818 \\
0.082 88.42319222 \\
0.084 88.6777798 \\
0.086 88.89920967 \\
0.088 89.14017078 \\
0.09 89.35524164 \\
0.092 89.54782885 \\
0.094 89.73519259 \\
0.096 89.90416061 \\
0.098 90.06109193 \\
0.1 90.22029433 \\
0.102 90.35110828 \\
0.104 90.48010538 \\
0.106 90.603879 \\
0.108 90.71720567 \\
0.11 90.8232649 \\
0.112 90.92341933 \\
0.114 91.01607922 \\
0.116 91.11418968 \\
0.118 91.22206577 \\
0.12 91.31517987 \\
0.122 91.40193496 \\
0.124 91.48051417 \\
0.126 91.56091025 \\
0.128 91.63994368 \\
0.13 91.71829578 \\
0.132 91.79483103 \\
0.134 91.87113917 \\
0.136 91.93382086 \\
0.138 91.99854651 \\
0.14 92.05941134 \\
0.142 92.12799782 \\
0.144 92.19567587 \\
0.146 92.25199855 \\
0.148 92.30605015 \\
0.15 92.36259993 \\
0.152 92.41347202 \\
0.154 92.4700218 \\
0.156 92.52975109 \\
0.158 92.58471112 \\
0.16 92.63558321 \\
0.162 92.68986192 \\
0.164 92.73664608 \\
0.166 92.79319586 \\
0.168 92.85042696 \\
0.17 92.89539426 \\
0.172 92.95330669 \\
0.174 93.00667696 \\
0.176 93.0618641 \\
0.178 93.11409884 \\
0.18 93.1760992 \\
0.182 93.23401163 \\
0.184 93.28011446 \\
0.186 93.3264444 \\
0.188 93.37345567 \\
0.19 93.41910429 \\
0.192 93.4704306 \\
0.194 93.51448946 \\
0.196 93.572629 \\
0.198 93.6237282 \\
0.2 93.67528161 \\
0.202 93.72388263 \\
0.204 93.78088663 \\
0.206 93.82108467 \\
0.208 93.87422783 \\
0.21 93.92487282 \\
0.212 93.96847747 \\
0.214 94.01866824 \\
0.216 94.06431686 \\
0.218 94.10065407 \\
0.22 94.14017078 \\
0.222 94.17855196 \\
0.224 94.2157976 \\
0.226 94.25463299 \\
0.228 94.2909702 \\
0.23 94.32526344 \\
0.232 94.35637718 \\
0.234 94.38635538 \\
0.236 94.41156432 \\
0.238 94.44154251 \\
0.24 94.46765988 \\
0.242 94.49400436 \\
0.244 94.51944041 \\
0.246 94.54646621 \\
0.248 94.5644077 \\
0.25 94.58439317 \\
0.252 94.60801235 \\
0.254 94.63503815 \\
0.256 94.65797602 \\
0.258 94.67796148 \\
0.26 94.70089935 \\
0.262 94.72179324 \\
0.264 94.74018895 \\
0.266 94.76494368 \\
0.268 94.78515625 \\
0.27 94.80377907 \\
0.272 94.82103924 \\
0.274 94.84193314 \\
0.276 94.86146439 \\
0.278 94.88485647 \\
0.28 94.90529615 \\
0.282 94.92619004 \\
0.284 94.94753815 \\
0.286 94.97138445 \\
0.288 94.99046148 \\
0.29 95.01430778 \\
0.292 95.03633721 \\
0.294 95.05518714 \\
0.296 95.07358285 \\
0.298 95.09538517 \\
0.3 95.12127544 \\
0.302 95.14035247 \\
0.304 95.16124637 \\
0.306 95.18373001 \\
0.308 95.2162064 \\
0.31 95.24096112 \\
0.312 95.26707849 \\
0.314 95.29569404 \\
0.316 95.31931323 \\
0.318 95.34633903 \\
0.32 95.37495458 \\
0.322 95.40084484 \\
0.324 95.42196584 \\
0.326 95.43899891 \\
0.328 95.46307231 \\
0.33 95.48237645 \\
0.332 95.4957758 \\
0.334 95.5168968 \\
0.336 95.53506541 \\
0.338 95.5527798 \\
0.34 95.57117551 \\
0.342 95.58866279 \\
0.344 95.6070585 \\
0.346 95.62817951 \\
0.348 95.64498547 \\
0.35 95.66656068 \\
0.352 95.68654615 \\
0.354 95.70494186 \\
0.356 95.72379179 \\
0.358 95.74286882 \\
0.36 95.76580669 \\
0.362 95.78601926 \\
0.364 95.80850291 \\
0.366 95.82757994 \\
0.368 95.84506722 \\
0.37 95.86868641 \\
0.372 95.88958031 \\
0.374 95.90956577 \\
0.376 95.93522892 \\
0.378 95.96498001 \\
0.38 95.99404978 \\
0.382 96.02039426 \\
0.384 96.04923692 \\
0.386 96.07694404 \\
0.388 96.10192587 \\
0.39 96.13167696 \\
0.392 96.15620458 \\
0.394 96.18550145 \\
0.396 96.21071039 \\
0.398 96.23455669 \\
0.4 96.26135538 \\
0.402 96.30064499 \\
0.404 96.32244731 \\
0.406 96.34538517 \\
0.408 96.37218387 \\
0.41 96.39693859 \\
0.412 96.4182867 \\
0.414 96.43804506 \\
0.416 96.45757631 \\
0.418 96.48551054 \\
0.42 96.51208212 \\
0.422 96.53433866 \\
0.424 96.55886628 \\
0.426 96.5804415 \\
0.428 96.60383358 \\
0.43 96.62904251 \\
0.432 96.65379724 \\
0.434 96.68513808 \\
0.436 96.71057413 \\
0.438 96.73941679 \\
0.44 96.77212028 \\
0.442 96.80073583 \\
0.444 96.82957849 \\
0.446 96.85524164 \\
0.448 96.8809048 \\
0.45 96.91042878 \\
0.452 96.94335938 \\
0.454 96.97515443 \\
0.456 97.00853924 \\
0.458 97.03783612 \\
0.46 97.07190225 \\
0.462 97.09665698 \\
0.464 97.12640807 \\
0.466 97.15888445 \\
0.468 97.18954397 \\
0.47 97.21838663 \\
0.472 97.25154433 \\
0.474 97.28220385 \\
0.476 97.31013808 \\
0.478 97.3376181 \\
0.48 97.36055596 \\
0.482 97.38690044 \\
0.484 97.41415334 \\
0.486 97.44095203 \\
0.488 97.47206577 \\
0.49 97.49841025 \\
0.492 97.52679869 \\
0.494 97.54905523 \\
0.496 97.57494549 \\
0.498 97.59833757 \\
0.5 97.62127544 \\
0.502 97.64444041 \\
0.504 97.67532703 \\
0.506 97.70099019 \\
0.508 97.73233103 \\
0.51 97.76480741 \\
0.512 97.78956214 \\
0.514 97.81477108 \\
0.516 97.84611192 \\
0.518 97.86746003 \\
0.52 97.88858103 \\
0.522 97.90811228 \\
0.524 97.93786337 \\
0.526 97.95989281 \\
0.528 97.98101381 \\
0.53 97.99940952 \\
0.532 98.02348292 \\
0.534 98.04755632 \\
0.536 98.06299964 \\
0.538 98.07957849 \\
0.54 98.10138081 \\
0.542 98.11795967 \\
0.544 98.13567406 \\
0.546 98.15316134 \\
0.548 98.16837754 \\
0.55 98.182004 \\
0.552 98.19676599 \\
0.554 98.21107376 \\
0.556 98.23151344 \\
0.558 98.24377725 \\
0.56 98.25490552 \\
0.562 98.2662609 \\
0.564 98.28147711 \\
0.566 98.29442224 \\
0.568 98.30350654 \\
0.57 98.31554324 \\
0.572 98.32689862 \\
0.574 98.33757267 \\
0.576 98.35051781 \\
0.578 98.36096475 \\
0.58 98.37390988 \\
0.582 98.38912609 \\
0.584 98.40161701 \\
0.586 98.41728743 \\
0.588 98.43432049 \\
0.59 98.45203488 \\
0.592 98.46429869 \\
0.594 98.47928779 \\
0.596 98.494504 \\
0.598 98.51199128 \\
0.6 98.52584484 \\
0.602 98.53992551 \\
0.604 98.55218932 \\
0.606 98.56695131 \\
0.608 98.57944222 \\
0.61 98.5907976 \\
0.612 98.60101744 \\
0.614 98.61146439 \\
0.616 98.62236555 \\
0.618 98.6298601 \\
0.62 98.63780887 \\
0.622 98.64598474 \\
0.624 98.65416061 \\
0.626 98.66120094 \\
0.628 98.6689226 \\
0.63 98.67414608 \\
0.632 98.68050509 \\
0.634 98.68550145 \\
0.636 98.69254179 \\
0.638 98.69731105 \\
0.64 98.70389717 \\
0.642 98.71252725 \\
0.644 98.72024891 \\
0.646 98.72842478 \\
0.648 98.73478379 \\
0.65 98.74318677 \\
0.652 98.74931868 \\
0.654 98.75704033 \\
0.656 98.76385356 \\
0.658 98.7708939 \\
0.66 98.77793423 \\
0.662 98.7831577 \\
0.664 98.78883539 \\
0.666 98.79292333 \\
0.668 98.7981468 \\
0.67 98.80564135 \\
0.672 98.80882086 \\
0.674 98.81540698 \\
0.676 98.8190407 \\
0.678 98.8251726 \\
0.68 98.82971475 \\
0.682 98.83698219 \\
0.684 98.83948038 \\
0.686 98.84243278 \\
0.688 98.8462936 \\
0.69 98.85106286 \\
0.692 98.85537791 \\
0.694 98.85923874 \\
0.696 98.86173692 \\
0.698 98.86605196 \\
0.7 98.87104833 \\
0.702 98.87672602 \\
0.704 98.88172238 \\
0.706 98.88785429 \\
0.708 98.89375908 \\
0.71 98.89966388 \\
0.712 98.90307049 \\
0.714 98.90783975 \\
0.716 98.91329033 \\
0.718 98.9182867 \\
0.72 98.92396439 \\
0.722 98.93123183 \\
0.724 98.93895349 \\
0.726 98.94508539 \\
0.728 98.95076308 \\
0.73 98.95553234 \\
0.732 98.96234557 \\
0.734 98.96734193 \\
0.736 98.97415516 \\
0.738 98.98233103 \\
0.74 98.99141533 \\
0.742 98.99845567 \\
0.744 99.00890262 \\
0.746 99.02003089 \\
0.748 99.03025073 \\
0.75 99.04478561 \\
0.752 99.05386991 \\
0.754 99.06408975 \\
0.756 99.07521802 \\
0.758 99.083621 \\
0.76 99.09293241 \\
0.762 99.09951853 \\
0.764 99.10837573 \\
0.766 99.11700581 \\
0.768 99.12654433 \\
0.77 99.13585574 \\
0.772 99.14516715 \\
0.774 99.15129906 \\
0.776 99.15924782 \\
0.778 99.16810501 \\
0.78 99.17628089 \\
0.782 99.18309411 \\
0.784 99.19217842 \\
0.786 99.20217115 \\
0.788 99.21102834 \\
0.79 99.22329215 \\
0.792 99.23532885 \\
0.794 99.24486737 \\
0.796 99.25395167 \\
0.798 99.2637173 \\
0.8 99.27529978 \\
0.802 99.28642805 \\
0.804 99.29823765 \\
0.806 99.30754906 \\
0.808 99.31981286 \\
0.81 99.33230378 \\
0.812 99.33934411 \\
0.814 99.34979106 \\
0.816 99.35864826 \\
0.818 99.36523438 \\
0.82 99.37091206 \\
0.822 99.37477289 \\
0.824 99.37976926 \\
0.826 99.38431141 \\
0.828 99.3870367 \\
0.83 99.39384993 \\
0.832 99.39839208 \\
0.834 99.40225291 \\
0.836 99.40747638 \\
0.838 99.41247275 \\
0.84 99.41769622 \\
0.842 99.4231468 \\
0.844 99.42927871 \\
0.846 99.43563772 \\
0.848 99.44176962 \\
0.85 99.44585756 \\
0.852 99.45403343 \\
0.854 99.45789426 \\
0.856 99.46289063 \\
0.858 99.47038517 \\
0.86 99.47787972 \\
0.862 99.48560138 \\
0.864 99.49445858 \\
0.866 99.50036337 \\
0.868 99.50399709 \\
0.87 99.51194586 \\
0.872 99.5221657 \\
0.874 99.52966025 \\
0.876 99.53488372 \\
0.878 99.54056141 \\
0.88 99.54782885 \\
0.882 99.55441497 \\
0.884 99.56054688 \\
0.886 99.56781432 \\
0.888 99.57394622 \\
0.89 99.57916969 \\
0.892 99.58711846 \\
0.894 99.59302326 \\
0.896 99.6034702 \\
0.898 99.61323583 \\
0.9 99.62050327 \\
0.902 99.62822493 \\
0.904 99.6364008 \\
0.906 99.64503089 \\
0.908 99.6513899 \\
0.91 99.6604742 \\
0.912 99.67046693 \\
0.914 99.67818859 \\
0.916 99.68886265 \\
0.918 99.69840116 \\
0.92 99.70998365 \\
0.922 99.72111192 \\
0.924 99.73224019 \\
0.926 99.7415516 \\
0.928 99.74927326 \\
0.93 99.75767624 \\
0.932 99.76380814 \\
0.934 99.77016715 \\
0.936 99.77834302 \\
0.938 99.78629179 \\
0.94 99.79287791 \\
0.942 99.80014535 \\
0.944 99.81172783 \\
0.946 99.81831395 \\
0.948 99.82694404 \\
0.95 99.83511991 \\
0.952 99.84397711 \\
0.954 99.85419695 \\
0.956 99.86078307 \\
0.958 99.87531795 \\
0.96 99.88485647 \\
0.962 99.89280523 \\
0.964 99.90211664 \\
0.966 99.9098383 \\
0.968 99.91710574 \\
0.97 99.92687137 \\
0.972 99.93572856 \\
0.974 99.94367733 \\
0.976 99.95094477 \\
0.978 99.95912064 \\
0.98 99.9670694 \\
0.982 99.97206577 \\
0.984 99.97524528 \\
0.986 99.978879 \\
0.988 99.98251272 \\
0.99 99.98637355 \\
0.992 99.99114281 \\
0.994 99.9938681 \\
0.996 99.99545785 \\
0.998 99.99750182 \\
1.0 100.0 \\
         };
         \addlegendentry{\textcolor{black}{Diff3F+$\chi_{\text{DINO+SD}}$ \textbf{(0.94)}}}
        
	\end{axis}
\end{tikzpicture}

%% file: tables/qualitative_matching.tex
\newcommand{\imageheightam}{0.095\textheight}
\newcommand{\imageheightbm}{0.085\textheight}
\newcommand{\imagespacingm}{\hspace{0.1cm}}

\begin{figure}[!ht]
    \centering
    \begin{tabular}{cccccc}
         Source &  Target & Diff3F & Ours\\
 
        \adjustbox{valign=m}{\includegraphics[height=\imageheightam]{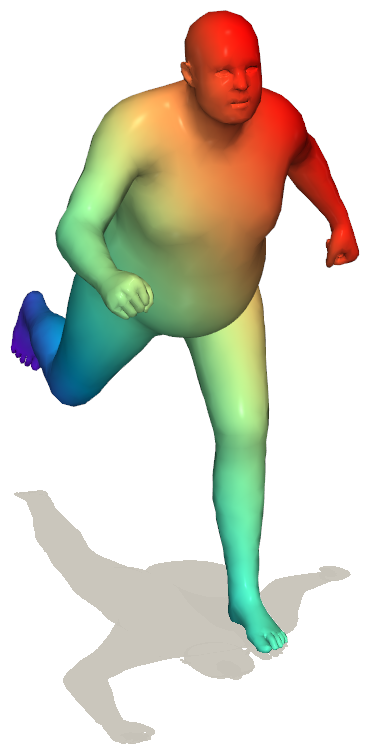}} & 
        \imagespacingm
        \adjustbox{valign=m}{\includegraphics[height=\imageheightam]{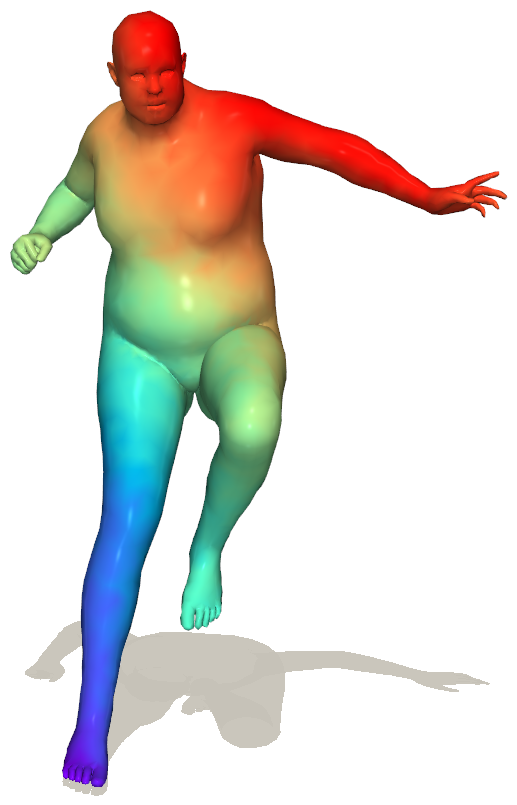}} &
        \imagespacingm
        \adjustbox{valign=m}{\includegraphics[height=\imageheightam]{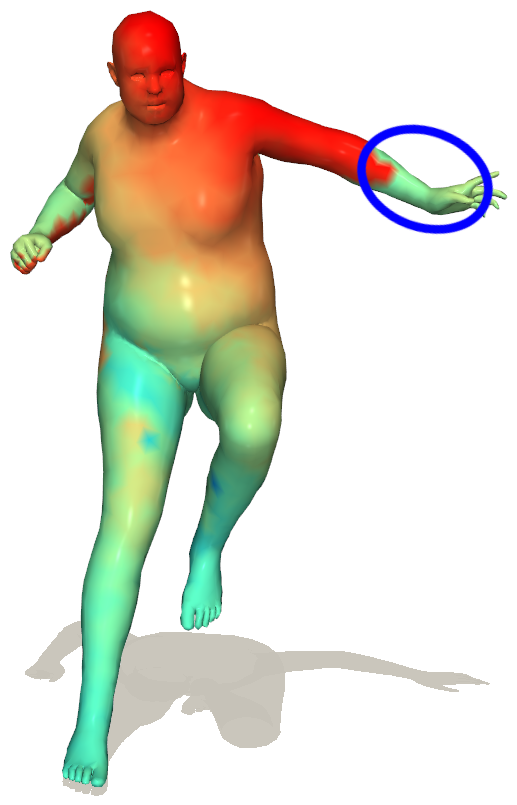}} & 
        \imagespacingm
        \adjustbox{valign=m}{\includegraphics[height=\imageheightam]{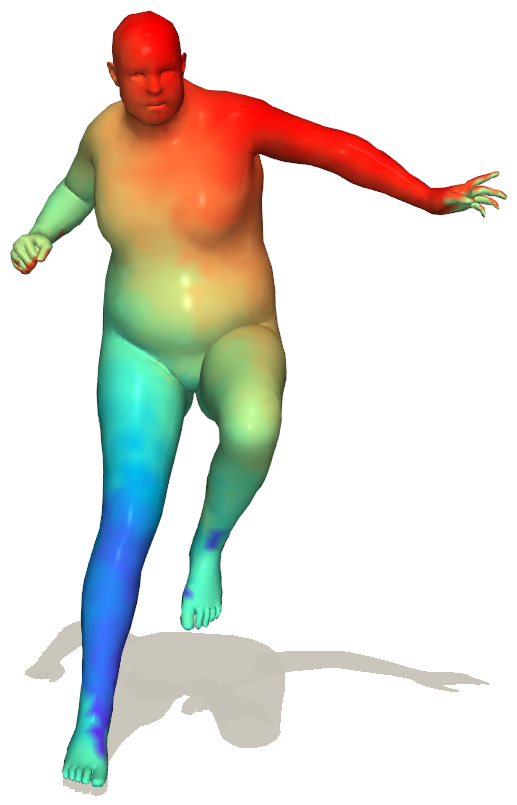}} \\

         \adjustbox{valign=m}{\includegraphics[height=\imageheightbm]{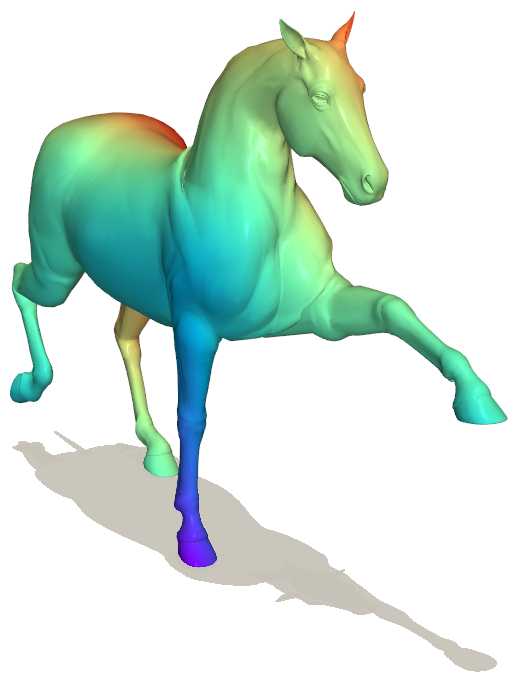}} &
         \imagespacingm
         \adjustbox{valign=m}{\includegraphics[height=\imageheightbm]{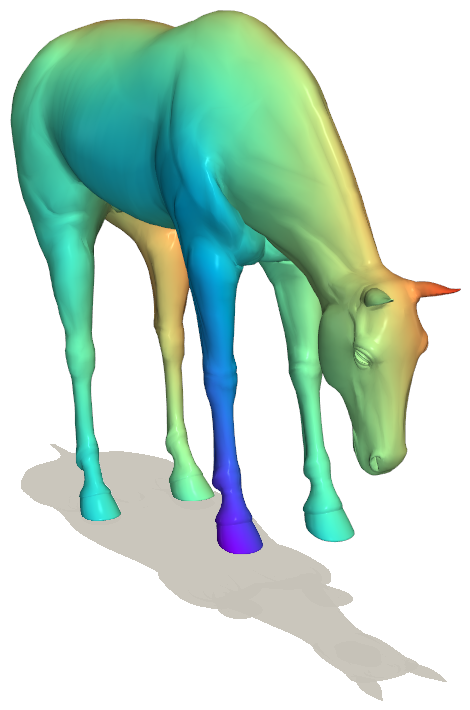}} &
         \imagespacingm
         \adjustbox{valign=m}{\includegraphics[height=\imageheightbm]{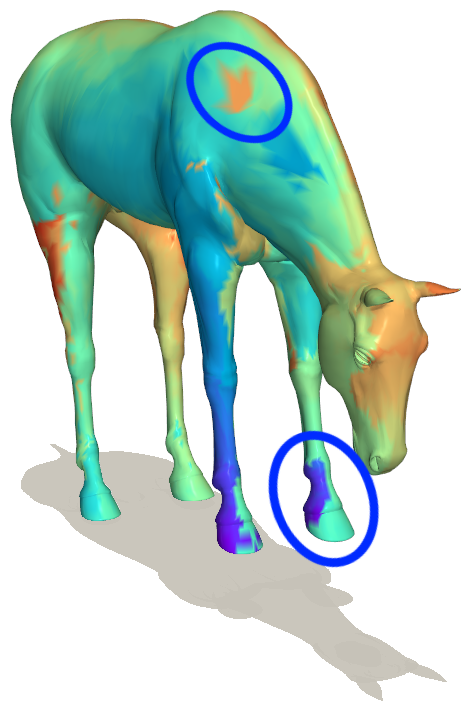}} &
         \imagespacingm
         \adjustbox{valign=m}{\includegraphics[height=\imageheightbm]{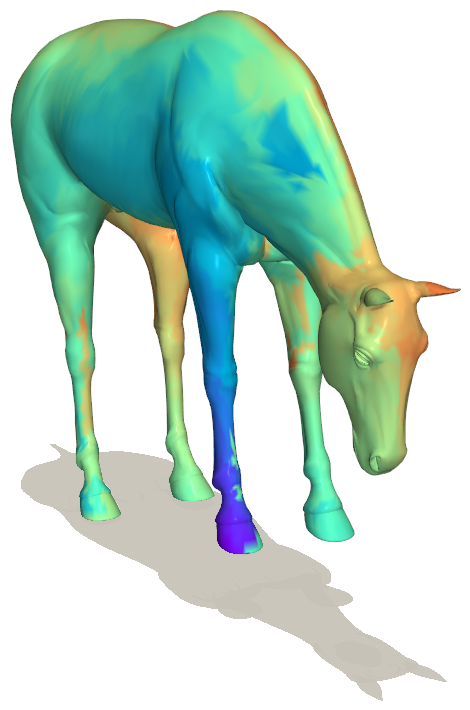}} \\

    \end{tabular}
    
    \caption{Qualitative results of shape correspondence using Diff3F and our chirality features. Our approach is able to correct for the left/right ambiguity as shown by the blue ovals.}
    \label{fig:qualitative_matching}
\end{figure}

%% file: tables/part_seg_fig.tex
\newcommand{\imageheightac}{0.08\textheight}
\newcommand{\imageheightbc}{0.08\textheight}
\newcommand{\imagespacingc}{\hspace{0.3cm}}

\begin{figure}[!ht]
    \centering
    \begin{tabular}{ccccc}
    
         \rotatebox[origin=c]{90}{\textbf{Diff3F}} &
         \adjustbox{valign=m}{\includegraphics[height=\imageheightac]{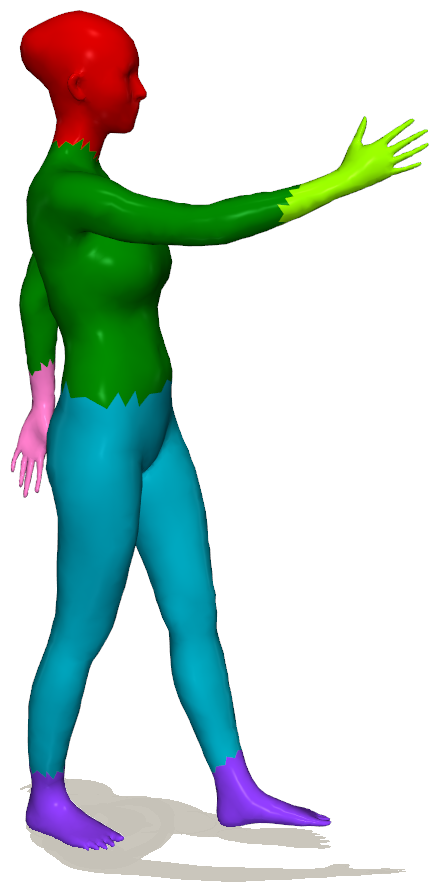}} &
         \imagespacingc
         \adjustbox{valign=m}{\includegraphics[height=\imageheightac]{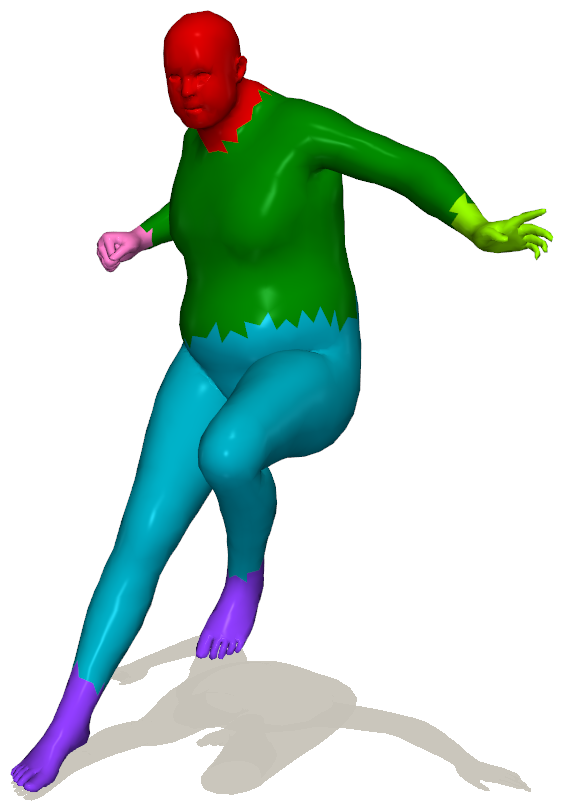}} & 
         \imagespacingc
         \adjustbox{valign=m}{\includegraphics[height=\imageheightac]{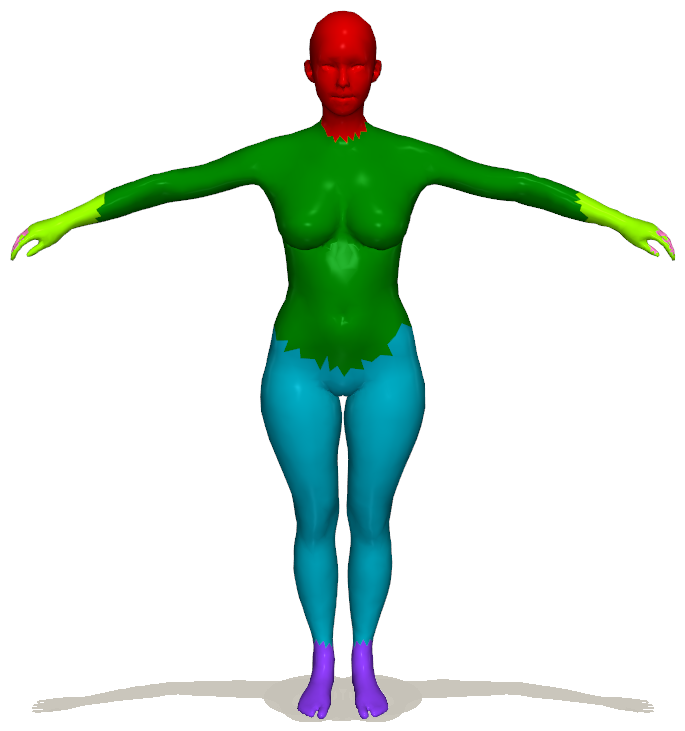}} & 
         \imagespacingc
         \adjustbox{valign=m}{\includegraphics[height=\imageheightac]{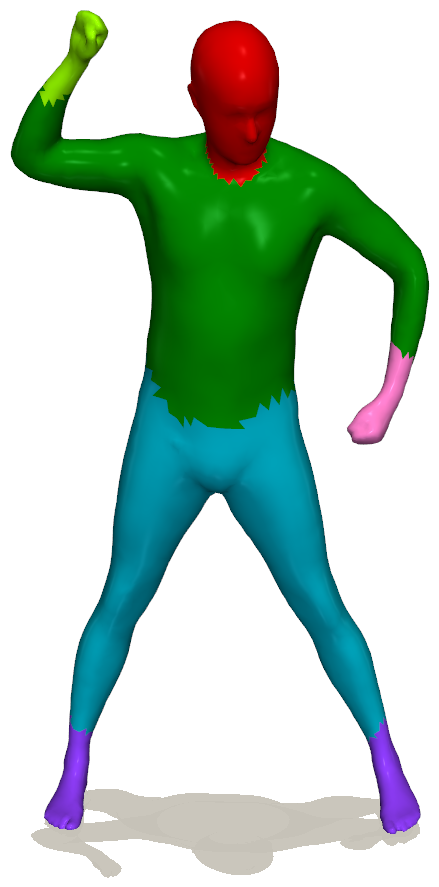}} \\

         \rotatebox[origin=c]{90}{\textbf{Ours}} &
         \adjustbox{valign=m}{\includegraphics[height=\imageheightbc]{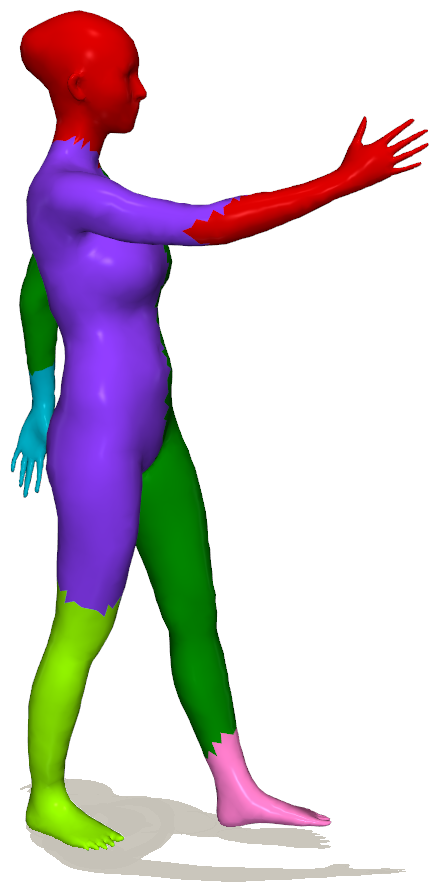}} &
         \imagespacingc
         \adjustbox{valign=m}{\includegraphics[height=\imageheightbc]{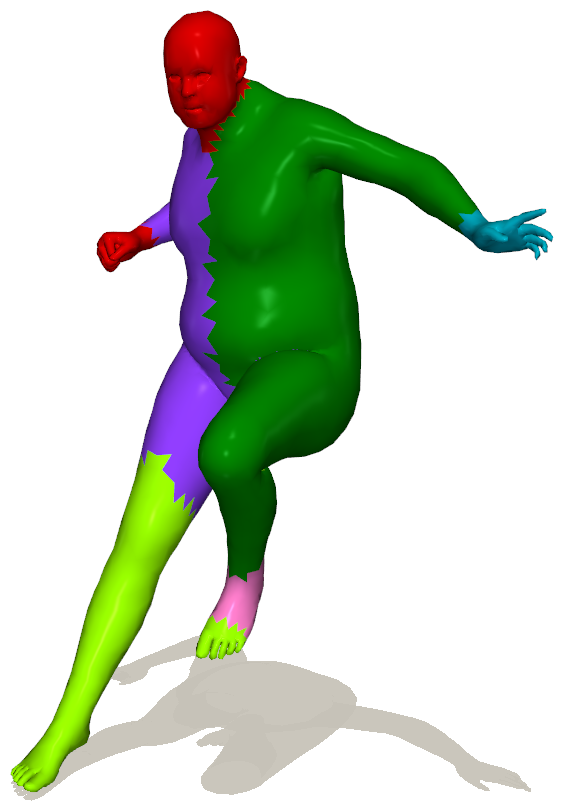}} & 
         \imagespacingc
         \adjustbox{valign=m}{\includegraphics[height=\imageheightbc]{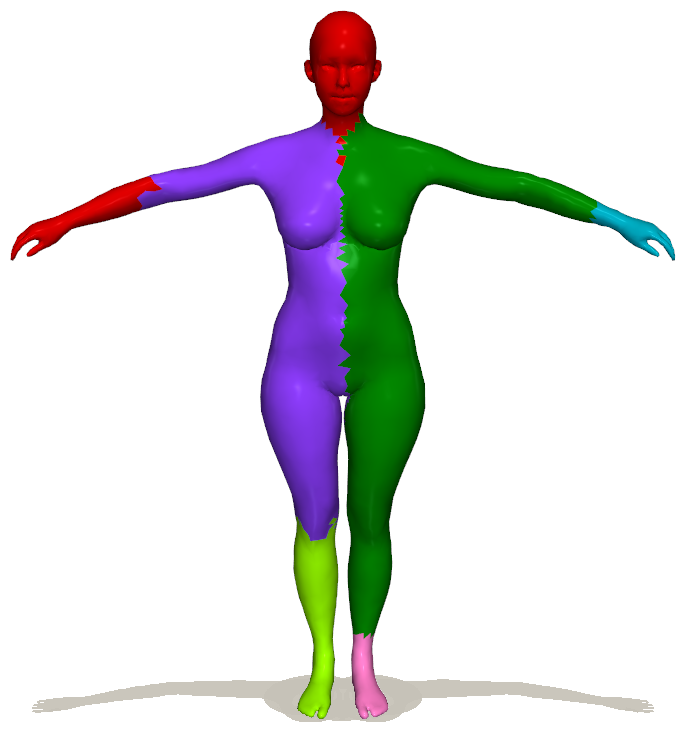}} & 
         \imagespacingc
         \adjustbox{valign=m}{\includegraphics[height=\imageheightbc]{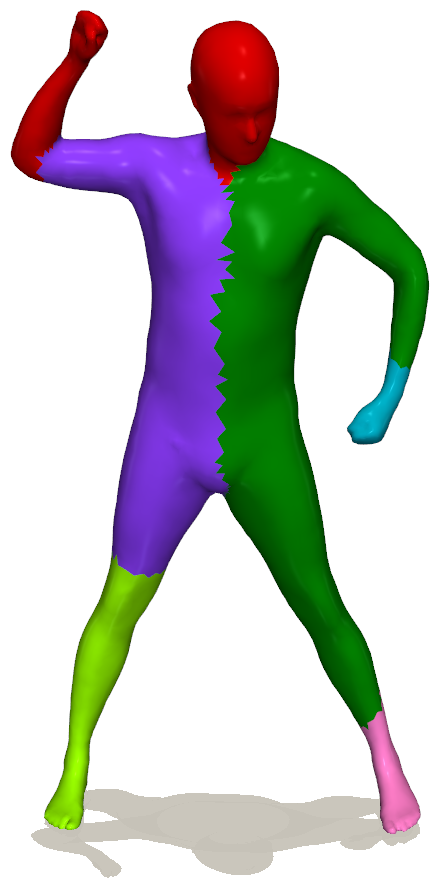}} \\

    \end{tabular}
   
    \caption{Visualisations of part segmentation using Diff3F \cite{dutt2024diffusion} and our features. Our chirality features can differentiate between the left and right parts of the body (e.g.~hands and legs), whereas Diff3F clusters the left and right hands and legs together.}
    \label{fig:part_seg}
\end{figure}

%% file: tables/partiality_fig.tex
\newcommand{\imagespacingpartial}{\hspace{0.6cm}}

\begin{figure}[!h]
    \centering
    \begin{tabular}{ccc}

      \adjustbox{valign=m}{\includegraphics[height=0.081\textheight]{tables/qualitaitve_results_left_right/partial_0_ours.png}} &
      \imagespacingpartial
     \adjustbox{valign=m}{\includegraphics[height=0.081\textheight]{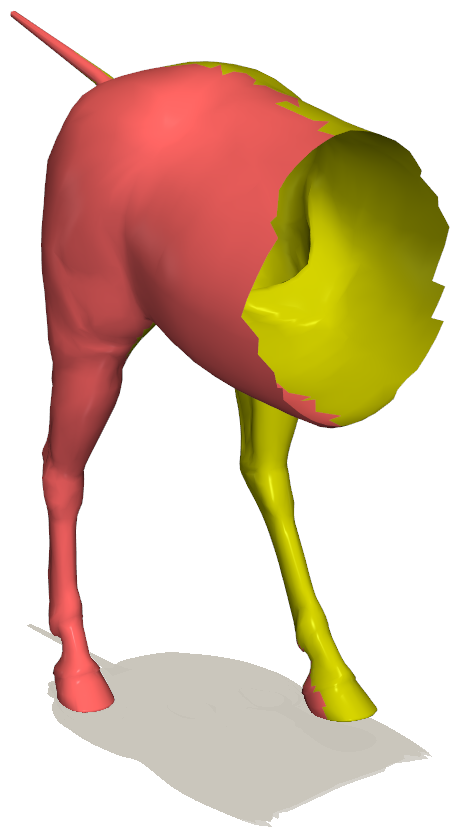}} &
     \imagespacingpartial
     \adjustbox{valign=m}{\includegraphics[height=0.081\textheight]{tables/qualitaitve_results_left_right/partial_2_ours.png}} \\

    \end{tabular}
    
    \caption{Our chirality features can distinguish left and right parts of partial shapes on both human and animal shapes.}
    \label{fig:partiality_fig}
\end{figure}

%% file: tables/anisotropic_fig.tex
\newcommand{\imageheightanisotropica}{0.11\textheight}
\newcommand{\imageheightanisotropicb}{0.13\textheight}
\newcommand{\imagespacinganisotropic}{\hspace{0.3cm}}

\begin{figure}[!ht]
    \centering
    \begin{tabular}{cccc}

      \adjustbox{valign=m}{\includegraphics[height=\imageheightanisotropica]{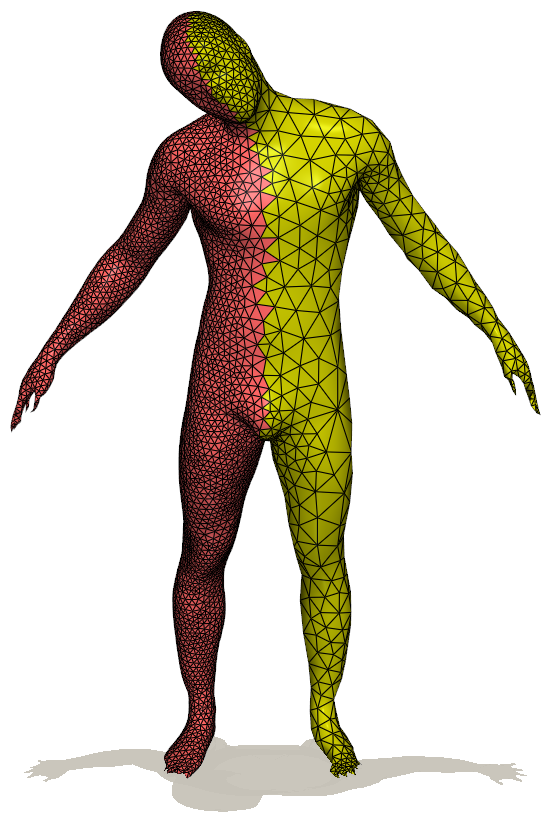}} &
      \imagespacinganisotropic
     \adjustbox{valign=m}{\includegraphics[height=\imageheightanisotropicb]{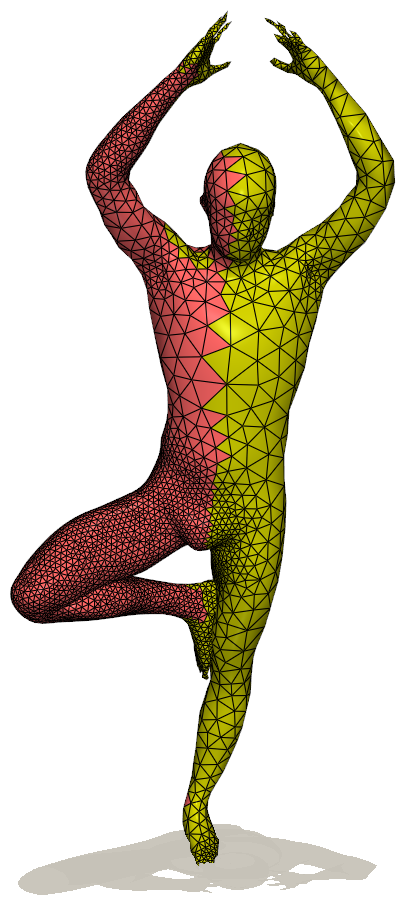}} &
    \imagespacinganisotropic
     \adjustbox{valign=m}{\includegraphics[height=\imageheightanisotropica]{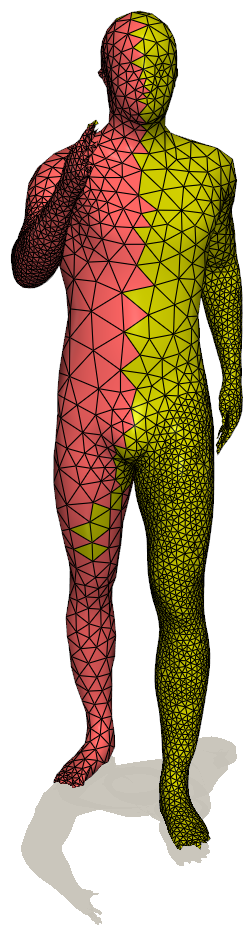}} & 
     \imagespacinganisotropic
     \adjustbox{valign=m}{\includegraphics[height=\imageheightanisotropicb]{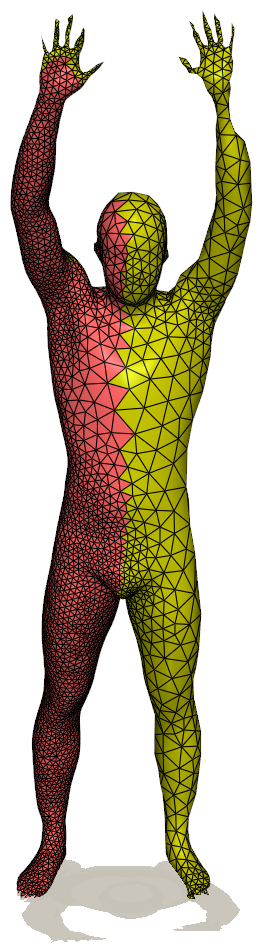}} \\

    \end{tabular}
    
    \caption{Our approach is robust to anisotropic meshes and can handle meshes with different discretisation. }
    \label{fig:anisotropic_shapes}
\end{figure}

%% file: sec/5_ablation.tex
\subsection{Ablation study}
\label{sec: ablation}
 
We conduct ablative experiments to verify our design choices, using use the FAUST \cite{bogo2014faust} and SMAL \cite{zheng2015skeleton} datasets. Tab.~\ref{tab:ablation} summarises our findings. By comparing the first four columns, we conclude all losses are crucial for obtaining accurate chirality features. By comparing the last three columns, we observe combining SD and DINO yields the best results across both human and animal datasets.
\input{tables/ablation}

%% file: tables/ablation.tex
\begin{table}
\setlength{\tabcolsep}{3.3pt}
  \centering
  \resizebox{1.0\columnwidth}{!}{%
  \begin{tabular}{@{}lccccccc@{}}
    \toprule
     w/o & $\mathcal{L}_{\text{dis}}$
    & $\mathcal{L}_{\text{var}}$
    & $\mathcal{L}_{\text{fif}}$
    & $\mathcal{L}_{\text{inv}}$
    & SD
    & DINO
    & full\\
    \midrule
   \textbf{FAUST} & 51.34 & 77.02 & 90.95 & \textbf{96.26} & 89.21 & 81.47 & \underline{95.79} \\
    & $\pm$ \textit{0.96} & $\pm$ \textit{8.87} & $\pm$ \textit{7.86} & $\pm$ \textit{0.49} & $\pm$ \textit{2.48} & $\pm$ \textit{24.96} & $\pm$ \textit{0.50} \\
    \midrule
   \textbf{SMAL} & 51.33 & 74.37 & \textbf{96.41} & 76.72 & 71.65 & 94.21 & \underline{94.71} \\
   & $\pm$ \textit{0.91} & $\pm$ \textit{0.97} & $\pm$ \textit{0.27} & $\pm$ \textit{3.80} & $\pm$ \textit{1.17} & $\pm$ \textit{3.89} & $\pm$ \textit{2.59} \\
    
    \bottomrule
  \end{tabular}
  }
  \caption{Ablation study on the FAUST and SMAL datasets. We use the chirality accuracy $acc_{\chi}$ ($\uparrow$) as evaluation metric. The best/second best results in each column are highlighted/underlined, respectively.}
  \label{tab:ablation}
\end{table}

%% file: sec/6_discussion.tex
\section{Limitations \& Future works}
\label{sec: discussion}

Although our proposed method shows superiority compared to other methods for left-and-right disentanglement, and the extracted chirality feature performs well on various downstream tasks, further improvements are still possible. Firstly, methods using 3D descriptors aggregated from 2D images (e.g.~Diff3F \cite{dutt2024diffusion}, DenseMatcher \cite{zhu2025densematcher}) struggle with occluded vertices and fail to learn reliable features. Our method simply inherits this shortcoming from Diff3F \cite{dutt2024diffusion}. Deforming shapes, to reduce occlusions, is an interesting direction for future work. Secondly, our model takes meshes as input, since $\mathcal{L}_{\text{var}}$ relies on the edges of the mesh, which limits the direct applicability of our method to point clouds. However, we show in Sec.~\textcolor{iccvblue}{C} of the supplementary material that our method (with a simple $k$-nearest neighbour approximation of the connectivity of the input point cloud) can give reasonable results. Additionally, despite incorporating the total variation loss, we occasionally observe patches with incorrect features, as seen in Fig.~\ref{fig:teaser} (left side, top right). Another type of inaccuracy might arise when two body parts with different chirality are in close proximity or touching each other (Fig.~\ref{fig:anisotropic_shapes}, middle left). Future works combining geometric constraints might alleviate these problems.

%% file: sec/7_conclusion.tex
\section{Conclusion}
\label{sec:conclusion}

Chirality information plays an important role in visual computing and has been severely under-explored in the shape analysis field. In this paper, we propose an unsupervised method to disentangle per vertex chirality features from semantic and geometric features aggregated from 2D foundation models (DINO-V2 \cite{oquabdinov2} and StableDiffusion \cite{rombach2022high}). The chirality features disentangled by our proposed pipeline show superiority compared to various other features/methods on left-right distinguishing tasks both quantitatively and qualitatively. Furthermore, combined with our chirality features, other vertex feature descriptors show better performance on both shape matching and part segmentation tasks on various datasets. Additionally, the generalisation tests on partial and anisotropic shapes confirm the robustness of our method and also enlarge the application scenarios of our model due to more realistic properties. To conclude, we believe that our proposed pipeline and extracted chirality features will benefit future research in shape analysis and other visual computing areas. 

%% file: sec/8_acknowledge.tex
\section*{Acknowledgments}
We thank Paul Roetzer for the valuable feedback on earlier drafts of this manuscript.
This work is supported by the ERC starting grant no.~101160648 (Harmony).

%% file: supplementary.tex
\clearpage
\setcounter{page}{1}
\maketitlesupplementary

\setcounter{figure}{0}
\setcounter{table}{0}
\setcounter{section}{0}
\renewcommand{\thesection}{\Alph{section}}
\renewcommand\thefigure{F.\arabic{figure}} \renewcommand\thetable{T.\arabic{table}}

In the supplementary materials, we include information about the data generation (Sec.~\ref{sec:data_gen}), architectural decisions (Sec.~\ref{sec:archit}), additional results (Sec.~\ref{sec:repr} \& Sec.~\ref{sec:faust}) and implementation details (Sec.~\ref{sec:impl}).

\section{BeCoS Data Generation}
\label{sec:data_gen}
When generating BeCoS [\textcolor{iccvblue}{18}]
with default settings, each shape is used multiple times to generate a wide variety of shape matching pairs. This results in a train/validation/test split of 
$20370/274/284$ shapes. Since we use the framework to propagate annotations to all shapes, effectively ignoring the pairings, we restrict BeCoS to use each shape at most once, resulting in the train/validation/test split of $1980/284/274$ shapes. To generate BeCoS\textsubscript{-h} and BeCoS\textsubscript{-a}, we filter the resulting dataset for human and animal entries, respectively. Note that we assign the \textit{centaur} shapes from TOSCA dataset [\textcolor{iccvblue}{7}] 
to BeCoS\textsubscript{-a}.

\section{Architectural Design}
\label{sec:archit}
In our method, each vertex gets assigned two scalar features $\chi_v$ and $\bar{\chi}_v$, derived from a non-linear projection $\tilde{g}$ of its original and mirrored Diff3F features. Specifically, we extract a specific component from $\tilde{g}(\mathcal{F}_v)$ and normalise it by the L2 Norm of the whole feature vector. The normalisation ensures that $\chi_v$ reflects the relative magnitude of the chosen component compared to the whole feature, while remaining invariant to scaling. Compared to a fixed non-linearity such as $\tanh{[\tilde{g}(\mathcal{F}_v)]_0}$, which only uses the first component, the normalisation incorporates additional information and promotes more stable learning dynamics. To empirically confirm our choice, we provide an ablation by training both architectures and evaluating them in the left-right disentanglement task. The results can be found in Table \ref{tab:chirality_accuracy_tanh}.

\input{tables/left_right_results_tanh}

\section{Different shape representations.}
\label{sec:repr}

Our $\mathcal{L}_{\text{var}}$ loss requires mesh connectivity to regularize the smoothness and boundary of our solution, effectively restricting the applicability of our method to meshes. However, connectivity information for different kinds of shape representations, like point clouds, can be approximated. For example using mutual k-nearest neighbors among vertices.
To get a first impression of the applicability of our method on point clouds, we train a network using SD + Dino features generated from \emph{rendered point cloud images}. We replace the edges used in $\mathcal{L}_{\text{var}}$ with a primitive mutual k-nearest neighbor ($k=5$) connectiviy approximation. The models are evaluated in the left/right classification on \textsc{FAUST}. We also include results for the point cloud setting without $\mathcal{L}_{\text{var}}$. The table and figure below show that our method with k-nn approximation performs robustly on most vertices of the point cloud. 

\begin{table}[H]
  \setlength{\tabcolsep}{5pt}
   \centering
  \begin{tabular}{@{}lccc@{}}
    \toprule
     \footnotesize{Input} & \multicolumn{2}{c}{\footnotesize{Point cloud}} & \footnotesize{Mesh} \\ 
     \footnotesize{Losses} & \footnotesize{w/o $\mathcal{L}_{\text{var}}$} & \footnotesize{KNN- $\mathcal{L}_{\text{var}}$} & \footnotesize{Full} \\
    \midrule
    \footnotesize{Acc} & 67.08 & 92.87 & 94.76 \\
    \bottomrule
  \end{tabular}
    \caption{The model trained on point clouds using approximate k-NN connectivity information reaches a high accuracy, compared to a model trained without connectiviy information.}
\end{table}

\newcommand{\imagespacingf}{\hspace{1.5cm}}

\begin{figure}[!ht]
\centering
\begin{tabular}{cc}
     \footnotesize{w/o $\mathcal{L}_{\text{var}}$} &  \footnotesize{KNN-$\mathcal{L}_{\text{var}}$} \\
    \includegraphics[height=0.3\textheight]{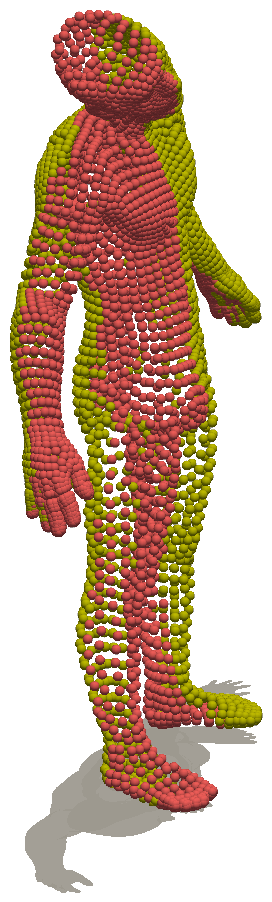} & 
    \imagespacingf
    \includegraphics[height=0.3\textheight]{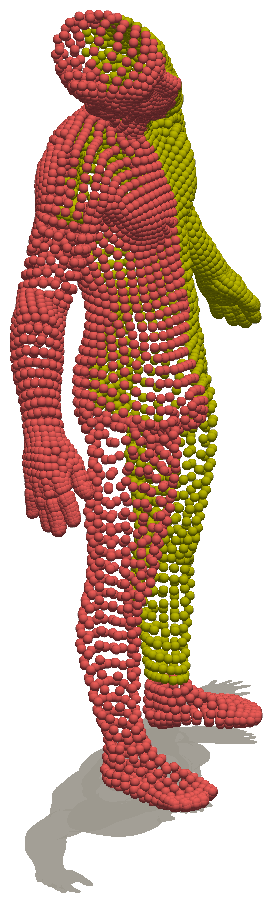}
\end{tabular}
\caption{Qualitative results of models trained on point clouds with and without approximated k-NN connectivity information. Without the $\mathcal{L}_{\text{var}}$ loss, the length of the boundary is not regularized, resulting in an inaccurate left/right split. With the approximate $\mathcal{L}_{\text{var}}$ loss, the model is able to correctly classify most of the points.}
\end{figure}
\noindent The inaccurate assignment of the left foot shows that there are remaining open challenges. Since our main focus is on 3D meshes, we leave this exploration for future work.

\section{Additional shape matching results.}
\label{sec:faust}
We provide additional results for the shape matching task on the \textsc{FAUST} benchmark [\textcolor{iccvblue}{6}]. 
We compute vertex correspondences using cosine similarity between the vertex features of the source and target shape. When combined with Diff3F features, our features achieve a $50.0\%$ decrease in error for the inter-subject and $42.2\%$ for the intra-subject task, compared to Diff3F features. Qualitative results are shown in Fig.~\ref{fig:additional_results}.

\newcommand{\imagespacingtf}{\hspace{0.2cm}}
\newcommand{\imagespacingtff}{\hspace{0.6cm}}
\begin{figure}[h]
    \centering
    \begin{tabular}{ccc}
    \multicolumn{3}{c}{\small{\textbf{Inter-Subject}}}\\
     \footnotesize{Source} & \footnotesize{Diff3F} & \footnotesize{Diff3F + Ours} \\
    \includegraphics[height=0.5\columnwidth]{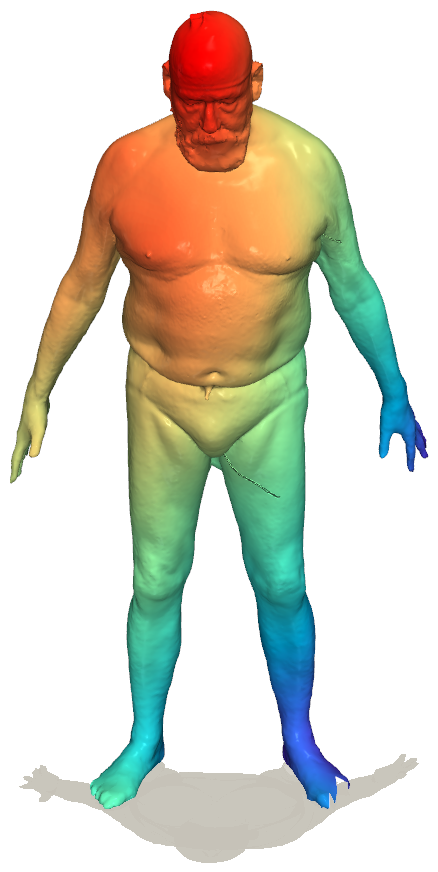} & 
    \imagespacingtf
    \includegraphics[height=0.5\columnwidth]{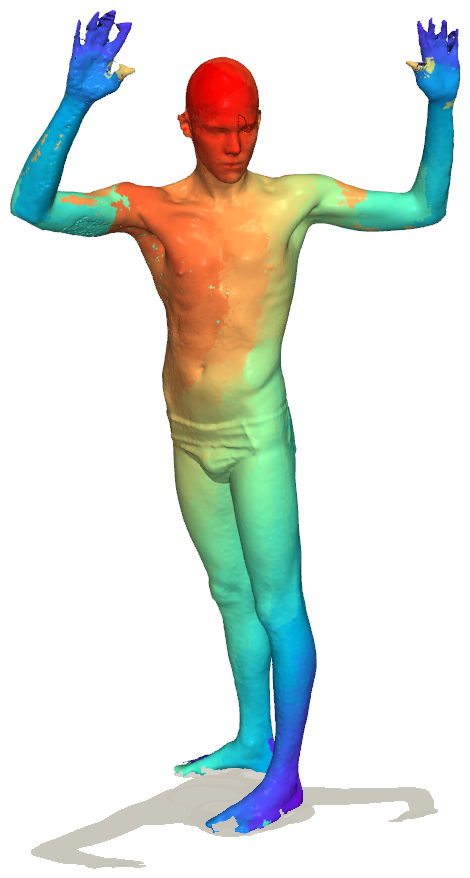} &
    \imagespacingtf
    \includegraphics[height=0.5\columnwidth]{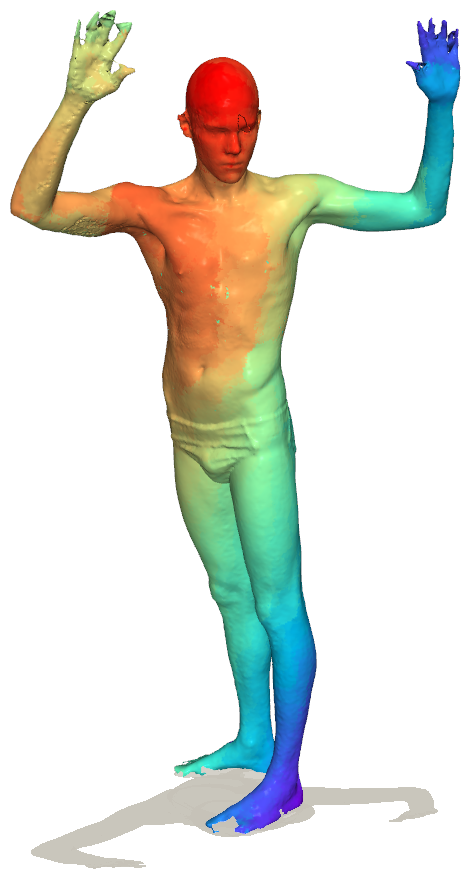}
    \end{tabular}
    \centering
    \begin{tabular}{ccc}
    \\
    \\
    \multicolumn{3}{c}{\small{\textbf{Intra-Subject}}}\\
     \footnotesize{Source} & \footnotesize{Diff3F} & \footnotesize{Diff3F + Ours} \\
    \includegraphics[height=0.5\columnwidth]{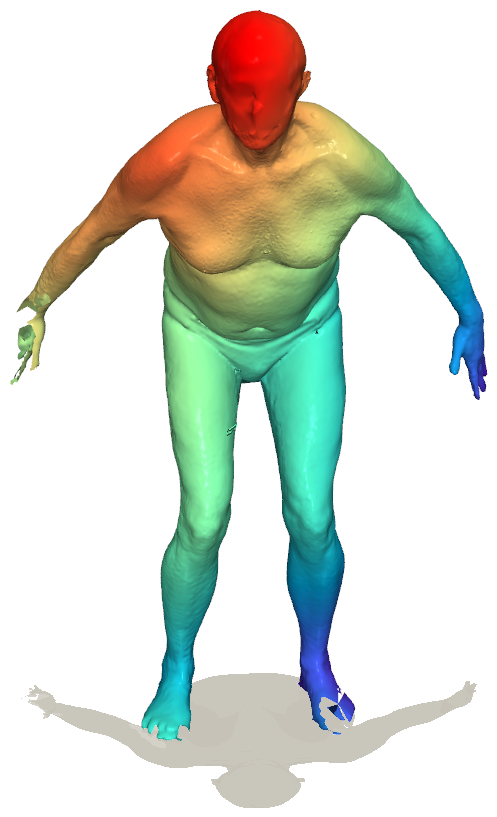} & 
    \imagespacingtff
    \includegraphics[height=0.55\columnwidth]{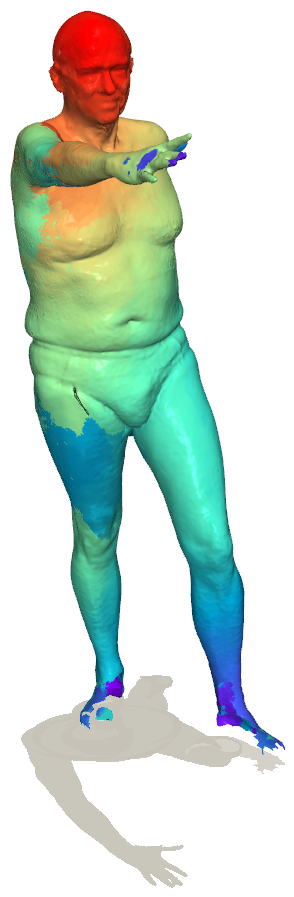} &
    \imagespacingtff
    \includegraphics[height=0.55\columnwidth]{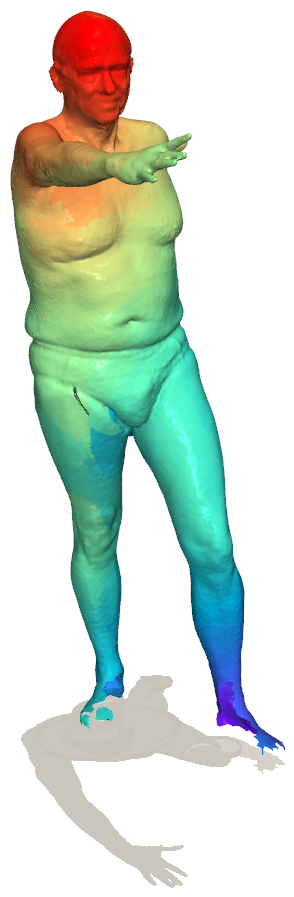}
    \end{tabular}
    \caption{Our method effectively resolves left/right ambiguity when matching the FAUST benchmark. Both in the inter- and intra-subject case.}
    \label{fig:additional_results}
\end{figure}

\section{Implementation details.}
\label{sec:impl}
We employ a lightweight two-layer MLP to implement $g_\Phi$, with a hidden dimension equal to the input dimension ($D = 3968$) and ReLU as the activation function. Experimentally, we find that using normalization on the output feature works better than sigmoid or tanh functions. 
The model is trained on a single NVIDIA A40 GPU using ADAM with a learning rate of $10^{-3}$. We precalculate the input features and run the training for $20000$ iterations, taking around $3$h. All details can be found in the code on \url{https://wei-kang-wang.github.io/chirality/}.

%% file: tables/left_right_results_tanh.tex
\begin{table*}[tbh]
\centering
\resizebox{0.95\textwidth}{!}{%
\setlength{\tabcolsep}{3.5pt}
  \centering
  \begin{tabular}{@{}lccccccccc@{}}
    \toprule
    Train & \textbf{BeCoS} & \multicolumn{2}{c}{\textbf{BeCoS\textsubscript{-h}}} & \multicolumn{2}{c}{\textbf{BeCoS\textsubscript{-a}}} & \multicolumn{2}{c}{\textbf{FAUST}} & \multicolumn{2}{c}{\textbf{SMAL}} \\
    \cmidrule(r){2-2}
    \cmidrule(r){3-4}
    \cmidrule(r){5-6}
    \cmidrule(r){7-8}
    \cmidrule(r){9-10}
    Test & \textbf{BeCoS} & \textbf{BeCoS\textsubscript{-h}} & \textbf{BeCoS\textsubscript{-a}} & \textbf{BeCoS\textsubscript{-h}} & \textbf{BeCoS\textsubscript{-a}} & \textbf{FAUST} & \textbf{SCAPE} & \textbf{SMAL} & \textbf{TOSCA} \\
    \midrule
    $\tanh$ & 75.46 & 92.51 & 83.45 & 73.71 & 75.87 & 91.84 & 94.93 & 71.04 & 68.46 \\
    \changed{Normalisation} & \textbf{91.84} & \textbf{94.09} & \textbf{84.19} & \textbf{90.36} & \textbf{91.10} & \textbf{94.76} & \textbf{95.51} & \textbf{96.59} & \textbf{94.09}\\
    \bottomrule
  \end{tabular}
  }
  \caption{Normalisation of the chirality feature $\chi_v$ with respect to the whole vector results in higher accuracy across all datasets but FAUST, compared to using $\tanh$.}
  \label{tab:chirality_accuracy_tanh}
\end{table*}